\documentclass{article}

\usepackage{microtype}
\usepackage{graphicx}
\usepackage{subfigure}
\usepackage{booktabs} 
\usepackage{amssymb}
\usepackage{amsmath}
\usepackage{amsthm}
\usepackage{mathtools}
\usepackage{enumerate}

\usepackage{hyperref}

\newcommand{\R}{\mathbb{R}}

\newcommand{\bigO}[1]{\mathcal{O}\big(#1\big)}
\newcommand{\pf}[1]{\nabla f (#1)}
\newcommand{\pF}[1]{\nabla F (#1)}
\newcommand{\pfij}[1]{\nabla f_{i_j}(#1)}
\newcommand{\norm}[1]{\lVert#1\rVert}
\newcommand{\normbig}[1]{\big\lVert#1\big\rVert}
\newcommand{\brk}[1]{\lbrack #1 \rbrack}
\newcommand{\Eij}[1]{\mathbb{E}_{i_j} \big[#1\big]}
\newcommand{\EijBig}[1]{\mathbb{E}_{i_j} \Big[#1\Big]}
\newcommand{\E}[1]{\mathbb{E} \big[#1\big]}
\newcommand{\mleq}[1]{\overset{\mathclap{(#1)}}{\leq}}
\newcommand{\meq}[1]{\overset{\mathclap{(#1)}}{=}}
\newcommand{\mgeq}[1]{\overset{\mathclap{(#1)}}{\geq}}
\newcommand{\mar}[1]{(#1)}

\DeclareMathOperator*{\argmin}{arg\,min}

\newtheorem{hood}{Definition}

\newtheorem{mig_hood}{Corollary}

\newtheorem{sconvex}{Assumption}
\newtheorem{nsconvex}[sconvex]{Assumption}
\newtheorem{asyspa_sc}[sconvex]{Assumption}
\newtheorem{asy_assump_indp}{Assumption}[section]
\newtheorem{asy_assump_sample}[asy_assump_indp]{Assumption}

\newtheorem{sconvex_rate}{Theorem}
\newtheorem{nsconvex_rate}[sconvex_rate]{Theorem}
\newtheorem{serial_sparse_frate}[sconvex_rate]{Theorem}
\newtheorem{serial_sparse_rate}[sconvex_rate]{Theorem}
\newtheorem{asy_sparse_rate}[sconvex_rate]{Theorem}

\newtheorem{s_variance}{Lemma}
\newtheorem{variance}[s_variance]{Lemma}
\newtheorem{three_point}[s_variance]{Lemma}
\newtheorem{sc_prox}[s_variance]{Lemma}
\newtheorem{its_variance}[s_variance]{Lemma}

\usepackage[accepted]{icml2018}

\icmltitlerunning{A Simple Stochastic Variance Reduced Algorithm with Fast Convergence Rates}

\begin{document}

\twocolumn[
\icmltitle{A Simple Stochastic Variance Reduced Algorithm with Fast Convergence Rates}



\icmlsetsymbol{equal}{*}

\begin{icmlauthorlist}
\icmlauthor{Kaiwen Zhou}{cuhk}
\icmlauthor{Fanhua Shang}{xidian}
\icmlauthor{James Cheng}{cuhk}

\end{icmlauthorlist}

\icmlaffiliation{cuhk}{Department of Computer Science and Engineering, The Chinese University of Hong Kong, Hong Kong}
\icmlaffiliation{xidian}{School of Artificial Intelligence, Xidian University, China}

\icmlcorrespondingauthor{Fanhua Shang}{fhshang@xidian.edu.cn}

\icmlkeywords{Stochastic Gradient Descent, Sparse, Asynchronous, Variance Reduction}

\vskip 0.3in
]



\printAffiliationsAndNotice{} 

\begin{abstract}
Recent years have witnessed exciting progress in the study of stochastic variance reduced gradient methods (e.g., SVRG, SAGA), their accelerated variants (e.g, Katyusha) and their extensions in many different settings (e.g., online, sparse, asynchronous, distributed). Among them, accelerated methods enjoy improved convergence rates but have complex coupling structures, which makes them hard to be extended to more settings (e.g., sparse and asynchronous) due to the existence of perturbation. In this paper, we introduce a simple stochastic variance reduced algorithm (MiG), which enjoys the best-known convergence rates for both strongly convex and non-strongly convex problems. Moreover, we also present its efficient sparse and asynchronous variants, and theoretically analyze its convergence rates in these settings. Finally, extensive experiments for various machine learning problems such as logistic regression are given to illustrate the practical improvement in both serial and asynchronous settings.
\end{abstract}

\section{Introduction}
\label{introduction}

In this paper, we consider the following convex optimization problem with a finite-sum structure, which is prevalent in machine learning and statistics such as regularized empirical risk minimization (ERM):
\begin{equation}
	\label{prob_def}
	\min_{x\in \mathbb{R}^d} \left\{F(x) \triangleq f(x) + g(x)\right\},
\end{equation}
where $f(x)\!=\!\frac{1}{n}\!\sum_{i=1}^{n}f_i(x)$ is a finite average of $n$ smooth convex function $f_i(x)$, and $g(x)$ is a relatively simple (but possibly non-differentiable) convex function.

For the strongly convex problem~(\ref{prob_def}), traditional gradient descent (GD) yields a linear convergence rate but with a high per-iteration cost. As an alternative, SGD \cite{robbins:sgd} enjoys significantly lower per-iteration complexity than GD, i.e., $O(d)$ vs.\ $O(nd)$. However, due to the variance of random sampling, standard SGD usually obtains slow convergence and poor performance~\cite{johnson:svrg}. Recently, many stochastic variance reduced methods (e.g., SAG~\cite{roux:sag}, SDCA~\cite{shalev-Shwartz:sdca}, SVRG~\cite{johnson:svrg}, SAGA~\cite{defazio:saga}, and their proximal variants, such as~\cite{schmidt:sag}, ~\cite{shalev-Shwartz:acc-sdca}, ~\cite{xiao:prox-svrg} and~\cite{koneeny:mini}) have been proposed to solve Problem~\eqref{prob_def}. All these methods enjoy low per-iteration complexities comparable with SGD, but with the help of certain variance reduction techniques, they obtain a linear convergence rate as GD. More accurately, these methods achieve an improved oracle complexity $\mathcal{O}\!\left((n\!+\!\kappa)\log({1}/{\epsilon})\right)$\footnote{We denote $\kappa \triangleq \frac{L}{\sigma}$ throughout the paper, known as the condition number of an $L$-smooth and $\sigma$-strongly convex function.
}, compared with $\mathcal{O}(n\sqrt{\kappa}\log({1}/{\epsilon}))$ for accelerated deterministic methods (e.g., Nesterov's accelerated gradient descent \cite{nesterov:co}). In summary, these methods dramatically reduce the overall computational cost compared with deterministic methods in theory.

More recently, researchers have proposed accelerated stochastic variance reduced methods for Problem~(\ref{prob_def}), which include Acc-Prox-SVRG~\cite{nitanda:svrg}, APCG~\cite{lin:APCG}, Catalyst~\cite{lin:vrsg}, SPDC~\cite{zhang:spdc}, point-SAGA~\cite{defazio:sagab}, and Katyusha~\cite{zhu:Katyusha}. For strongly convex problems, both Acc-Prox-SVRG \cite{nitanda:svrg} and Catalyst \cite{lin:vrsg} make good use of the ``\textit{Nesterov's momentum}" in \cite{nesterov:co} and attain the corresponding oracle complexities $\mathcal{O}((n\!+\!\!b\sqrt{\kappa})\log({1}/{\epsilon}))$ (with a sufficiently large mini-batch size $b$) and $\mathcal{O}((n\!+\!\!\sqrt{\kappa n})\log(\kappa)\log({1}/{\epsilon}))$. APCG, SPDC, point-SAGA and Katyusha essentially achieve the best-known oracle complexity $\mathcal{O}((n\!+\!\sqrt{\kappa n})\log(1/\epsilon))$.

Inspired by emerging multi-core computer architectures, asynchronous variants of the above stochastic gradient methods have been proposed in recent years, e.g., Hogwild! \cite{rec:hogwild}, Lock-Free SVRG \cite{reddi:sgd}, KroMagnon \cite{man:perturbed} and ASAGA \cite{leb:asaga}. Among them, KroMagnon and ASAGA (as the sparse and asynchronous variants of SVRG and SAGA) enjoy a fast linear convergence rate for strongly convex objectives. However, there still lacks a variant of accelerated algorithms in these settings.

The main issue for those accelerated algorithms is that most of their algorithm designs (e.g., \cite{zhu:Katyusha} and \cite{hien:asmd}) involve tracking at least two highly correlated coupling vectors\footnote{Here we refer to the number of variable vectors involved in one update.} (in the inner loop). This kind of algorithm structure prevents us from deriving efficient (lock-free) asynchronous sparse variants for those algorithms. More critically, when the number of concurrent threads is large (e.g., 20 threads), the high perturbation (i.e., updates on shared variables from concurrent threads) may even destroy their convergence guarantees. This leads us to the key question we study in this paper:

\textbf{Can we design an accelerated algorithm that keeps track of only one variable vector?}

We answer this question by a simple stochastic variance reduced algorithm (MiG), which has the following features:
\begin{itemize}
	\item \textbf{Simple.} The algorithm construction of MiG requires tracking only one variable vector in the inner loop, which means its computational overhead and memory overhead are exactly the same as SVRG (or Prox-SVRG \cite{xiao:prox-svrg}). This feature allows MiG to be extended to more strict settings such as the sparse and asynchronous settings. We theoretically analyze its variants in Section~\ref{sparse_async}.
	\item \textbf{Theoretically Fast.} MiG achieves the best-known oracle complexity of $\bigO{(n\!+\!\!\sqrt{\kappa n})\log{({1}/{\epsilon})}}$ for strongly convex problems. For non-strongly convex problems, MiG also achieves an optimal convergence rate $\bigO{{1}/{T^2}}$, where $T$ is the total number of stochastic iterations. These rates keep up with those of Katyusha and are consistently faster than non-accelerated algorithms, e.g., SVRG and SAGA.
	\item \textbf{Practically Fast.} Due to its light-weighted algorithm structure, our experiments verify that the running time of MiG is shorter than its counterparts in the serial dense setting. In the sparse and asynchronous settings, MiG achieves significantly better performance than KroMagnon and ASAGA in terms of both gradient evaluations and running time.
	\item \textbf{Implementable.} Unlike many incremental gradient methods (e.g., SAGA), MiG does not require an additional gradient table which is not practical for large-scale problems. Our algorithm layout is similar to SVRG, which means that most existing techniques designed for SVRG (such as a distributed variant) can be modified for MiG without much effort.
\end{itemize}
We summarize some properties of the existing methods and MiG in Table~\ref{comparison-table}.
\begin{table}[t]
	\caption{Comparison of different stochastic variance reduced algorithms. (``Complexity'' is for strongly-convex problems. ``Memory'' is those used to store variables.``S\&A'' refers to efficient (lock-free) Sparse \& Asynchronous variant.)}
	\label{comparison-table}
	\begin{center}
		\begin{small}
		\tabcolsep=0.11cm
		\def\arraystretch{1.5}
		\begin{tabular}{cccc}
			Algorithm & Complexity & Memory & S\&A \\
			\midrule
			SVRG&$\bigO{(n+\kappa)\log{\frac{1}{\epsilon}}}$&1 Vec.& $\surd$ \\
			SAGA & $\bigO{(n+\kappa)\log{\frac{1}{\epsilon}}}$&1 Vec. 1$\nabla$ Table.& $\surd$\\
			Katyusha    &$\bigO{(n+\sqrt{\kappa n})\log{\frac{1}{\epsilon}}}$& 2 Vec.& $\times$ \\
			MiG    &$\bigO{(n+\sqrt{\kappa n})\log{\frac{1}{\epsilon}}}$& 1 Vec.&$\surd$         \\
			\bottomrule
		\end{tabular}
		\end{small}
	\end{center}
\end{table}

\section{Notations}
We mainly consider Problem~(\ref{prob_def}) in standard Euclidean space with the Euclidean norm denoted by $\norm{\cdot}$. We use $\mathbb{E}$ to denote that the expectation is taken with respect to all randomness in one epoch. To further categorize the objective functions, we define that a convex function $f:\R^n \rightarrow \R$ is said to be $L$-smooth if for all $x, y \in \R^d$, it holds that
\begin{equation} \label{l-smooth}
	f(x) \leq f(y) + \langle \pf{y}, x-y \rangle + \frac{L}{2} \norm{x-y}^2,
\end{equation}
and $\sigma$-strongly convex if for all $x, y \in \R^d$,
\begin{equation} \label{s-convex}
	f(x) \geq f(y) + \langle \mathcal{G}, x-y \rangle + \frac{\sigma}{2} \norm{x-y}^2,
\end{equation}
where $\mathcal{G}\!\in\!\partial f(y)$, the set of sub-gradient of $f$ at $y$. If $f$ is differentiable, we replace $\mathcal{G}\!\in\!\partial f(y)$ with $\mathcal{G}\!=\!\pf{y}$. Then we make the following assumptions to categorize Problem~(\ref{prob_def}):

\begin{sconvex}[Strongly Convex]\label{assump1}
	In Problem~(\ref{prob_def}), each $f_i(\cdot)$\footnote{In fact, if each $f_i(\cdot)$ is $L$-smooth, the averaged function $f(\cdot)$ is itself $L$-smooth --- but probably with a smaller $L$. We keep using $L$ as the smoothness constant for a consistent analysis.} is $L$-smooth and convex, $g(\cdot)$ is $\sigma$-strongly convex.
\end{sconvex}

\begin{nsconvex}[Non-strongly Convex]\label{assump2}
	In Problem~(\ref{prob_def}), each $f_i(\cdot)$ is $L$-smooth and convex, and $g(\cdot)$ is convex.
\end{nsconvex}

\section{A Simple Accelerated Algorithm}
\label{sequential}
In this section, we introduce a simple accelerated stochastic algorithm (MiG) for both strongly convex and non-strongly convex problems.

\subsection{MiG for Strongly Convex Objectives}
We first consider Problem~(\ref{prob_def}) that satisfies Assumption~\ref{assump1}.

\begin{algorithm}[tb]
	\caption{MiG}
	\label{alg:mig-sc}
	\begin{algorithmic}[1]
		\STATE {\bfseries Input:} Initial vector $x_0$, epoch length $m$, learning rate $\eta$, parameter $\theta$.
		\STATE $\tilde{x}_0 = x^1_0 = x_0$, $\omega = 1 + \eta\sigma$;
		\FOR{$s = 1\ldots \mathcal{S} $}
			\STATE$\mu_s = \nabla f\left(\tilde{x}_{s-1}\right)$;
			\FOR{$j = 1 \ldots m$}
				\STATE Sample $i_j$ uniformly in $\lbrace 1\ldots n\rbrace$;
				\STATE $y_{j-1} = \theta x^s_{j-1} + (1\!-\!\theta) \tilde{x}_{s-1}$;
				$//$temp variable $y$
				\STATE $\tilde{\nabla} = \nabla\! f_{i_j}(y_{j-1}) - \nabla\! f_{i_j}(\tilde{x}_{s-1}) + \mu_s$;
				\STATE $x^s_j = \argmin_{x} \!\left\lbrace\! \frac{1}{2\eta} \norm{x - x^s_{j-\!1}}^2 \!+\! \langle \tilde{\nabla}\!, x \rangle \!+\! g(x) \!\right\rbrace$;
			\ENDFOR
			\STATE $\tilde{x}_s = \theta{\big(\sum_{j=0}^{m-1}\!\omega^{j}\big)}^{-1}\! \sum_{j = 0}^{m-1}\!{\omega^{j} x^s_{j+1}}\! + (1\!-\!\theta) \tilde{x}_{s-1}$;
			\STATE $x^{s+1}_0 = x^s_m$;
		\ENDFOR
		\STATE \textbf{return} $\tilde{x}_{\mathcal{S}}$.
	\end{algorithmic}
\end{algorithm}

Now we formally introduce MiG in Algorithm~\ref{alg:mig-sc}. In order to further illustrate some ideas behind the algorithm structure, we make the following remarks:

\begin{itemize}
	\item \textit{Temp variable $y$.} As we can see in Algorithm~\ref{alg:mig-sc}, $y$ is a convex combination of $x$ and $\tilde{x}$ with the parameter $\theta$. So for implementation, we do not need to keep track of $y$ in the whole inner loop. For the purpose of giving a clean proof, we mark $y$ with iteration number $j$.
	\item \textit{Fancy update for $\tilde{x}_s$.} One can easily verify that this update for $\tilde{x}_s$ is equivalent to using $\omega^j$ weighted averaged $y_{j+1}$ to update $\tilde{x}_s$, which is written as:
	$\tilde{x}_s \!= \!{\big(\sum_{j=0}^{m-1} \omega^{j}\big)}^{-1} \!\sum_{j = 0}^{m-1}{\omega^{j} y_{j+1}}$. Since we only keep track of $x$, we adopt this expended fancy update for $\tilde{x}_s$ --- but it is still quite simple in implementation.
	\item \textit{Choice of $x^{s+1}_0$.} In recent years, some existing stochastic algorithms such as~\cite{zhang:svrg,xiao:prox-svrg} choose to use $\tilde{x}_s$ as the initial vector for new epoch. For MiG, when using $\tilde{x}_s$, the overall oracle complexity will degenerate to a non-accelerated one for some ill-conditioned problems, which is $\bigO{(n\!+\!\kappa)\log{({1}/{\epsilon})}}$. It is reported that even in practice, using the last iterate yields a better performance as discussed in \cite{zhu:vrnc}.
\end{itemize}

Next we give the convergence rate of MiG in terms of oracle complexity as follows (the proofs to all theorems in this paper are given in the Supplementary Material):

\begin{sconvex_rate}[Strongly Convex]\label{theorem1}
Let $x^{*}$ be the optimal solution of Problem~(\ref{prob_def}). If Assumption~\ref{assump1} holds, then by choosing $m=\Theta(n)$, \textup{MiG} achieves an $\epsilon$-additive error with the following oracle complexities in expectation:
	\[
		\begin{cases}
			\bigO{\sqrt{\kappa n} \log{\frac{F(x_0) - F(x^*)}{\epsilon}}}, & \text{if }\;\; \frac{m}{\kappa} \leq \frac{3}{4}, \\
			\bigO{n  \log{\frac{F(x_0) - F(x^*)}{\epsilon}}}, & \text{if } \;\;\frac{m}{\kappa} > \frac{3}{4}.
		\end{cases}
	\]
	In other words, the overall oracle complexity of \textup{MiG} is $\bigO{(n+\!\sqrt{\kappa n})\log{\frac{F(x_0) - F(x^*)}{\epsilon}}}$.
\end{sconvex_rate}

\begin{table}[t]
	\caption{The theoretical settings of the parameters $\eta$ and $\theta$.}
	\label{sc_parameter-table}
	\begin{center}
		\begin{small}
			\begin{sc}
				\begin{tabular}{ccc}
					\toprule
					Condition &Learning rate $\eta$ & Parameter $\theta$\\
					\midrule
					$\frac{m}{\kappa} \leq \frac{3}{4}$ &$\sqrt{\frac{1}{3\sigma m L}}$ & $\sqrt{\frac{m}{3\kappa}}$  \\
					$\frac{m}{\kappa} > \frac{3}{4}$ &$\frac{2}{3L}$ &$\frac{1}{2}$\\
					\bottomrule
				\end{tabular}
			\end{sc}
		\end{small}
	\end{center}
	\vskip -0.1in
\end{table}

This result implies that in the strongly convex setting, MiG enjoys the best-known oracle complexity for stochastic first-order algorithms, e.g., APCG, SPDC, and Katyusha. The theoretical suggestions\footnote{We recommend users to tune these two parameters for better performance in practice, or to use the tuning criteria mentioned in Table~\ref{Moment_table} with only tuning $\theta$.} of the learning rate $\eta$ and the parameter $\theta$ are shown in Table~\ref{sc_parameter-table}.

\subsubsection{Comparison Between MiG And Related Methods}
\label{comparisons}
We carefully compare the algorithm structure of MiG with Katyusha, and find that MiG corresponds to a case of Katyusha, when $1 \!-\! \tau_1\! -\! \tau_2\! =\! 0$, and Option II in Katyusha is used. However, this setting is neither suggested nor analyzed in \cite{zhu:Katyusha}, and thus without a convergence guarantee. In some sense, MiG can be regarded as a ``simplified Katyusha'', while this simplification does not hurt its oracle complexity. Since this simplification discards all the proximal gradient steps in Katyusha, MiG enjoys a lower memory overhead in practice and a cleaner proof in theory. Detailed comparison with Katyusha can be found in the Supplementary Material~\ref{detailed_comp_with_katyu}.

MiG does not use any kind of ``\textit{Nesterov's Momentum}", which is used in some accelerated algorithms, e.g., Acc-Prox-SVRG \cite{nitanda:svrg} and Catalyst \cite{lin:vrsg}.


\subsection{MiG for Non-strongly Convex Objectives}
In this part, we consider Problem~(\ref{prob_def}) when Assumption~\ref{assump2} holds. Since non-strongly convex optimization problems (e.g., LASSO) are becoming prevalent these days, making a direct variant of MiG for these problems is of interest.

In this setting, we summarize MiG\textsuperscript{NSC} with the optimal $\bigO{{1}/{T^2}}$ convergence rate in Algorithm~\ref{alg:mig-nsc}.

\begin{algorithm}[tb]
	\caption{MiG\textsuperscript{NSC}}
	\label{alg:mig-nsc}
	\begin{algorithmic}[1]
		\STATE {\bfseries Input:} Initial vector $x_0$, epoch length $m$, learning rate $\eta$, parameter $\theta$.
		\STATE $\tilde{x}_0 = x^1_0 = x_0$;
		\FOR{$s = 1\ldots \mathcal{S} $}
		\STATE $\mu_s = \nabla f\left(\tilde{x}_{s-1}\right)$, $\theta = \frac{2}{s + 4}$, $\eta = \frac{1}{4L\theta}$;
		\FOR{$j = 1 \ldots m$}
		\STATE Sample $i_j$ uniformly in $\lbrace 1\ldots n\rbrace$;
		\STATE $y_{j-1} = \theta x^s_{j-1} + (1 - \theta) \tilde{x}_{s-1}$;
		$//$temp variable $y$
		\STATE $\tilde{\nabla} = \nabla\! f_{i_j}(y_{j-1}) - \nabla\! f_{i_j}(\tilde{x}_{s-1}) + \mu_s$;
		\STATE $x^s_j = \argmin_{x} \left\lbrace \frac{1}{2\eta} \!\norm{x \!-\! x^s_{j-\!1}}^2 \!+\! \langle \tilde{\nabla}\!, x \rangle\! +\! g(x) \right\rbrace$;
		\ENDFOR
		\STATE $\tilde{x}_s =  \frac{\theta}{m}\! \sum_{j=1}^{m} {x^s_j} + (1\! - \!\theta) \tilde{x}_{s-1}$;
		\STATE $x^{s+1}_0 = x^s_m$;
		\ENDFOR
		\STATE \textbf{return} $\tilde{x}_{\mathcal{S}}$.
	\end{algorithmic}
\end{algorithm}

\begin{nsconvex_rate}[Non-strongly Convex]
\label{nsconvex_rate}
	If Assumption~\ref{assump2} holds, then by choosing $m\!=\!\Theta(n)$, \textup{MiG\textsuperscript{NSC}} achieves the following oracle complexity in expectation:
	\[
		\mathcal{O}\!\left(n\sqrt{\frac{F(x_{0}) - F(x^*)}{\epsilon}} + \sqrt{\frac{nL\norm{x_{0} - x^*}^2}{\epsilon}}\right).
	\]
This result implies that \textup{MiG\textsuperscript{NSC}} attains the optimal convergence rate $\mathcal{O}({1}/{T^2})$, where $T \!=\!\mathcal{S}(m\!+\!n)$ is the total number of stochastic iterations.
\end{nsconvex_rate}

The result in Theorem~\ref{nsconvex_rate} shows that \textup{MiG\textsuperscript{NSC}} enjoys the same oracle complexity as Katyusha\textsuperscript{ns} \cite{zhu:Katyusha}, which is close to the best-known complexity in this case\footnote{Note that the best-known oracle complexity for non-strongly convex problems is $\mathcal{O}(n\log(1/\epsilon)\!+\!\sqrt{nL/\epsilon})$.}. If the reduction techniques in \cite{zhu:box,xu:hs} are used in our algorithm, our algorithm can obtain the best-known oracle complexity.

\subsection{Extensions}
It is common to apply reductions to extend the algorithms designed for $L$-smooth and $\sigma$-strongly convex objectives to other cases (e.g., non-strongly convex or non-smooth). For example, \citet{zhu:box} proposed several reductions for the algorithms that satisfy \textit{homogeneous objective decrease} (\textsf{HOOD}), which is defined as follows.

\begin{hood}[\citet{zhu:box}]\label{hood}
An algorithm $\mathcal{A}\!$ that solves Problem~(\ref{prob_def}) with Assumption~\ref{assump1} satisfies \textup{\textsf{HOOD}} if, for every starting vector $x_0$, $\mathcal{A}$ produces an output $x'$ satisfying\footnote{This definition can be extended to the probabilistic guarantee, which is $\E{F(x')}\!-\!F(x^*)\!\leq\!\frac{F(x_0)- F(x^*)}{4}$ \cite{zhu:box}.} $F(x')\!-\! F(x^*)\!\leq\! \frac{F(x_0)- F(x^*)}{4}$ in \textup{\textsf{Time($L$, $\sigma$)}}.
\end{hood}
Based on Theorem~\ref{theorem1}, it is a direct corollary that MiG (refers to Algorithm~\ref{alg:mig-sc}) satisfies \textsf{HOOD}.
\begin{mig_hood}\label{mig_hood}
	\textup{MiG} satisfies the \textup{\textsf{HOOD}} property in \textup{\textsf{Time($n+\sqrt{\kappa n}$)}}.
\end{mig_hood}

The reductions in~\cite{zhu:box} use either decaying regularization (\textsf{AdaptReg}) or certain smoothing trick (\textsf{AdaptSmooth}) to achieve optimal reductions that shave off a non-optimal log factor comparing to other reduction techniques. Thus, we can apply \textsf{AdaptReg} to MiG and get an improved $\mathcal{O}(n\log(1/\epsilon)\!+\!\sqrt{nL/\epsilon})$ rate for non-strongly convex problems. Moreover, we can also apply \textsf{AdaptSmooth} to MiG to tackle non-smooth optimization problems, e.g., SVM.

\section{Sparse and Asynchronous Variants}\label{sparse_async}

In order to further elaborate the importance of keeping track of only one variable vector (in the inner loop), in this section we propose the variants of MiG for both the serial sparse and asynchronous sparse settings.

Adopting the sparse update technique in \cite{man:perturbed} for sparse datasets is a very practical choice to reduce collisions between threads. However, due to additional sparse approximating variance and asynchronous perturbation, we need to compensate it with a slower theoretical speed. On the other hand, asynchrony (in the lock-free style) may even destroy convergence guarantees if the algorithm requires tracking many highly correlated vectors. In practice, it is reported that maintaining more atomic\footnote{Atomic write of some necessary variables is a requirement to achieve high precision in practice \cite{leb:asaga}.} variables also degrades the performance. Thus, in this section, we mainly focus on practical issues and experimental performance.

As we can see, MiG has only one variable vector. This feature gives us convenience in both theoretical analysis and practical implementation. In order to give a clean proof, we first make a simpler assumption on the objective function, which is identical to those in~\cite{rec:hogwild,man:perturbed,leb:asaga}:
\begin{equation}\label{prob_def2}
	\min_{x\in \mathbb{R}^d} {F(x) \triangleq \frac{1}{n} \sum_{i=1}^{n} f_i(x)}.
\end{equation}
\begin{asyspa_sc}[Sparse and Asynchronous Settings]\label{asyspa_assump}
	In Problem~(\ref{prob_def2}), each $f_i(\cdot)$ is $L$-smooth, and the averaged function $F(\cdot)$ is $\sigma$-strongly convex.
\end{asyspa_sc}

Next we start with analyzing MiG in the serial sparse setting and then extend it to a sparse and asynchronous one.

\subsection{Serial Sparse MiG}
\label{sec_serial_sparse_mig}

\begin{algorithm}[tb]
	\caption{Serial Sparse MiG}
	\label{sparse_mig}
	\begin{algorithmic}[1]
		\STATE {\bfseries Input:} Initial vector $x_0$, epoch length $m$, learning rate $\eta$, parameter $\theta$.
		\STATE $\tilde{x}_0 = x^1_0 = x_0$;
		\FOR{$s = 1\ldots \mathcal{S} $}
			\STATE$\mu_s = \nabla F\left(\tilde{x}_{s-1}\right)$;
			\FOR{$j = 1 \ldots m$}
				\STATE $T_{i_j}:=$ support of sample $i_j$;
				\STATE $\brk{y_{j-1}}_{T_{i_j}} \!= \theta\cdot \brk{x^s_{j-1}}_{T_{i_j}} + (1 - \theta)\cdot\brk{\tilde{x}_{s-1}}_{T_{i_j}}$;
				\STATE $\tilde{\nabla}_{\!S} \!=\! \nabla\!f_{i_j}(\brk{y_{j-1}}_{T_{i_j}}) - \nabla\!f_{i_j}(\brk{\tilde{x}_{s-1}}_{T_{i_j}}) + D_{i_j}\mu_s$;
				\STATE $\brk{x^s_j}_{T_{i_j}} = \brk{x^s_{j-1}}_{T_{i_j}} - \eta \cdot \tilde{\nabla}_{\!S}$;
			\ENDFOR
			\STATE Option I: $x^{s+1}_0 \!= \tilde{x}_{s} = \frac{\theta}{m} \!\sum_{j=0}^{m-1} \!{x^s_j} + \!(1\! -\! \theta)\tilde{x}_{s-1}$;
			\STATE Option II: $x^{s+\!1}_0 \!=\! x^s_m$, $\tilde{x}_{s} \!=\! \frac{\theta}{m}\! \sum_{j=1}^{m} \!{x^s_j} \!+\! (1 \!-\! \theta) \tilde{x}_{s-\!1}$;
		\ENDFOR
		\STATE \textbf{return} $\tilde{x}_{\mathcal{S}}$.
	\end{algorithmic}
\end{algorithm}

The sparse variant (as shown in Algorithm~\ref{sparse_mig}) of MiG is slightly different from MiG in the dense case. We explain these differences by making the following remarks:

\begin{itemize}
	\item \textit{Sparse approximate gradient $\tilde{\nabla}_{\!S}$.} In order to perform fully sparse updates, following~\cite{man:perturbed}, we use a diagonal matrix $D$ to re-weigh the dense vector $\mu_s$, whose entries are the inverse probabilities $\lbrace p_k^{-1} \rbrace$ of the corresponding coordinates $\lbrace k\!\mid\! k\!=\!1,\ldots, d\rbrace$ belonging to a uniformly sampled support $T_{i_j}$ of sample $i_j$. $P_{i_j}$ is the projection matrix for the support $T_{i_j}$. We define $D_{i_j} \!= \!P_{i_j}D$, which ensures the unbiasedness $\Eij{D_{i_j}\mu_s} \!=\! \mu_s$. Here we also define $D_m \!=\! \max_{k= 1\ldots d}{p_k^{-1}}$ for future usage. Note that we only need to compute $y$ on the support of sample $i_j$, and hence the entire inner loop updates sparsely.
	\item \textit{Update $\tilde{x}$ with uniform average.} In the sparse and asynchronous setting, a weighted average in Algorithm~\ref{alg:mig-sc} is not effective due to the perturbation both in theory and in practice. Thus, we choose a simple uniform average scheme for a better practical performance.
\end{itemize}
%
We now consider the convergence property of Algorithm~\ref{sparse_mig}.

\begin{serial_sparse_rate}[\textup{Option I}]\label{serial_sparse_rate}
Let $x^{*}$ be the optimal solution of Problem~(\ref{prob_def2}). If Assumption~\ref{asyspa_assump} holds, then by choosing $\eta \!=\!{1}/{L}$, $\theta\! =\! {1}/{10}$, $m \!=\! 25\kappa$, Algorithm~\ref{sparse_mig} with Option I satisfies the following inequality in one epoch $s$:
	\[
		\E{\big(F(\tilde{x}_s) - F(x^*)\big)} \leq 0.75\cdot \big(F(\tilde{x}_{s-1}) - F(x^*)\big),
	\]
	which means that the total oracle complexity of the serial sparse \textup{MiG} is $\bigO{(n+\kappa)\log{\frac{F(x_0) - F(x^*)}{\epsilon}}}$.
\end{serial_sparse_rate}

Since it is natural to ask whether we can get an improved bound for the Serial Sparse MiG, we analyze Algorithm~\ref{sparse_mig} with Option II and a somewhat intriguing restart scheme. The convergence result is given as follows:

\begin{serial_sparse_frate}[\textup{Option II}]\label{serial_sparse_frate}
If Assumption~\ref{asyspa_assump} holds, then by executing Algorithm~\ref{sparse_mig} with Option II and restarting\footnote{We set $x_0 \!=\!\frac{1}{\mathcal{S}} \sum_{s=1}^{\mathcal{S}} {\tilde{x}_s}$ as the initial vector after each restart.} the algorithm every $\mathcal{S}\!=\!\Big\lceil{2\cdot\frac{(1-\theta)\cdot( 1 + \zeta ) + \frac{\theta}{\eta m\sigma}}{\theta + \zeta\theta - \zeta}}\Big\rceil$ epochs, where $\zeta \!=\! D_m^2 \!-\! D_m$, the oracle complexity of the entire procedure is divided into the cases,
	\[
	\begin{cases}
		\bigO{\sqrt{\kappa n} \log{\frac{F(x_0) - F(x^*)}{\epsilon}}}, & \text{if }\: \frac{m}{\kappa} \leq \frac{3}{4}\text{ with }\zeta \leq \sqrt{\frac{m}{4\kappa}}, \\
		\bigO{n  \log{\frac{F(x_0) - F(x^*)}{\epsilon}}}, & \text{if }\: \frac{m}{\kappa} > \frac{3}{4}\text{ with }\zeta \leq C_{\zeta},
	\end{cases}
	\]
	where $C_{\zeta}$ is a constant for the sparse estimator variance. Detailed parameter settings are given in the Supplementary Material~\ref{proof_T3}.
\end{serial_sparse_frate}

\textit{Remark:} This result indeed imposes some strong assumptions on $D_m$, which may not be true for real world datasets, because the variance bound used for Option II highly correlates to $D_m$, and $D_m$ can be as large as $n$ for extreme datasets. Detailed discussion is given in the Supplementary Material~\ref{discuss_variance_bound}.

The result of Theorem~\ref{serial_sparse_frate} shows that under some constraints on sparse variance, Serial Sparse MiG attains a faster convergence rate than Sparse SVRG~\cite{man:perturbed} and Sparse SAGA~\cite{leb:asaga}. Although these constraints are strong and the restart scheme is not quite practical, we keep the result here as a reference for both the sparse ($\zeta \!> \!0$) and dense ($\zeta \!=\! 0$) cases.


\subsection{Asynchronous Sparse MiG}
In this part, we extend the Serial Sparse MiG to the Asynchronous Sparse MiG.

Our algorithm is given in Algorithm~\ref{async_mig}. Notice that Option I and II correspond to the update options mentioned in Algorithm~\ref{sparse_mig}. The difference is that Option II corresponds to averaging ``fake'' iterates defined at~\eqref{fake_iterates}, while Option I is the average of inconsistent read\footnote{We could use ``\textit{fake average}'' in Option I, but it leads to a complex proof and a worse convergence rate with factor ($\propto \kappa^{-2}$).} of $x$. Since the averaging scheme in Option II is not proposed before, we refer to it as ``\textit{fake average}''. Just like the analysis in the serial sparse case, Option I leads to a direct and clean proof while Option II may require restart and leads to troublesome theoretical analysis. So we only analyze Option I in this setting.

However, Option II is shown to be highly practical since the ``\textit{fake average}'' scheme only requires updates on the support of samples and enjoys strong robustness when the actual number of inner loops does not equal to $m$\footnote{This phenomenon is prevalent in the asynchronous setting.}. Thus, Option II leads to a very practical implementation.
\begin{algorithm}[t]
	\caption{Asynchronous Sparse MiG}
	\label{async_mig}
	\begin{algorithmic}[1]
		\STATE {\bfseries Input:} Initial vector $x_0$, epoch length $m$, learning rate $\eta$, parameter $\theta$.
		\STATE $x:=$ shared variable, $\bar{x}:=$ average of $x$;
		\STATE $\tilde{x}_0 = x = x_0$;
		\FOR{$s = 1\ldots \mathcal{S} $}
		\STATE Compute $\mu_s = \nabla F\left(\tilde{x}_{s-1}\right)$ in parallel;
		\STATE Option I: $\bar{x} = \mathbf{0}$;
		\STATE Option II: $\bar{x} = x$;
		\STATE $j=0$; \COMMENT{inner loop counter}
		\WHILE[in parallel]{$j < m$}
		\STATE $j = j + 1$; $\ \ \ \ //$ atomic increase counter $j$
		\STATE Sample $i_j$ uniformly in $\lbrace 1\ldots n\rbrace$;
		\STATE $T_{i_j}\!:=$ support of sample $i_j$;
		\STATE $\brk{\hat{x}}_{T_{i_j}}\!:=$ inconsistent read of $\brk{x}_{T_{i_j}}$;
		\STATE $\brk{\hat{y}}_{T_{i_j}} \!= \theta\cdot\brk{\hat{x}}_{T_{i_j}} + (1 - \theta)\cdot \brk{\tilde{x}_{s-1}}_{T_{i_j}}$;
		\STATE $\tilde{\nabla}(\hat{y}) = \nabla\! f_{i_j}(\brk{\hat{y}}_{T_{i_j}}) - \nabla\! f_{i_j}(\brk{\tilde{x}_{s-1}}_{T_{i_j}}) + D_{i_j}\mu_s$;
		\STATE $\brk{u}_{T_{i_j}} \!= - \eta \cdot \tilde{\nabla}(\hat{y})$;
		\STATE $//$ atomic write $x$, $\bar{x}$ for each coordinate
		\STATE $\brk{x}_{T_{i_j}}\! = \brk{x}_{T_{i_j}} + \brk{u}_{T_{i_j}}$;
		\STATE Option I: $\bar{x} = \bar{x} + \frac{1}{m} \cdot \hat{x}$;
		\STATE Option II: $\brk{\bar{x}}_{T_{i_j}} \!= \brk{\bar{x}}_{T_{i_j}} + \brk{u}_{T_{i_j}} \!\cdot \frac{(m + 1 - j)_+}{m}$;
		\ENDWHILE
		\STATE $\tilde{x}_s = \theta \bar{x} + (1 - \theta) \tilde{x}_{s-1}$;
		\STATE Option I: $x = \tilde{x}_{s}$;
		\STATE Option II: keep $x$ unchanged;
		\ENDFOR
		\STATE \textbf{return} $\tilde{x}_{\mathcal{S}}$.
	\end{algorithmic}
\end{algorithm}

Following~\cite{man:perturbed}, our analysis is based on the ``fake'' iterates $x$ and $y$, which are defined as:
\begin{equation}\label{fake_iterates}
	x_{j} = x_0 - \eta \sum_{i=0}^{j-1} \tilde{\nabla}(\hat{y}_i)\text{, }y_{j} = \theta x_{j} + (1 - \theta) \tilde{x}_{s-1},
\end{equation}
where the ``perturbed'' iterates $\hat{y}$, $\hat{x}$ with perturbation $\xi$ are defined as
\begin{equation}\label{async_petb}
	\hat{y}_j = \theta \hat{x}_j + (1 - \theta) \tilde{x}_{s-1}\text{, }\hat{x}_j = x_j + \xi_j.
\end{equation}
The labeling order and detailed analysis framework are given in the Supplementary Material~\ref{async_analysis}.

Note that $y$ is a temp variable, so the only source of perturbation comes from $x$. This is the benefit of keeping track of only one variable vector since it controls the perturbation and allows us to give a smooth analysis in asynchrony.

Next we give our convergence result as follows:
\begin{asy_sparse_rate}\label{asy_sparse_rate}
	If Assumption~\ref{asyspa_assump} holds, then by choosing $m \!=\! 60\kappa$, $\eta\! =\!{1}/{(5L)}$, $\theta\! = \!{1}/{6}$, suppose $\tau$ satisfies $\tau \!\leq\! \min{\lbrace \frac{5}{4\sqrt{\Delta}}, 2\kappa, \sqrt{\frac{2\kappa}{\sqrt{\Delta}}} \rbrace}$ (the linear speed-up condition), Algorithm~\ref{async_mig} with Option I has the following oracle complexity:
	\[
		\mathcal{O}\!\left((n+\kappa)\log{\frac{F(x_0) - F(x^*)}{\epsilon}}\right),
	\]
	where $\tau$ represents the maximum number of overlaps between concurrent threads~\cite{man:perturbed} and $\Delta\!=\! \max_{k=1\ldots d}{p_k}$, which is a measure of sparsity~\cite{leb:asaga}.
\end{asy_sparse_rate}
This result is better than that of KroMagnon, which correlates to $\kappa^2$~\cite{man:perturbed}, and keeps up with ASAGA~\cite{leb:asaga}. Although without significant improvement on theoretical bounds due to the existence of perturbation, the coupling step of MiG can still be regarded as a simple add-on boosting and stabilizing the performance of SVRG variants. We show this improvement by empirical evaluations in Section~\ref{async_experiments}.

\section{Experiments}\label{experiments}
In this section, we evaluate the performance of MiG on real-world datasets for both serial dense and asynchronous sparse\footnote{Experiments for the serial sparse variant are omitted since it corresponds to the asynchronous sparse variant with $1$ thread.} cases. All the algorithms were implemented in C++ and executed through MATLAB interface for a fair comparison. Detailed experimental setup is given in the Supplementary Material~\ref{experiment_setup}.

We first give a detailed comparison between MiG and other algorithms in the sequential dense setting.

\begin{figure}[t]
\begin{center}
\subfigure[Relatively small regularization parameter $\lambda \!=\! 10^{-8}$]{
			\includegraphics[width=\columnwidth / 2]{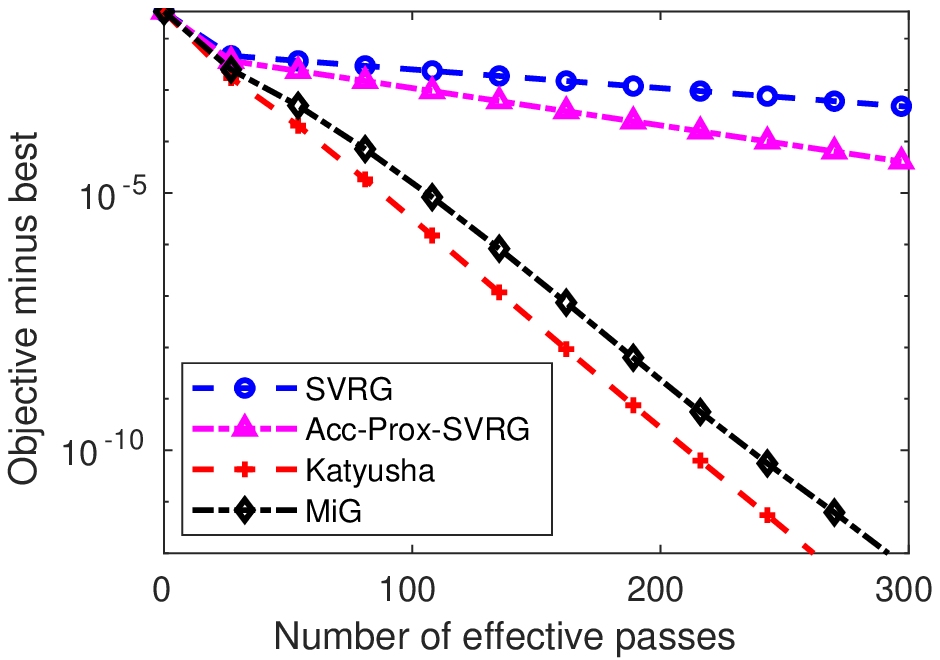}
			\includegraphics[width=\columnwidth / 2]{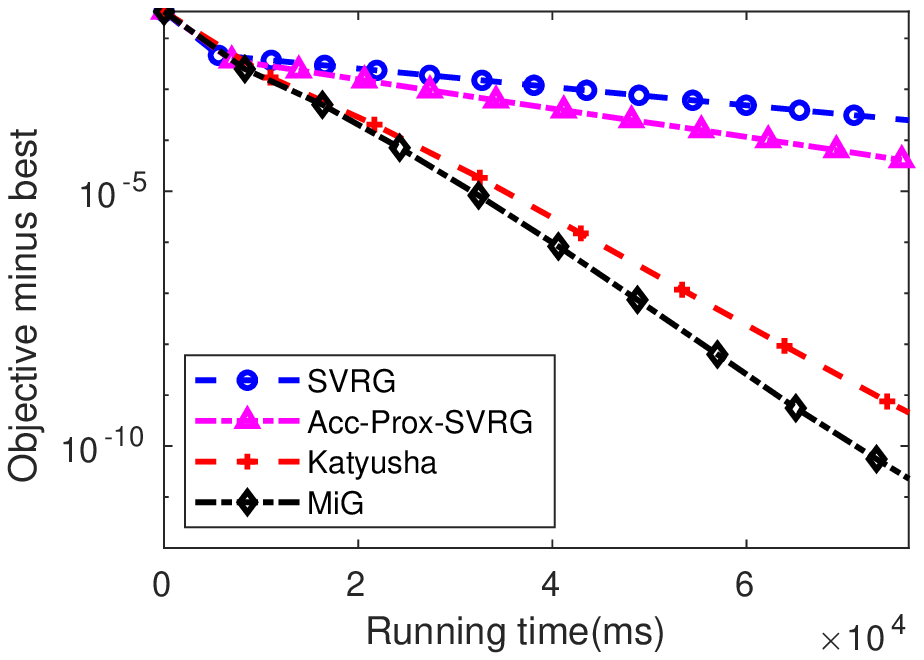}}
\subfigure[Relatively large regularization parameter $\lambda\! =\! 10^{-4}$]{
			\includegraphics[width=\columnwidth / 2, height=1.165in]{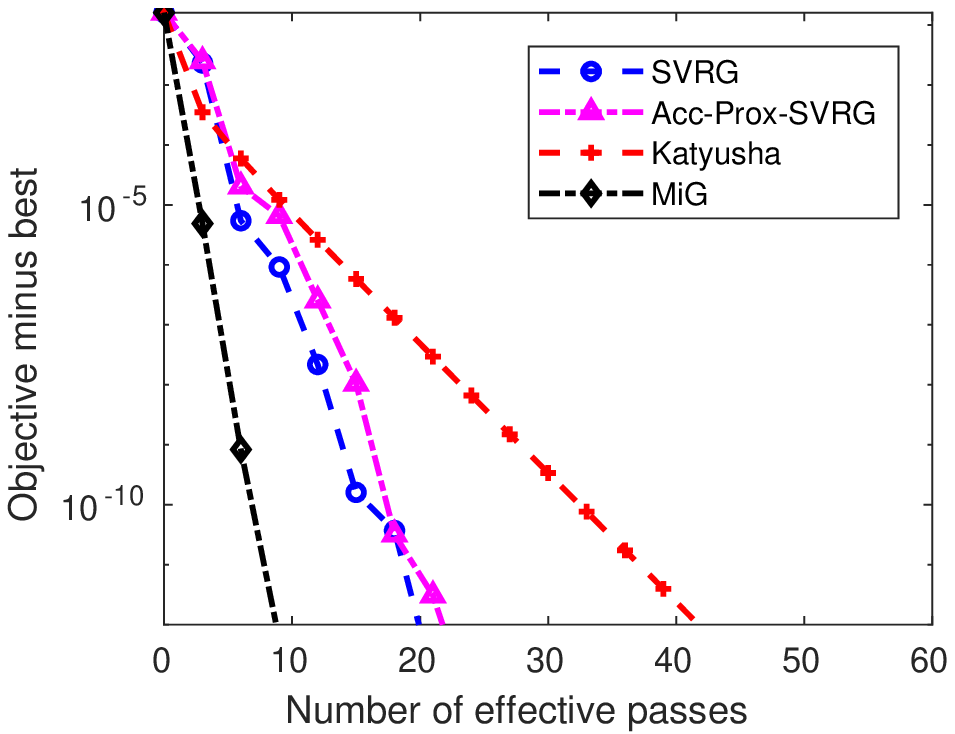}
			\includegraphics[width=\columnwidth / 2, height=1.165in]{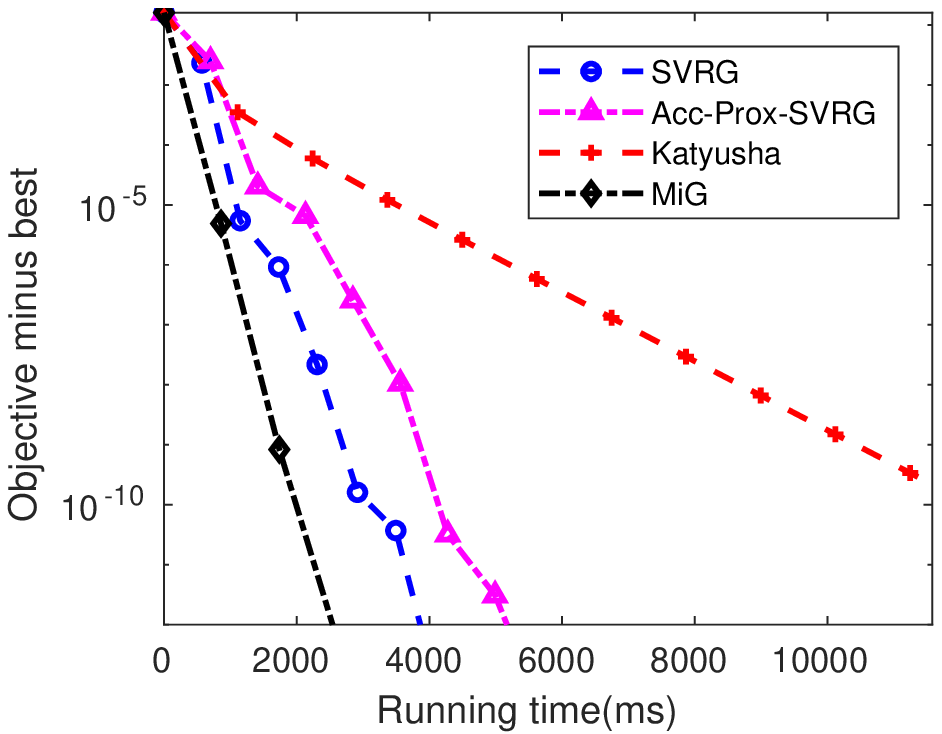}}

\caption{Comparison of the effect of the momentum techniques used in Acc-Prox-SVRG~\cite{nitanda:svrg}, Katyusha~\cite{zhu:Katyusha} and MiG for $\ell2$-logistic regression on \textsf{covtype}.}
\label{MiG_Katyusha}
\end{center}
\end{figure}

\subsection{Comparison of Momentums}

The parameter $\theta$ in MiG, similar to the parameter $\tau_2$ in Katyusha, is referred as the parameter for ``\textit{Katyusha Momentum}'' in~\cite{zhu:Katyusha}. So intuitively, MiG can be regarded as adding ``\textit{Katyusha Momentum}'' on top of SVRG. Katyusha equipped with the linear coupling framework in~\cite{zhu:linear} and thus can be regarded as the combination of ``\textit{Nesterov's Momentum}'' with ``\textit{Katyusha Momentum}''.

We empirically evaluate the effect of the two kinds of momentums. Moreover, we also examine the performance of Acc-Prox-SVRG~\cite{nitanda:svrg}, which can be regarded as SVRG with pure ``\textit{Nesterov's Momentum}''. Note that the mini-batch size used in Acc-Prox-SVRG is set to $1$. The algorithms and parameter settings are listed in Table~\ref{Moment_table} (we use the same notations as in their original papers).

\begin{table}[t]
	\begin{center}
		\caption{Compared algorithms and parameter tuning criteria.}
		\vskip 0.15in
		\label{Moment_table}
		\tabcolsep=0.05cm
		\def\arraystretch{1.3}
		\begin{small}
			\begin{tabular}{lcc}
				\toprule
				& Momentum& Parameter Tuning \\
				\midrule
				SVRG           & None   &  learning rate $\eta$\\
				Acc-Prox-SVRG  & Nestrv.& same $\eta$, tune momentum $\beta$\\
				Katyusha      & Nestrv.\&Katyu.& $\tau_2 \!=\! \frac{1}{2}$, $\alpha \!=\! \frac{1}{3\tau_1 L}$, tune $\tau_1$ \\
				MiG			   & Katyu.  & $\eta \!=\! \frac{1}{3\theta L}$, tune $\theta$ \\
				\bottomrule
			\end{tabular}
		\end{small}
	\end{center}
\end{table}

The results in Figure~\ref{MiG_Katyusha} correspond to the two typical conditions with relatively large $\lambda\!=\!10^{-4}$ and relatively small $\lambda\!=\!10^{-8}$. One can verify that with the epoch length $m \!= \!2n$ for all algorithms, these two conditions fall into the two regions correspondingly in Table~\ref{sc_parameter-table}.

For the case of $\frac{m}{\kappa} \!\leq\! \frac{3}{4}$ (see the first row in Figure~\ref{MiG_Katyusha}), we set the parameters for MiG and Katyusha\footnote{We choose to implement Katyusha with Option I, which is analyzed theoretically in~\cite{zhu:Katyusha}.} with their theoretical suggestions (e.g., $\theta\!=\!\tau_1\!=\!\!\sqrt{\frac{m}{3\kappa}}$). For fair comparison, we set the learning rate $\eta \!=\! \frac{1}{4L}$ for SVRG and Acc-Prox-SVRG, which is theoretically reasonable. The results imply that MiG and Katyusha have close convergence results and outperform SVRG and Acc-Prox-SVRG. This justifies the improvement of $\sqrt{\kappa n}$ convergence rate in theory.

We notice that Katyusha is slightly faster than MiG in terms of the number of oracle calls, which is reasonable since Katyusha has one more ``\textit{Nesterov's Momentum}". From the result of Acc-Prox-SVRG, we see that ``\textit{Nesterov's Momentum}" is effective in this case, but without significant improvement. As analyzed in~\cite{nitanda:svrg}, using large enough mini-batch is a requirement to make Acc-Prox-SVRG improve its convergence rate in theory (see Table 1 in~\cite{nitanda:svrg}), which also explains the limited difference between MiG and Katyusha.

When comparing running time (millisecond, ms), MiG outperforms other algorithms. Using ``\textit{Nesterov's Momentum}" requires tracking of at least two variable vectors, which increases both memory consumption and computational overhead. More severely, it prevents the algorithms with this trick to have an efficient sparse and asynchronous variant.

\begin{figure}[t]
	\begin{center}
		\centerline{
			\includegraphics[width=\columnwidth / 2]{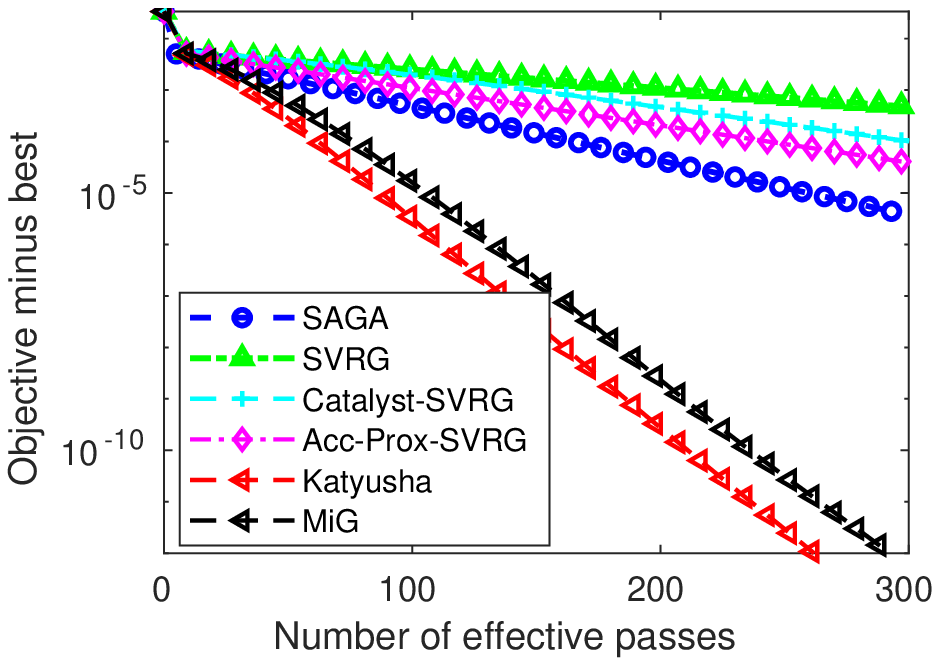}
			\includegraphics[width=\columnwidth / 2]{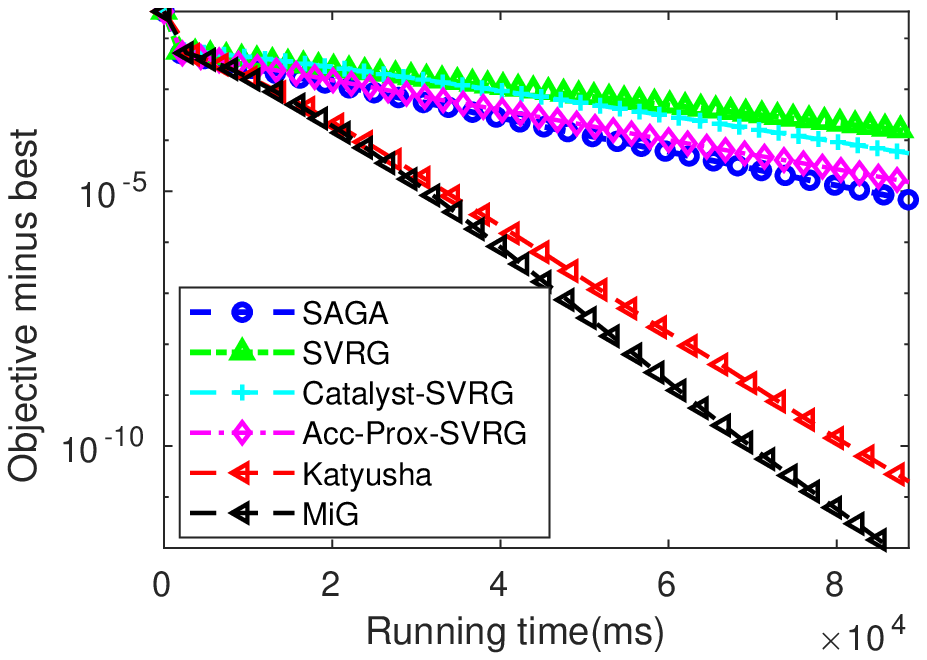}}
		\centerline{
			\includegraphics[width=\columnwidth / 2]{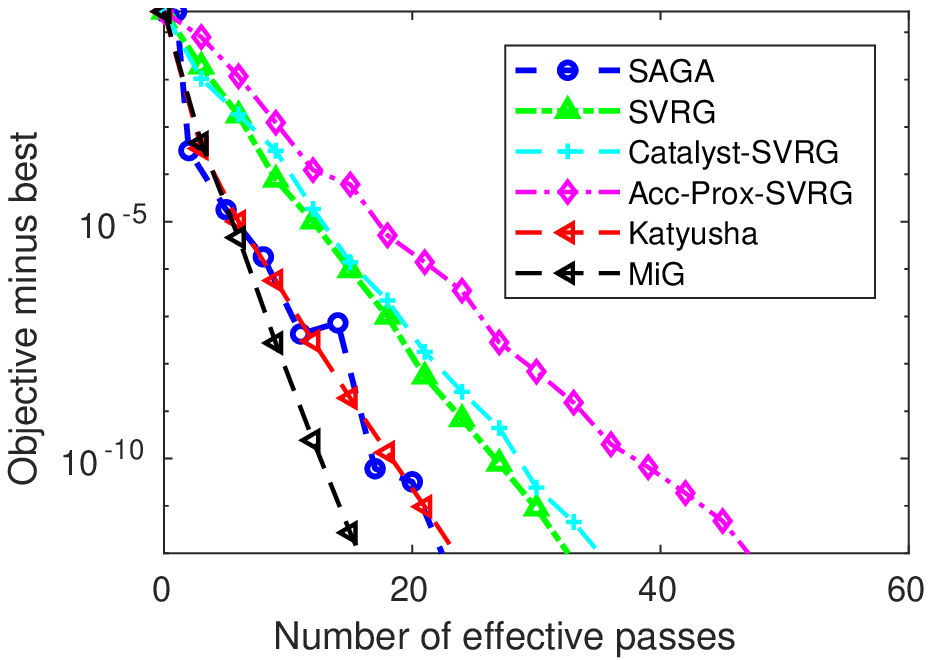}
			\includegraphics[width=\columnwidth / 2]{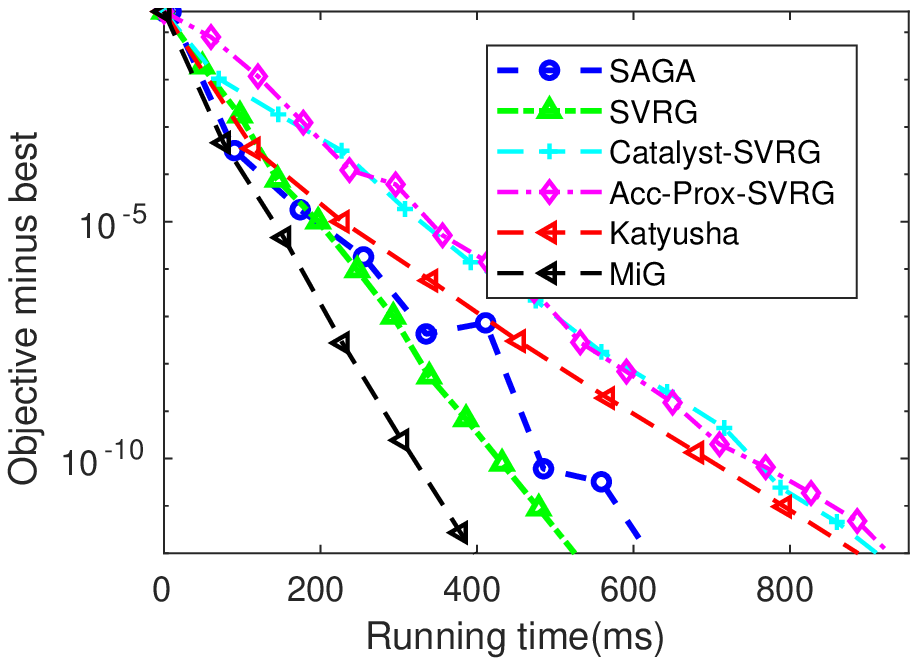}}
		\caption{First row: Theoretical evaluation of MiG and state-of-the-art algorithms for $\ell2$-logistic regression ($\lambda\!=\!10^{-8}$) on \textsf{covtype}. Second row: Practical evaluation of MiG and the state-of-the-art algorithms for ridge regression ($\lambda\!=\!10^{-4}$) on \textsf{a9a}.}
		\label{MiG_All}
	\end{center}
\end{figure}

For the case of $\frac{m}{\kappa}\!>\!\frac{3}{4}$ (see the second row in Figure~\ref{MiG_Katyusha}), we tuned all the parameters in Table~\ref{Moment_table}. From the parameter tuning of Acc-Prox-SVRG, we found that using a smaller momentum parameter $\beta$ yields a better performance, but still worse than the original SVRG. This result seems to indicate that the ``\textit{Nesterov's Momentum}" is not effective in this case. Katyusha yields a poor performance in this case because the parameter suggestion limits $\tau_1\!\leq\!\frac{1}{2}$. When tuning both $\tau_2$ and $\tau_1$, Katyusha performs much better, but with increasing difficulty of parameter tuning. MiG performs quite well with tuning only one parameter $\theta$, which further verifies its simplicity and efficiency.
\subsection{Comparison with state-of-the-art Algorithms}
We compare MiG with many state-of-the-art accelerated algorithms (e.g., Acc-Prox-SVRG \cite{nitanda:svrg}, Catalyst (based on SVRG) \cite{lin:vrsg}, and Katyusha \cite{zhu:Katyusha}) and non-accelerated algorithms (e.g., SVRG \cite{johnson:svrg} or Prox-SVRG \cite{xiao:prox-svrg} for non-smooth regularizer, and SAGA \cite{defazio:saga}), as shown in Figure~\ref{MiG_All}.

In order to give clear comparisons, we designed two different types of experiments. One is called ``theoretical evaluation'' with a relatively small $\lambda$\footnote{Note that we normalize data vectors to ensure a uniform $L$.}, where most of the parameter settings follow the corresponding theoretical recommendations\footnote{Except for Acc-Prox-SVRG and Catalyst, we carefully tuned the parameters for them, and the detailed parameter settings are given in the Supplementary Material~\ref{exp_parameter_settings}.} to justify the improvement of $\sqrt{\kappa n}$ convergence rate. Another is ``practical evaluation'' for a relatively large $\lambda$, where we carefully tuned the parameters for all the algorithms since in this condition, all the algorithms have similar convergence rates.

For a relatively large $\lambda$, the results (see the Supplementary Material~\ref{exp_parameter_settings} for more results) show that MiG performs consistently better than Katyusha in terms of both oracle calls and running time. In other words, we see that MiG achieves satisfactory performance in both conditions. Moreover, experimental results for non-strongly convex objectives are also given in the Supplementary Material~\ref{exp_parameter_settings}.

\begin{figure}[t]
	\begin{center}
		\centerline{
			\includegraphics[width=\columnwidth / 2]{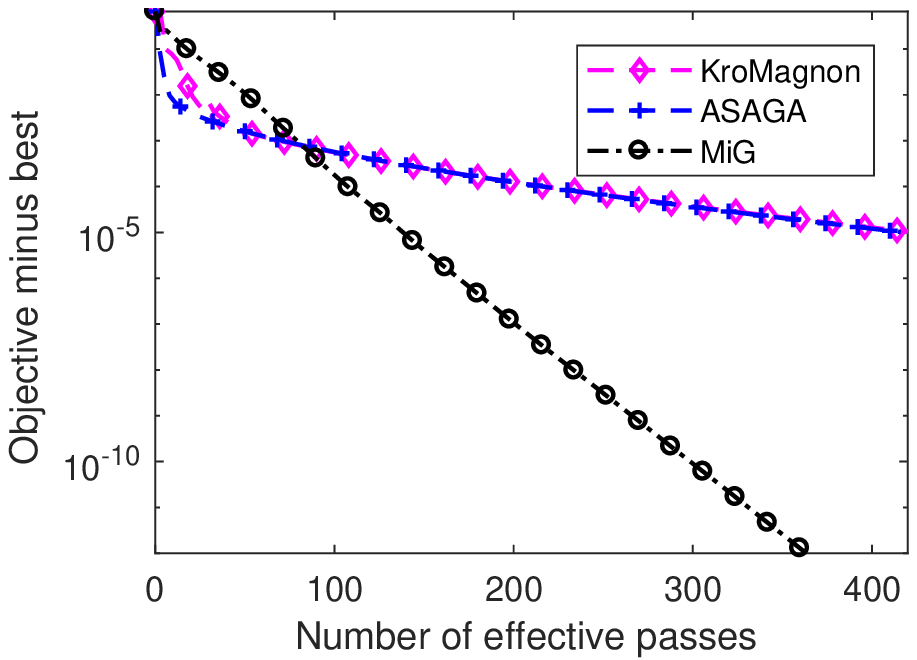}
			\includegraphics[width=\columnwidth / 2]{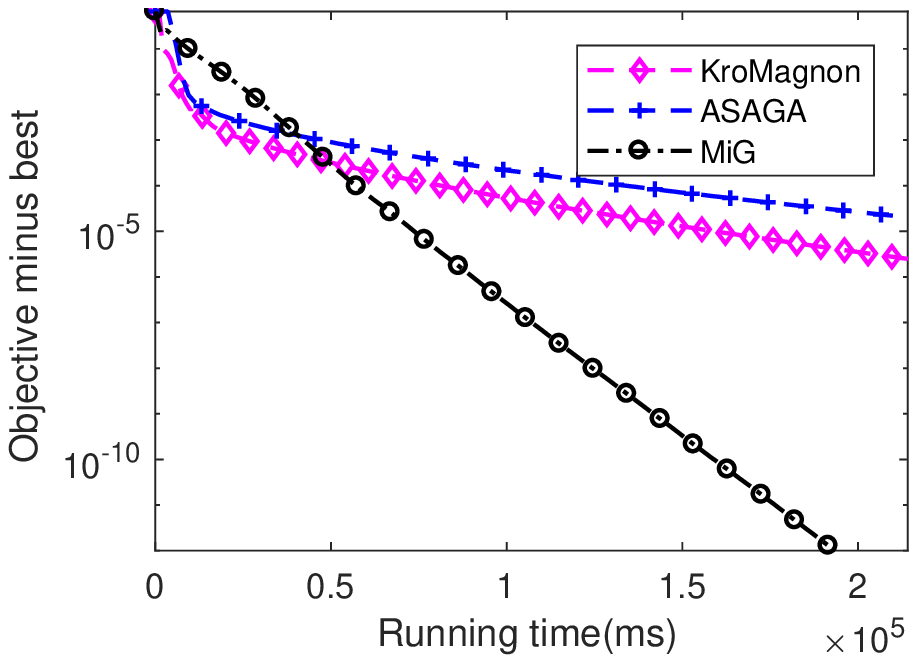}}

		\centerline{
			\includegraphics[width=\columnwidth / 2]{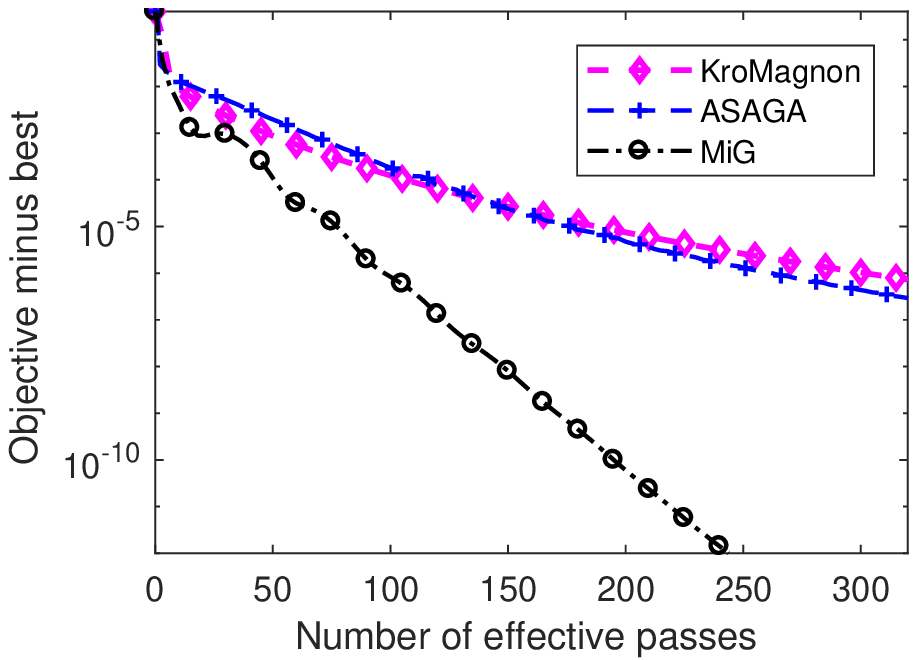}
			\includegraphics[width=\columnwidth / 2]{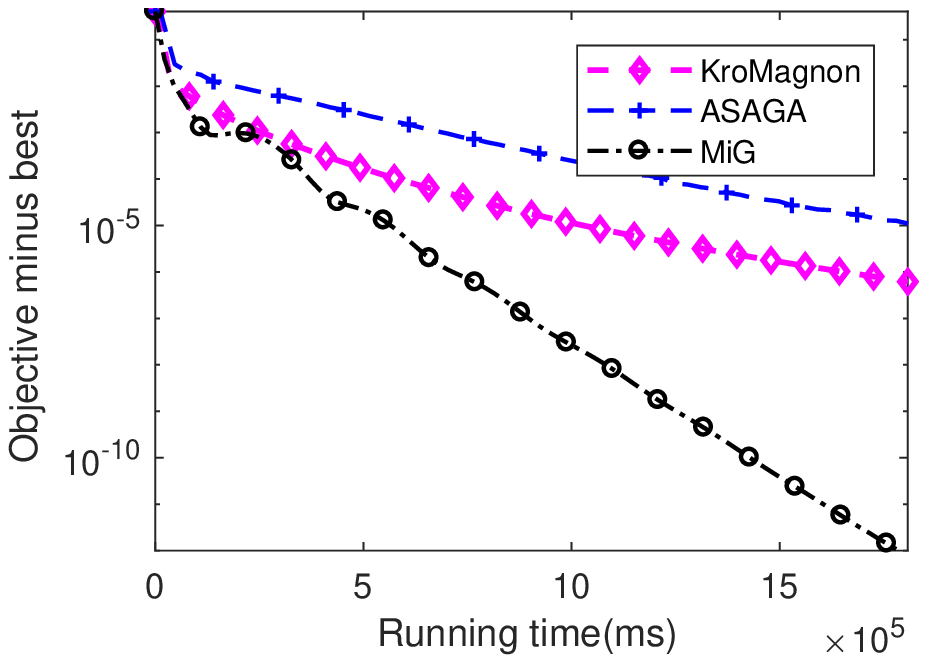}}
		\caption{Comparison of KroMagnon \cite{man:perturbed}, ASAGA \cite{leb:asaga}, and MiG with 16 threads. First row: \textsf{RCV1}, $\ell_2$-logistic regression with $\lambda\! =\! 10^{-9}$. Second row: \textsf{KDD2010}, $\ell_2$-logistic regression with $\lambda \!=\!10^{-10}$.}
		\label{Async_MiG}
	\end{center}
\end{figure}

\subsection{In Sparse and Asynchronous Settings}
\label{async_experiments}
To further stress the simplicity and implementability of MiG, we make some experiments to assess the performance of its asynchronous variant. We also compare MiG (i.e., Algorithm \ref{async_mig} with Option II\footnote{We omit the tricky restart scheme required in theory to examine the most practical variant.}) with KroMagnon \cite{man:perturbed} and ASAGA \cite{leb:asaga}.

Unlike in the serial dense case where we have strong theoretical guarantees, in these settings, we mainly focus on practical performance and stability. So we carefully tuned the parameter(s) for each algorithm to achieve a best-tuned performance (detailed setting and parameter tuning criteria is given in the Supplementary Material \ref{async_exp_parameter_settings}). We measure the performance on the two sparse datasets listed in Table~\ref{Table4}.


When comparing performance in terms of oracle calls, MiG significantly outperforms other algorithms, as shown in Figure~\ref{Async_MiG}. When considering running time, the difference is narrowed due to the high simplicity of KroMagnon (which only uses one atomic vector) compared with ASAGA (which uses atomic gradient table and atomic gradient average vector) and MiG (which only uses atomic ``\textit{fake average}'').

\begin{figure}[t]
	\begin{center}
		\centerline{
			\includegraphics[width=0.57\columnwidth]{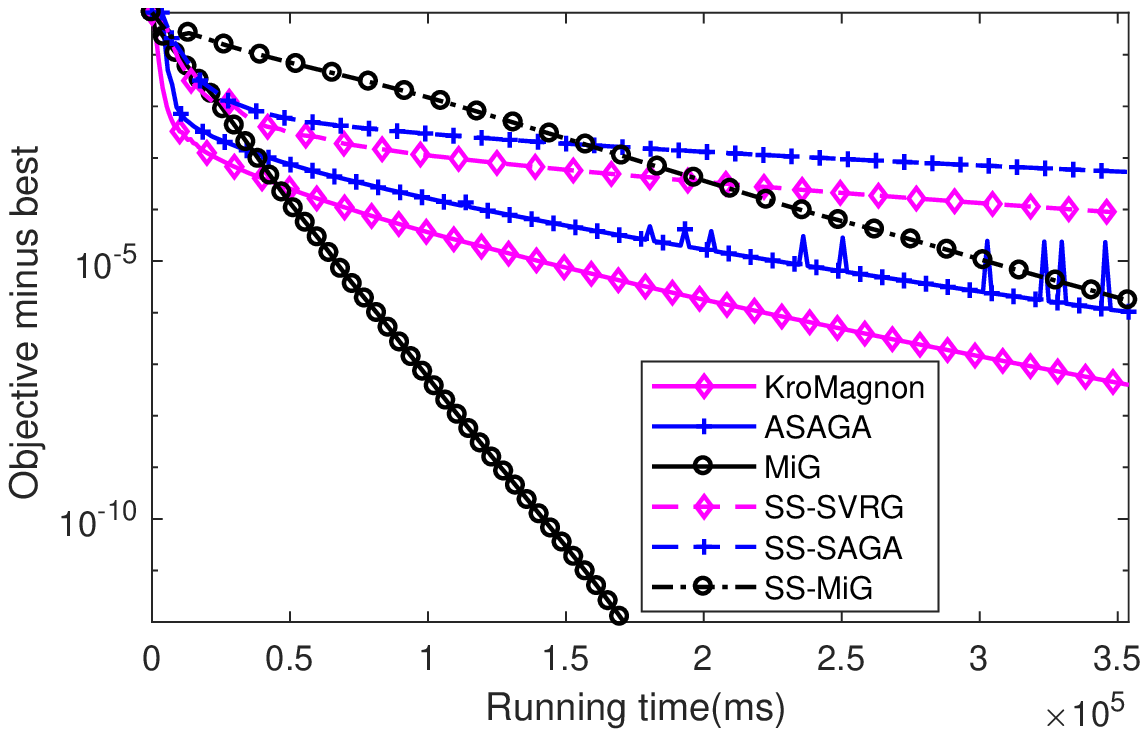}
			\includegraphics[width=0.40\columnwidth]{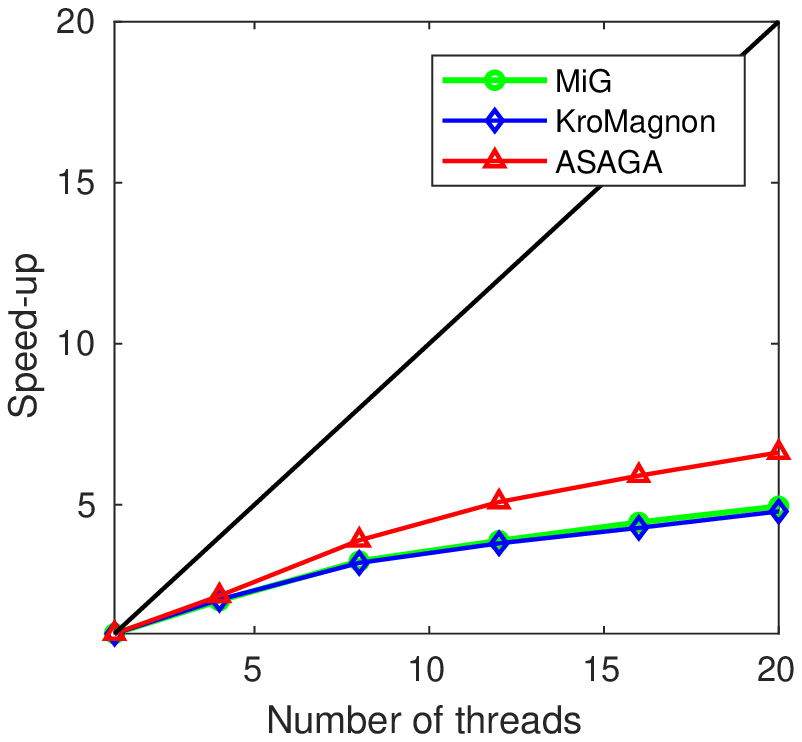}}
		\caption{Speed-up evaluation on \textsf{RCV1}. Left: Evaluation of sub-optimality in terms of running time for asynchronous versions (20 threads) and SS (Serial Sparse) versions. Right: Speed-up of achieving $10^{-5}$ sub-optimality in terms of the number of threads.}
		\label{speed_up}
	\end{center}
\end{figure}

We then examine the speed-up gained from more parallel threads on \textsf{RCV1}. We evaluate the improvement of using asynchronous variants (20 threads) and the speed-up ratio as a function of the number of threads as shown in Figure~\ref{speed_up}. For the latter evaluation, the running time is recorded when the algorithms achieve $10^{-5}$ sub-optimality. The speed-up ratio is calculated based on the running time of a single core.

\begin{table}[t]
	\caption{Summary of the two sparse data sets.}
	\vskip 0.1in
	\label{Table4}
	\begin{center}
		\begin{tabular}{cccc}
			\toprule
			Dataset & \# Data & \# Features & Density \\
			\midrule
			RCV1    & 697,641&47,236& 1.5 $\times$ $10^{-3}$ \\
			KDD2010 & 19,264,097&1,163,024 & $10^{-6}$\\
			\bottomrule
		\end{tabular}
	\end{center}
\end{table}

Since we used MiG with Option II for the above experiments which only has a theoretical analysis (with restart) in the serial case, we further designed an experiment to evaluate the effectiveness of $\theta$. The results in the Supplementary Material~\ref{effective_theta} indicate the effectiveness of our acceleration trick.

\section{Conclusion}\label{conclusions}

We proposed a simple stochastic variance reduction algorithm (MiG) with the best-known oracle complexities for stochastic first-order algorithms. These elegant results further reveal the mystery of acceleration tricks in stochastic first-order optimization. Moreover, the high simplicity of MiG allows us to derive the variants for the asynchronous and sparse settings, which shows its potential to be applied to other cases (e.g., online~\cite{borsos:online}, distributed \cite{lee:dsgd,lian:parallel}) as well as to tackle more complex problems (e.g., structured prediction). In general, our approach can be implemented to boost the speed of large-scale real-world optimization problems.

\section*{Acknowledgements}
We thank the reviewers for their valuable comments. This
work was supported in part by Grants (CUHK 14206715 \&
14222816) from the Hong Kong RGC, the Major Research Plan of the National Natural Science Foundation of China (Nos. 91438201 and 91438103), Project supported the Foundation for Innovative Research Groups of the National Natural Science Foundation of China (No. 61621005), the National Natural Science Foundation of China (Nos. U1701267, 61573267, 61502369 and 61473215), the Program for Cheung Kong Scholars and Innovative Research Team in University(No. IRT\_15R53) and the Fund for Foreign Scholars in University Research and Teaching Programs (the 111 Project) (No. B07048).


\bibliography{icml_2018}
\bibliographystyle{icml2018}


\newpage
\onecolumn
\icmltitlerunning{Supplementary Materials for ``A Simple Stochastic Variance Reduced Algorithm with Fast Convergence Rates''}
\icmltitle{Supplementary Materials for \\ ``A Simple Stochastic Variance Reduced Algorithm with Fast Convergence Rates''}

\renewcommand\thefigure{\arabic{figure}}  
\setcounter{figure}{0}   
\appendix
{\textbf{Notations.} We use $\mathbb{E}_{i_j}$ to denote that the expectation is taken with respect to the $j$th sample in one epoch, while $\mathbb{E}$ means taking expectation with respect to all randomness in one epoch. We use $O(\cdot)$ to denote computational complexity and $\mathcal{O}(\cdot)$ to denote oracle complexity. $x^*$ refers to the solution to Problem~\eqref{prob_def} (the proofs for Section~\ref{sequential}) or Problem~\eqref{prob_def2} (the proofs for Section~\ref{sparse_async}). $\mathcal{S}$ is the total number of epochs to be executed. Boldface number like $\mathbf{0}$ refers to a vector with all 0.

{\textbf{Note:} }In order to give a clean proof, we omit the superscripts for iterates in the same epoch $s$ as $x_j$ instead of $x^s_j$ unless otherwise specified.}

\section{Useful Lemmas}
\begin{variance}\label{variance}
	\textup{(Variance Bound)} Suppose each component function $f_i$ is $L$-smooth, let $\tilde{\nabla} = \pfij{y_{j-1}} - \pfij{\tilde{x}_{s-1}} + \pf{\tilde{x}_{s-1}}$, which is the approximate gradient used in \textup{MiG}. Then the following inequality holds:
	\[
		\Eij{\norm{\pf{y_{j-1}} - \tilde{\nabla}}^2} \leq 2L\big(f(\tilde{x}_{s-1}) - f(y_{j-1}) - \langle \pf{y_{j-1}}, \tilde{x}_{s-1} - y_{j-1} \rangle \big).
	\]
	\begin{proof}
		This lemma is identical to Lemma 3.4 in~\cite{zhu:Katyusha}, which provides a tighter upper bound on the gradient estimator variance than those in~\cite{johnson:svrg, xiao:prox-svrg}.
	\end{proof}
\end{variance}

\vspace{3mm}

\begin{three_point}\label{3_point}
	\textup{(3-points property)} Assume that $z^*$ is an optimal solution to the following problem,
	\[
		\min_{x} \frac{\tau}{2} \norm{x - z_0}^2 + \psi(x),
	\]
	where $\tau > 0$, and $\psi(\cdot)$ is a convex function (but possibly non-differentiable). Then for all $z \in \R^d$, there exists a vector $ \mathcal{G} \in \partial \psi(z^*)$ with
	\[
		\langle \mathcal{G}, z - z^* \rangle = \frac{\tau}{2} \norm{z^* - z_0}^2 - \frac{\tau}{2} \norm{z - z_0}^2 + \frac{\tau}{2} \norm{z - z^*}^2,
	\]
   where $\partial \psi(z^*)$ denotes the sub-differential of $\psi(\cdot)$ at $z^*$. If $\psi(\cdot)$ is differentiable, we can simply replace $\mathcal{G} \in \partial \psi(\cdot)$ with $\mathcal{G} = \nabla \psi(\cdot)$.
	\begin{proof}
		By the optimality of $z^*$, there exists a vector $\mathcal{G} \in \partial \psi(z^*)$ (or $\mathcal{G} = \nabla \psi(z^*)$ for differentiable $\psi(\cdot)$) satisfying
		\[
			\tau(z^* - z_0) + \mathcal{G} = \mathbf{0}.
		\]
		Thus for all $z\in \R^d$,
		\[
			\begin{aligned}
				0 & = \langle \tau(z^* - z_0) + \mathcal{G}, z^* - z  \rangle\\
				& = \tau\langle z^* - z_0, z^* - z \rangle + \langle \mathcal{G}, z^* - z \rangle\\
				&\meq{\star} \frac{\tau}{2} \norm{z^* - z_0}^2 - \frac{\tau}{2} \norm{z - z_0}^2 + \frac{\tau}{2} \norm{z - z^*}^2 + \langle \mathcal{G}, z^* - z \rangle,
			\end{aligned}
		\]
		where \mar{$\star$} uses the fact that $\langle a - b, a - c \rangle = \frac{1}{2}\norm{a-b}^2 - \frac{1}{2} \norm{b-c}^2 + \frac{1}{2} \norm{a-c}^2$.
	\end{proof}
\end{three_point}

\vspace{3mm}

\begin{sc_prox}\label{g_prox}
	If two vector $x_j$, $x_{j-1}\in \R^d$ satisfy $x_j = \argmin_{x} \lbrace \frac{1}{2\eta} \norm{x - x_{j-1}}^2 + \langle \tilde{\nabla}, x \rangle + g(x) \rbrace$ with a constant vector $\tilde{\nabla}$ and a general convex function $g(\cdot)$, then for all $u\in \R^d$, we have
	\[
		\langle \tilde{\nabla}, x_{j} - u \rangle \leq -\frac{1}{2\eta}\norm{x_{j-1} - x_{j}}^2 + \frac{1}{2\eta}\norm{x_{j-1} - u}^2 - \frac{1}{2\eta}\norm{x_{j}-u}^2 + g(u) - g(x_{j}).
	\]
	Moreover, if $g(\cdot)$ is $\sigma$-strongly convex, the above inequality becomes
	\[
		\langle \tilde{\nabla}, x_{j} - u \rangle \leq -\frac{1}{2\eta}\norm{x_{j-1} - x_{j}}^2 + \frac{1}{2\eta}\norm{x_{j-1} - u}^2 - \frac{1+\eta\sigma}{2\eta}\norm{x_{j}-u}^2 + g(u) - g(x_{j}).
	\]
	\begin{proof}
		Applying Lemma~\ref{3_point} with $z = u$, $z_0 = x_{j-1}$, $z^* = x_j$, $\tau = \frac{1}{\eta}$, $\psi(x) = \langle \tilde{\nabla}, x \rangle + g(x)$, there exists a vector $\mathcal{G} \in \partial g(x_j)$ (or $\mathcal{G} = \nabla g(x_j)$ for differentiable $g(\cdot)$) satisfying
		\[
			\langle \tilde{\nabla}, u - x_j \rangle +  \langle \mathcal{G}, u-x_j \rangle = \frac{1}{2\eta} \norm{x_{j-1} - x_{j}}^2 - \frac{1}{2\eta} \norm{x_{j-1} - u}^2 + \frac{1}{2\eta} \norm{x_{j} - u}^2.
		\]
		
		Using the convexity of $g(\cdot)$, we get $g(u) - g(x_j) \geq \langle \mathcal{G}, u - x_j \rangle$ by definition. After rearranging, we conclude that
		\[
		\begin{aligned}
			\langle \tilde{\nabla}, x_{j} - u \rangle \leq -\frac{1}{2\eta}\norm{x_{j-1} - x_{j}}^2 + \frac{1}{2\eta}\norm{x_{j-1} - u}^2 - \frac{1}{2\eta}\norm{x_{j}-u}^2 + g(u) - g(x_{j}).
		\end{aligned}
		\]
		
		If $g(\cdot)$ is further $\sigma$-strongly convex, we have $g(u) - g(x_j) \geq \langle \mathcal{G}, u - x_j \rangle + \frac{\sigma}{2} \norm{x_j - u}^2$ by~(\ref{s-convex}). Similarly, we can write
		\[
			\langle \tilde{\nabla}, x_{j} - u \rangle \leq -\frac{1}{2\eta}\norm{x_{j-1} - x_{j}}^2 + \frac{1}{2\eta}\norm{x_{j-1} - u}^2 - \frac{1+\eta\sigma}{2\eta}\norm{x_{j}-u}^2 + g(u) - g(x_{j}).
		\]
	\end{proof}
\end{sc_prox}

\section{Proofs for Section~\ref{sequential}}
\label{proofs_sec2}
\subsection{Proof of Theorem~\ref{theorem1}}
First, we add the following constraint on the parameters $\eta$ and $\theta$, which is crucial in the proof of Theorem~\ref{theorem1}:
\begin{equation} \label{constraint}
	L\theta + \frac{L\theta}{1 - \theta} \leq \frac{1}{\eta}\text{ , or equivalently } \eta \leq \frac{1 - \theta}{L\theta(2-\theta)}.
\end{equation}
We start with convexity of $f(\cdot)$ at $y_{j-1}$. By definition, for any vector $u\in \R^d$, we have
\begin{align}
	f(y_{j-1}) - f(u) &\leq \langle \pf{y_{j-1}}, \;y_{j-1} - u \rangle \nonumber\\
	&= \langle \pf{y_{j-1}}, \;y_{j-1} - x_{j-1} \rangle + \langle \pf{y_{j-1}},\; x_{j-1} - u \rangle  \nonumber\\
	&\meq{\star} \frac{1 - \theta}{\theta}\langle \pf{y_{j-1}}, \;\tilde{x}_{s-1} - y_{j-1} \rangle +  \langle \pf{y_{j-1}}, \;x_{j-1} - u \rangle \label{T1_convx},
\end{align}
where \mar{$\star$} follows from the fact that $y_{j-1} = \theta x_{j-1} + (1 - \theta) \tilde{x}_{s-1}$.

Then we further expand $\langle \pf{y_{j-1}}, \;x_{j-1} - u \rangle$ as
\begin{equation} \label{T1_exp1}
	\langle \pf{y_{j-1}}, \;x_{j-1} - u \rangle = \langle \pf{y_{j-1}} - \tilde{\nabla}, \;x_{j-1} - u \rangle + \langle \tilde{\nabla}, \;x_{j-1} - x_j \rangle + \langle \tilde{\nabla},\; x_j - u \rangle .
\end{equation}
Using $L$-smooth~(\ref{l-smooth}) of $f(\cdot)$ at $(y_{j}, y_{j-1})$, we get
\[
\begin{aligned}
	f(y_j)-f(y_{j-1}) &\leq \langle \pf{y_{j-1}}, \;y_j - y_{j-1} \rangle + \frac{L}{2} \norm{y_j - y_{j-1}}^2 \\
	&\meq{\star} \theta \langle \pf{y_{j-1}}, \;x_j - x_{j-1} \rangle + \frac{L\theta^2}{2} \norm{x_j - x_{j-1}}^2 \\
	&= \theta \big[\langle \pf{y_{j-1}} - \tilde{\nabla}, \;x_j - x_{j-1} \rangle + \langle \tilde{\nabla}, \;x_j - x_{j-1} \rangle\big] + \frac{L\theta^2}{2} \norm{x_j - x_{j-1}}^2, \\
	\langle \tilde{\nabla}, x_{j-1} - x_{j} \rangle &\leq \frac{1}{\theta}\big(f(y_{j-1}) - f(y_{j})\big) + \langle \pf{y_{j-1}} - \tilde{\nabla}, x_j - x_{j-1} \rangle + \frac{L\theta}{2} \norm{x_j - x_{j-1}}^2,
\end{aligned}
\]
where \mar{$\star$} uses the definition of $y_{j-1}$.

After plugging in the constraint~(\ref{constraint}), we have
\begin{equation} \label{T1_Lsmooth}
	\langle \tilde{\nabla}, \;x_{j-1} - x_{j} \rangle \leq \frac{1}{\theta}\big(f(y_{j-1}) - f(y_{j})\big) + \langle \pf{y_{j-1}} - \tilde{\nabla}, \;x_j - x_{j-1} \rangle + \frac{1}{2\eta} \norm{x_j - x_{j-1}}^2 - \frac{L\theta}{2(1-\theta)}\norm{x_j - x_{j-1}}^2.
\end{equation}

Then we are ready to combine~(\ref{T1_convx}),~(\ref{T1_exp1}),~(\ref{T1_Lsmooth}), as well as Lemma~\ref{g_prox} (here $g(x)$ is $\sigma$-strongly convex by Assumption~\ref{assump1}), which gives
\begin{align}
	f(y_{j-1}) - f(u)
	&\leq \frac{1 - \theta}{\theta}\langle \pf{y_{j-1}}, \tilde{x}_{s-1} - y_{j-1} \rangle +  \langle \pf{y_{j-1}} - \tilde{\nabla}, x_{j} - u \rangle + \frac{1}{\theta} (f(y_{j-1}) - f(y_j)) \nonumber\\
	& \ \ \ \ - \frac{L\theta}{2(1-\theta)}\norm{x_j - x_{j-1}}^2 + \frac{1}{2\eta}\norm{x_{j-1} - u}^2 - \frac{1+\eta\sigma}{2\eta}\norm{x_{j}-u}^2 + g(u) - g(x_{j}). \nonumber
\end{align}

After taking expectation with respect to the sample $i_j$, we obtain
\[
\begin{aligned}
	f(y_{j-1}) - f(u)
	&\mleq{a} \frac{1 - \theta}{\theta}\langle \pf{y_{j-1}}, \tilde{x}_{s-1} - y_{j-1} \rangle +  \Eij{\langle \pf{y_{j-1}} - \tilde{\nabla}, x_{j} - x_{j-1} \rangle} + \frac{1}{\theta} (f(y_{j-1}) -\Eij{f(y_j)}) \\
	& \ \ \ \ - \frac{L\theta}{2(1-\theta)}\Eij{\norm{x_j - x_{j-1}}^2} + \frac{1}{2\eta}\norm{x_{j-1} - u}^2 - \frac{1+\eta\sigma}{2\eta}\Eij{\norm{x_{j}-u}^2} + g(u) - \Eij{g(x_{j})} \\
	&\mleq{b} \frac{1 - \theta}{\theta}\langle \pf{y_{j-1}}, \tilde{x}_{s-1} - y_{j-1} \rangle + \frac{1}{2\beta}\Eij{\norm{\pf{y_{j-1}} - \tilde{\nabla}}^2}+\frac{\beta}{2}\Eij{\norm{x_{j} - x_{j-1}}^2} \\
	& \ \ \ \  + \frac{1}{\theta} (f(y_{j-1}) -\Eij{f(y_j)})- \frac{L\theta}{2(1-\theta)}\Eij{\norm{x_j - x_{j-1}}^2} + \frac{1}{2\eta}\norm{x_{j-1} - u}^2 \\
	& \ \ \ \ - \frac{1+\eta\sigma}{2\eta}\Eij{\norm{x_{j}-u}^2} + g(u) - \Eij{g(x_{j})},
\end{aligned}
\]
where \mar{$a$} holds due to the unbiasedness of the gradient estimator $\Eij{\pf{y_{j-1}} - \tilde{\nabla}} = \mathbf{0}$, and \mar{$b$} uses the Young's inequality to expand $\Eij{\langle \pf{y_{j-1}} - \tilde{\nabla}, x_{j} - x_{j-1} \rangle}$ with the parameter $\beta > 0$.

Applying Lemma~\ref{variance} to bound the variance term $\Eij{\norm{\pf{y_{j-1}} - \tilde{\nabla}}^2}$, we get
\[
\begin{aligned}
 f(y_{j-1}) - f(u)
	&\leq \frac{1 - \theta}{\theta}\langle \pf{y_{j-1}}, \tilde{x}_{s-1} - y_{j-1} \rangle +  \frac{L}{\beta}\big(f(\tilde{x}_{s-1}) - f(y_{j-1}) - \langle \pf{y_{j-1}}, \tilde{x}_{s-1} - y_{j-1} \rangle \big)  \\
	& \ \ \ \ +\frac{\beta}{2}\Eij{\norm{x_{j} - x_{j-1}}^2} + \frac{1}{\theta} (f(y_{j-1}) -\Eij{f(y_j)})- \frac{L\theta}{2(1-\theta)}\Eij{\norm{x_j - x_{j-1}}^2} + \frac{1}{2\eta}\norm{x_{j-1} - u}^2 \\
	& \ \ \ \ - \frac{1+\eta\sigma}{2\eta}\Eij{\norm{x_{j}-u}^2} + g(u) - \Eij{g(x_{j})}.
\end{aligned}
\]

Let $\beta = \frac{L\theta}{1-\theta} > 0$, by rearranging the above inequality, we obtain
\begin{align}
	0 &\leq \frac{1 - \theta}{\theta}f(\tilde{x}_{s-1}) - \frac{1}{\theta}\Eij{f(y_j)} + F(u) - \Eij{g(x_{j})} + \frac{1}{2\eta}\norm{x_{j-1} - u}^2  - \frac{1+\eta\sigma}{2\eta} \Eij{\norm{x_{j}-u}^2} \nonumber \\
	&\mleq{\star} \frac{1 - \theta}{\theta}F(\tilde{x}_{s-1}) - \frac{1}{\theta}\Eij{F(y_j)} + F(u)+ \frac{1}{2\eta}\norm{x_{j-1} - u}^2  - \frac{1+\eta\sigma}{2\eta} \Eij{\norm{x_{j}-u}^2}, \nonumber \\
	&\frac{1}{\theta}\big(\Eij{F(y_j)} - F(u)\big) \leq \frac{1 - \theta}{\theta}\big(F(\tilde{x}_{s-1}) - F(u)\big) + \frac{1}{2\eta}\norm{x_{j-1} - u}^2  - \frac{1+\eta\sigma}{2\eta} \Eij{\norm{x_{j}-u}^2},\label{T1_scope_y}
\end{align}
where \mar{$\star$} follows from the Jensen's inequality and the definition of $y_{j-1}$, which leads to $-g(x_j) \leq \frac{1-\theta}{\theta}g(\tilde{x}_{s-1}) - \frac{1}{\theta}g(y_j)$.

Let $u = x^*$, using our choice of $\omega = 1 + \eta\sigma$ to sum~(\ref{T1_scope_y}) over $j=1\ldots m$ with increasing weight $\omega^{j-1}$. After taking expectation with respect to all randomness in this epoch, we have
\[
\begin{aligned}
	\frac{1}{\theta}\sum_{j = 0}^{m-1}\omega^j\big(\E{F(y_{j+1})} - F(x^*)\big) &+ \frac{\omega^m}{2\eta}\E{\norm{x_{m}-x^*}^2} \leq \frac{1 - \theta}{\theta}\sum_{j = 0}^{m-1}\omega^j\big(F(\tilde{x}_{s-1}) - F(x^*)\big) + \frac{1}{2\eta}\norm{x_{0} - x^*}^2.
\end{aligned}
\]

Using the Jensen's inequality and $\tilde{x}_s = \theta{\big(\sum_{j=0}^{m-1} \omega^{j}\big)}^{-1} \sum_{j = 0}^{m-1}{\omega^{j} x_{j+1}} + (1 - \theta) \tilde{x}_{s-1} = (\sum_{j = 0}^{m-1}\omega^j)^{-1} \sum_{j = 0}^{m-1}{\omega^jy_{j+1}}$, we have
\begin{equation}\label{T1_contraction}
	\frac{1}{\theta}\sum_{j = 0}^{m-1}\omega^j\big(\E{F(\tilde{x}_s)} - F(x^*)\big) + \frac{\omega^m}{2\eta}\E{\norm{x_{m}-x^*}^2} \leq \frac{1 - \theta}{\theta}\sum_{j = 0}^{m-1}\omega^j\big(F(\tilde{x}_{s-1}) - F(x^*)\big) + \frac{1}{2\eta}\norm{x_{0} - x^*}^2.
\end{equation}

\textbf{(I)} Consider the first case in Theorem~\ref{theorem1} with $\frac{m}{\kappa} \leq \frac{3}{4}$, we set $\eta = \sqrt{\frac{1}{3\sigma m L}}$, $\theta = \sqrt{\frac{m}{3\kappa}} \leq \frac{1}{2}$, and $m = \Theta(n)$.

First, we evaluate the crucial constraint~(\ref{constraint}). By substituting in our parameter settings, the constraint becomes
\[
	L\theta + \frac{L\theta}{1 - \theta} \leq \frac{1}{\eta} \rightarrow \sqrt{\frac{m}{\kappa}} \leq \frac{\sqrt{3}}{2}.
\]
Thus the constraint is satisfied by meeting the case assumption.

Then we focus on $(1-\theta)\omega^m$, observed that
\[
	(1-\theta)\omega^m = (1 - \sqrt{\frac{m}{3\kappa}})\cdot (1 + \sqrt{\frac{1}{3m\kappa}})^m.
\]
Let $\zeta = \sqrt{\frac{m}{\kappa}}$, $\zeta \in (0, \frac{\sqrt{3}}{2}]$, we can denote
\[
	\phi(\zeta) = (1-\frac{\sqrt{3}}{3}\zeta)\cdot(1+\frac{\sqrt{3}}{3}\cdot \frac{\zeta}{m})^m
\]
as a function of $\zeta$.

By taking derivative with respect to $\zeta$, we find that $\phi(\zeta)$ is monotonically decreasing on $[0, \frac{\sqrt{3}}{2}]$ for any $m >0$, which means
\[
	(1-\theta)\omega^m \leq \max_{\zeta \in (0, \frac{\sqrt{3}}{2}]} {\phi(\zeta)} \leq \phi(0) = 1.
\]
Thus we have $\frac{1}{\theta} \geq \frac{1-\theta}{\theta}\omega^m$. By using this inequality in~(\ref{T1_contraction}), we get
\[
\begin{aligned}
	& \ \ \ \ \frac{1-\theta}{\theta}\sum_{j = 0}^{m-1}\omega^j\big(\E{F(\tilde{x}_s)} - F(x^*)\big) + \frac{1}{2\eta}\E{\norm{x_{m}-x^*}^2}\\
	&\leq \omega^{-m}\cdot\Big(\frac{1 - \theta}{\theta}\sum_{j = 0}^{m-1}\omega^j\big(F(\tilde{x}_{s-1}) - F(x^*)\big) + \frac{1}{2\eta}\norm{x_{0} - x^*}^2\Big).
\end{aligned}
\]
Dividing both sides of the above inequality by $\frac{1-\theta}{\theta}\sum_{j = 0}^{m-1}\omega^j$, we get
\[
\begin{aligned}
	& \ \ \ \ \big(\E{F(\tilde{x}_s)} - F(x^*)\big) + \frac{\theta}{2\eta(1-\theta)\sum_{j = 0}^{m-1}\omega^j}\E{\norm{x_{m}-x^*}^2}\\
	&\leq \omega^{-m}\cdot\Big(\big(F(\tilde{x}_{s-1}) - F(x^*)\big) + \frac{\theta}{2\eta(1-\theta)\sum_{j = 0}^{m-1}\omega^j}\norm{x_{0} - x^*}^2\Big).
\end{aligned}
\]
Summing the above inequality over $s = 1\ldots \mathcal{S}$, we get
\[
\begin{aligned}
	& \ \ \ \ \big(\E{F(\tilde{x}_{\mathcal{S}})} - F(x^*)\big) + \frac{\theta}{2\eta(1-\theta)\sum_{j = 0}^{m-1}\omega^j}\E{\norm{x^{\mathcal{S}}_{m}-x^*}^2}\\
	&\leq \omega^{-\mathcal{S}m}\cdot\Big(\big(F(\tilde{x}_0) - F(x^*)\big) + \frac{\theta}{2\eta(1-\theta)\sum_{j = 0}^{m-1}\omega^j}\norm{x^1_{0} - x^*}^2\Big).
\end{aligned}
\]
Notice that in order to prevent confusion, we mark iterates with epoch number, such as $x^{\mathcal{S}}_m$ represent the last iterate in epoch $\mathcal{S}$.

Using the fact that $\sum_{j = 0}^{m-1}\omega^j \geq m$, we have
\[
\begin{aligned}
	\big(\E{F(\tilde{x}_{\mathcal{S}})} - F(x^*)\big) \leq \omega^{-\mathcal{S}m}\cdot\Big(\big(F(\tilde{x}_0) - F(x^*)\big) + \frac{\theta}{2\eta(1-\theta)m}\norm{x^1_{0} - x^*}^2\Big).
\end{aligned}
\]
Using the $\sigma$-strongly convexity of $F(\cdot)$ to bound $\norm{x^1_0 - x^*}^2$, which is $\norm{x^1_0 - x^*}^2 \leq \frac{2}{\sigma} \big(F(x^1_0) - F(x^*)\big)$, we obtain
\[
\begin{aligned}
	\E{F(\tilde{x}_\mathcal{S}) - F(x^*)} \leq (1 + \eta\sigma)^{-\mathcal{S}m} \cdot \Big(1 + \frac{\theta}{\eta(1-\theta)m\sigma}\Big)\cdot \big(F(\tilde{x}_0) - F(x^*)\big).
\end{aligned}
\]
Note that $\tilde{x}_0 = x^1_0 = x_0$.

By substituting with our parameters setting, we get
\[
\begin{aligned}
	\E{F(\tilde{x}_\mathcal{S}) - F(x^*)} &\mleq{\star} \big(O(1 + \sqrt{\frac{1}{3n\kappa}})\big)^{-\mathcal{S}m} \cdot O\Big(1 + 2\theta\sqrt{\frac{\kappa}{n}}\Big)\cdot \big(F(\tilde{x}_0) - F(x^*)\big) \\
	&\leq \big(O(1 + \sqrt{\frac{1}{3n\kappa}})\big)^{-\mathcal{S}m}\cdot O\big(F(\tilde{x}_0) - F(x^*)\big),
\end{aligned}
\]
where \mar{$\star$} holds due to the fact that $\theta \leq \frac{1}{2}$.

The above result implies that the oracle complexity in the case $\frac{m}{\kappa} \leq \frac{3}{4}$ to achieve an $\epsilon$-additive error is
$
	\mathcal{O}\Big(\sqrt{\kappa n}\log{\frac{F(\tilde{x}_0) - F(x^*)}{\epsilon }}\Big)
$.

\textbf{(II)} For another case with $\frac{m}{\kappa} > \frac{3}{4}$, we set $\eta = \frac{2}{3L}$,  $\theta = \frac{1}{2}$, and $m = \Theta(n)$.

Again, we evaluate the constraint~(\ref{constraint}) first. By substituting the parameter setting, the constraint becomes
\[
	L\theta + \frac{L\theta}{1 - \theta} \leq \frac{1}{\eta} \rightarrow \eta \leq \frac{2}{3L}.
\]
Thus the constraint is satisfied by our parameter choice.

Substituting the parameter setting into~(\ref{T1_contraction}), we get
\[
	2\sum_{j = 0}^{m-1}\omega^j\big(\E{F(\tilde{x}_s)} - F(x^*)\big) + \frac{3L\omega^m}{4}\E{\norm{x_{m}-x^*}^2} \leq \sum_{j = 0}^{m-1}\omega^j\big(F(\tilde{x}_{s-1}) - F(x^*)\big) + \frac{3L}{4}\norm{x_{0} - x^*}^2  .
\]
Notice that based on the Bernoulli's inequality, $\omega^m = (1 + \frac{2}{3\kappa})^m \geq 1 + \frac{2m}{3\kappa} \geq \frac{3}{2}$, which leads to
\[
\begin{aligned}
	\frac{3}{2}\sum_{j = 0}^{m-1}\omega^j\big(\E{F(\tilde{x}_s)} - F(x^*)\big) &+ \frac{9L}{8}\E{\norm{x_{m}-x^*}^2} \leq \sum_{j = 0}^{m-1}\omega^j\big(F(\tilde{x}_{s-1}) - F(x^*)\big) + \frac{3L}{4}\norm{x_{0} - x^*}^2,  \\
	\sum_{j = 0}^{m-1}\omega^j\big(\E{F(\tilde{x}_s)} - F(x^*)\big) &+ \frac{3L}{4}\E{\norm{x_{m}-x^*}^2} \leq \frac{2}{3}\cdot\big(\sum_{j = 0}^{m-1}\omega^j\big(F(\tilde{x}_{s-1}) - F(x^*)\big) + \frac{3L}{4}\norm{x_{0} - x^*}^2\big).
\end{aligned}
\]
Again, by telescoping the above inequality from $s = 1\ldots \mathcal{S}$, we get
\[
\begin{aligned}
	\sum_{j = 0}^{m-1}\omega^j\big(\E{F(\tilde{x}_{\mathcal{S}})} - F(x^*)\big) &+ \frac{3L}{4}\E{\norm{x^{\mathcal{S}}_{m}-x^*}^2} \leq \big(\frac{2}{3}\big)^{\mathcal{S}}\cdot\big(\sum_{j = 0}^{m-1}\omega^j\big(F(\tilde{x}_0) - F(x^*)\big) + \frac{3L}{4}\norm{x^1_{0} - x^*}^2\big).
\end{aligned}
\]
Since $\sum_{j=0}^{m-1} \omega^j \geq m$, the above inequality can be rewritten as follows:
\[
\begin{aligned}
	\big(\E{F(\tilde{x}_{\mathcal{S}})} - F(x^*)\big) &\mleq{\star} \big(\frac{2}{3}\big)^{\mathcal{S}}\cdot\big(1 + \frac{3\kappa}{2m}\big)\cdot\big(F(\tilde{x}_0) - F(x^*)\big) \\
	&\leq \big(\frac{2}{3}\big)^{\mathcal{S}}\cdot O\big(F(\tilde{x}_0) - F(x^*)\big),
\end{aligned}
\]
where \mar{$\star$} uses the $\sigma$-strongly convexity of $F(\cdot)$, that is, $\norm{x^1_0 - x^*}^2 \leq \frac{2}{\sigma} \big(F(x^1_0) - F(x^*)\big)$.

This result implies that the oracle complexity in this case is $\bigO{n\log{\frac{F(\tilde{x}_0) - F(x^*)}{\epsilon}}}$.

\subsubsection{Detailed comparison between MiG and Katyusha}
\label{detailed_comp_with_katyu}
As mentioned in Section~\ref{comparisons}, MiG is a special case of Katyusha. However, it is a non-trivial special case and there are some significant differences from the theoretical perspective between them for the ill-conditioned problems: 

The intuition of Katyusha is that Katyusha Momentum is used to further reduce the variance of the iterates so as to make Nesterov's Momentum effective, which can be clearly seen from the parameter choice: the author fixed Katyusha Momentum (for $\tilde{x}$, in the notation of original work) as $\tau_2=1/2$, and chose Nesterov's Momentum (for $y$ and $z$) with ($1/2 - \tau_1$, $\tau_1$). This idea is similar to Acc-Prox-SVRG~\cite{nitanda:svrg}, which uses sufficiently large mini-batch to make the Nesterov's Momentum effective. Thus, the acceleration comes from Nesterov's Momentum, which is common in previous work (e.g. Catalyst).

In contrast, MiG uses only Katyusha Momentum (negative momentum) to achieve acceleration (i.e., choosing $\tau_2$ in a dynamic way), which is somehow counter-intuitive. To the best of our knowledge, this is the first work to yield an acceleration without using Nesterov's Momentum (which is at the heart of all existing accelerated first-order methods, as stated in~\cite{zhu:Katyusha}).

\subsection{Proof of Theorem~\ref{nsconvex_rate}}

Again, we first impose the following constraint on $\eta$ and $\theta$:
\begin{equation} \label{T2_cons}
	L\theta + \frac{L\theta}{1 - \theta} \leq \frac{1}{\eta}\text{ , or equivalently } \eta \leq \frac{1 - \theta}{L\theta(2-\theta)}.
\end{equation}
We start with convexity of $f(\cdot)$ at $y_{j-1}$. By definition, for any vector $u\in \R^d$, we have
\begin{equation}\label{T2_convx}
	\begin{aligned}
		f(y_{j-1}) - f(u) &\leq \langle \pf{y_{j-1}}, y_{j-1} - u \rangle \\
		&\meq{\star} \frac{1 - \theta}{\theta}\langle \pf{y_{j-1}}, \tilde{x}_{s-1} - y_{j-1} \rangle +  \langle \pf{y_{j-1}}, x_{j-1} - u \rangle,
	\end{aligned}
\end{equation}
where \mar{$\star$} follows from the fact that $y_{j-1} = \theta x_{j-1} + (1 - \theta) \tilde{x}_{s-1}$.

Then we further expand $\langle \pf{y_{j-1}}, x_{j-1} - u \rangle$ as
\begin{equation} \label{T2_exp1}
	\langle \pf{y_{j-1}}, x_{j-1} - u \rangle = \langle \pf{y_{j-1}} - \tilde{\nabla}, x_{j-1} - u \rangle + \langle \tilde{\nabla}, x_{j-1} - x_j \rangle + \langle \tilde{\nabla}, x_j - u \rangle.
\end{equation}
Using $L$-smooth~(\ref{l-smooth}) of $f(\cdot)$ at $(y_{j}, y_{j-1})$, we get
\[
\begin{aligned}
	f(y_j)-f(y_{j-1}) &\leq \langle \pf{y_{j-1}}, y_j - y_{j-1} \rangle + \frac{L}{2} \norm{y_j - y_{j-1}}^2 \\
	&= \theta \big[\langle \pf{y_{j-1}} - \tilde{\nabla}, x_j - x_{j-1} \rangle + \langle \tilde{\nabla}, x_j - x_{j-1} \rangle\big] + \frac{L\theta^2}{2} \norm{x_j - x_{j-1}}^2, \\
	\langle \tilde{\nabla}, x_{j-1} - x_{j} \rangle &\leq \frac{1}{\theta}\big(f(y_{j-1}) - f(y_{j})\big) + \langle \pf{y_{j-1}} - \tilde{\nabla}, x_j - x_{j-1} \rangle + \frac{L\theta}{2} \norm{x_j - x_{j-1}}^2.
\end{aligned}
\]
Using the constraint~(\ref{T2_cons}), we have
\begin{equation} \label{T2_Lsmooth}
	\langle \tilde{\nabla}, x_{j-1} - x_{j} \rangle \leq \frac{1}{\theta}\big(f(y_{j-1}) - f(y_{j})\big) + \langle \pf{y_{j-1}} - \tilde{\nabla}, x_j - x_{j-1} \rangle + \frac{1}{2\eta} \norm{x_j - x_{j-1}}^2 - \frac{L\theta}{2(1-\theta)}\norm{x_j - x_{j-1}}^2.
\end{equation}
Then we can combine~(\ref{T2_convx}),~(\ref{T2_exp1}),~(\ref{T2_Lsmooth}), as well as Lemma~\ref{g_prox} (with $g(x)$ convex), which leads to
\begin{align}
	f(y_{j-1}) - f(u)&\leq \frac{1 - \theta}{\theta}\langle \pf{y_{j-1}}, \tilde{x}_{s-1} - y_{j-1} \rangle +  \langle \pf{y_{j-1}} - \tilde{\nabla}, x_{j} - u \rangle + \frac{1}{\theta} (f(y_{j-1}) - f(y_j)) \nonumber\\
	& \ \ \ \ - \frac{L\theta}{2(1-\theta)}\norm{x_j - x_{j-1}}^2 + \frac{1}{2\eta}\norm{x_{j-1} - u}^2 - \frac{1}{2\eta}\norm{x_{j}-u}^2 + g(u) - g(x_{j}). \nonumber
\end{align}

After taking expectation with respect to the sample $i_j$, we get
\[
\begin{aligned}
	f(y_{j-1}) - f(u)&\leq \frac{1 - \theta}{\theta}\langle \pf{y_{j-1}}, \tilde{x}_{s-1} - y_{j-1} \rangle + \frac{1}{2\beta} \Eij{\norm{\pf{y_{j-1}} - \tilde{\nabla}}^2}+ \frac{\beta}{2}\Eij{\norm{x_{j} - x_{j-1}}^2} \\
	& \ \ \ \ + \frac{1}{\theta} (f(y_{j-1}) - \Eij{f(y_j)}) - \frac{L\theta}{2(1-\theta)}\Eij{\norm{x_j - x_{j-1}}^2} + \frac{1}{2\eta}\norm{x_{j-1} - u}^2\\
	&\ \ \ \  - \frac{1}{2\eta}\Eij{\norm{x_{j}-u}^2} + g(u) - \Eij{g(x_{j})},
\end{aligned}
\]
which the inequality uses the unbiasedness of the gradient estimator and the Young's inequality with the parameter $\beta > 0$.

Using Lemma~\ref{variance} to bound the variance term and choosing $\beta = \frac{L\theta}{1-\theta}$, the above inequality becomes
\[
	 \frac{1}{\theta} \Eij{f(y_j)} - F(u)\leq \frac{1-\theta}{\theta}f(\tilde{x}_{s-1}) + \frac{1}{2\eta}\norm{x_{j-1} - u}^2- \frac{1}{2\eta}\Eij{\norm{x_{j}-u}^2}- \Eij{g(x_{j})}.
\]
Let $u = x^*$ and using the fact that $-g(x_j) \leq \frac{1-\theta}{\theta}g(\tilde{x}_{s-1}) - \frac{1}{\theta}g(y_j)$, we get
\[
	\frac{1}{\theta} \Eij{F(y_j) - F(x^*)}\leq \frac{1-\theta}{\theta}\big(F(\tilde{x}_{s-1}) - F(x^*)\big) + \frac{1}{2\eta}\norm{x_{j-1} - x^*}^2- \frac{1}{2\eta}\Eij{\norm{x_{j}-x^*}^2}.
\]
Summing the above inequality over $j=1\ldots m$ and taking expectation with respect to all randomness in this epoch, we obtain
\begin{align}
	\frac{1}{\theta} \E{\frac{1}{m}\sum_{j=1}^m F(y_j) - F(x^*)} &\mleq{a} \frac{1-\theta}{\theta}\big(F(\tilde{x}_{s-1}) - F(x^*)\big) + \frac{2\theta L}{m}\norm{x_{0} - x^*}^2- \frac{2\theta L}{m}\E{\norm{x_{m}-x^*}^2}, \nonumber\\
	\frac{1}{\theta^2} \E{ F(\tilde{x}_s) - F(x^*)}&\mleq{b} \frac{1-\theta}{\theta^2}\big(F(\tilde{x}_{s-1}) - F(x^*)\big) + \frac{2L}{m}\norm{x_{0} - x^*}^2- \frac{2L}{m}\E{\norm{x_{m}-x^*}^2},\label{T2_contraction}
\end{align}
where \mar{$a$} follows from our parameter choice of $\eta = \frac{1}{4L\theta}$, and \mar{$b$} uses the Jensen's inequality.

At this point, we first check the constraint~(\ref{T2_cons}),
\[
	\eta \leq \frac{1 - \theta}{L\theta(2-\theta)} \rightarrow \theta \leq \frac{2}{3},
\]
which is satisfied by the parameter setting $\theta = \frac{2}{s + 4} \leq \frac{1}{2}$.

Then by evaluating~(\ref{T2_contraction}) with $s = 1$, we have
\[
	\frac{1}{\theta_1^2} \E{F(\tilde{x}_1) - F(x^*)} +\frac{2 L}{m}\E{\norm{x^1_{m}-x^*}^2} \leq \frac{1-\theta_1}{\theta_1^2}\big(F(\tilde{x}_{0}) - F(x^*)\big) + \frac{2L}{m}\norm{x^1_{0} - x^*}^2.
\]
Notice that here we mark iterates and $\theta$ with epoch number to prevent confusion.

Since $\theta_s = \frac{2}{s + 4}$, one can easily verify that $ \frac{1 - \theta_{s+1}}{\theta_{s+1}^2} \leq \frac{1}{\theta_s^2}$. Thus we can telescope the above inequality from $s=1\dots \mathcal{S}$,
\[
\begin{gathered}
	\frac{1}{\theta_{\mathcal{S}}^2} \E{F(\tilde{x}_{\mathcal{S}}) - F(x^*)} + \frac{2L}{m}\E{\norm{x^{\mathcal{S}}_{m}-x^*}^2} \leq \frac{1-\theta_1}{\theta_1^2}\big(F(\tilde{x}_{0}) - F(x^*)\big) + \frac{2L}{m}\norm{x^1_{0} - x^*}^2, \\
	\E{F(\tilde{x}_{\mathcal{S}}) - F(x^*)} \leq \frac{4(1-\theta_1)}{(\mathcal{S} + 4)^2\theta_1^2}\big(F(\tilde{x}_{0}) - F(x^*)\big) + \frac{8L}{(\mathcal{S} + 4)^2 m}\norm{x^1_{0} - x^*}^2.
\end{gathered}
\]
In other words, by choosing $m=\Theta(n)$, the total oracle complexity is $\bigO{n\sqrt{\frac{F(\tilde{x}_{0}) - F(x^*)}{\epsilon}} + \sqrt{\frac{nL\norm{x^1_{0} - x^*}^2}{\epsilon}}}$.
\section{Proofs for Section~\ref{sparse_async}}

\begin{its_variance}[Sparse Variance Bound 1]\label{its_variance}
	If Assumption~\ref{asyspa_assump} holds, for any $y,\tilde{x} \in \R^d$ and the sample $i_j$, denote $\tilde{\nabla} = \pfij{y} - \pfij{\tilde{x}} + D_{i_j}\pF{\tilde{x}}$, where $D_{i_j}$ is defined in Section~\ref{sec_serial_sparse_mig}, we can bound the variance $\Eij{\norm{\tilde{\nabla}}^2}$ as
	\[
		\Eij{\norm{\tilde{\nabla}}^2} \leq 4L\big(F(y) - F(x^*)\big) + 4L\big(F(\tilde{x}) - F(x^*)\big).
	\]
	\begin{proof}
		From Lemma 10 in~\cite{man:perturbed}, we have
		\begin{equation}\label{S3_1}
			\Eij{\norm{\tilde{\nabla}}^2} \leq 2\Eij{\norm{\pfij{y} - \pfij{x^*}}^2} + 2\Eij{\norm{\pfij{\tilde{x}} - \pfij{x^*}}^2},
		\end{equation}
		which provides an upper bound for the variance of the sparse stochastic variance reduced gradient estimator.
		
		From Theorem 2.1.5 in~\cite{nesterov:co}, we have
		\[
			\norm{\nabla f_{i_j} (y) - \nabla f_{i_j} (x^*)}^2
			\leq 2L\big( f_{i_j}(y) - f_{i_j}(x^*) - \langle \nabla f_{i_j}(x^*),
			y - x^*  \rangle \big).
		\]
		Taking expectation with the sample $i_j$, the above inequality becomes
		\[
			\Eij{\norm{\nabla f_{i_j} (y) - \nabla f_{i_j} (x^*)}^2}
			\leq 2L\big( F(y) - F(x^*)\big).
		\]
		Similarly, we have
		\[
			\Eij{\norm{\nabla f_{i_j} (\tilde{x}) - \nabla f_{i_j} (x^*)}^2}
			\leq 2L\big( F(\tilde{x}) - F(x^*)\big).
		\]
		Substituting the above inequalities into~(\ref{S3_1}) yields the desired result.
	\end{proof}
\end{its_variance}

\begin{s_variance}[Sparse Variance Bound 2]\label{s_variance}
	If Assumption~\ref{asyspa_assump} holds, using same notations as in Algorithm~\ref{sparse_mig}, we can bound the variance as
	\[
	\begin{aligned}
		\Eij{\norm{\pF{y_{j-1}} - \tilde{\nabla}_{\!S}}^2}&\leq 2L\big( F(\tilde{x}_{s-1}) - F(y_{j-1}) - \langle \pF{y_{j-1}},
		\tilde{x}_{s-1} - y_{j-1}  \rangle \big)\\
		& \ \ \ \  + 2L(D_m^2 - D_m)\cdot\big(F(\tilde{x}_{s-1}) - F(x^*)\big),
	\end{aligned}
	\]
	where $D_m = \max_{k=1\ldots d} {\frac{1}{p_k}}$, i.e., the inverse probability of the most sparse coordinates.
	
	\begin{proof}	
	First, we expand the variance term as
	\begin{align}
		&\ \ \ \ \Eij{\norm{\pF{y_{j-1}} - \tilde{\nabla}_{\!S}}^2} \nonumber
		\\&= \EijBig{\normbig{\big(\nabla f_{i_j} (y_{j-1}) - \nabla f_{i_j} (\tilde{x}_{s-1})\big) - \big(\pF{y_{j-1}} - D_{i_j}\pF{\tilde{x}_{s-1}}\big)}^2} \nonumber\\
		&= \Eij{\norm{\nabla f_{i_j} (y_{j-1}) - \nabla f_{i_j} (\tilde{x}_{s-1})}^2} -2\Eij{\langle \nabla f_{i_j} (y_{j-1}) - \nabla f_{i_j} (\tilde{x}_{s-1}),\pF{y_{j-1}} - D_{i_j}\pF{\tilde{x}_{s-1}} \rangle}\label{S3_2}\\
		& \ \ \ \ + \Eij{\norm{\pF{y_{j-1}} - D_{i_j}\pF{\tilde{x}_{s-1}}}^2}. \nonumber
	\end{align}
	Then we focus on the last two terms in~(\ref{S3_2}),
	\begin{align}
		& \ \ \ \ \Eij{\norm{\pF{y_{j-1}} - D_{i_j}\pF{\tilde{x}_{s-1}}}^2}\nonumber \\
		&= \norm{\pF{y_{j-1}}}^2 - 2\Eij{\langle \pF{y_{j-1}}, D_{i_j} \pF{\tilde{x}_{s-1}} \rangle} + \Eij{\norm{D_{i_j}\pF{\tilde{x}_{s-1}}}^2} \nonumber\\
		&\meq{\star} \norm{\pF{y_{j-1}}}^2 - 2\langle \pF{y_{j-1}}, \pF{\tilde{x}_{s-1} }\rangle + \langle \pF{\tilde{x}_{s-1}}, D\pF{\tilde{x}_{s-1}} \rangle, \label{L1_term1}
	\end{align}
	where \mar{$\star$} uses the fact that $P_{{i_j}} = P_{{i_j}}\cdot P_{{i_j}}$ and the unbiased sparse estimator $\Eij{D_{i_j}\pF{x}} = \pF{x}$.
	\begin{align}
		&\ \ \ \ \Eij{\langle \nabla f_{i_j} (y_{j-1}) - \nabla f_{i_j} (\tilde{x}_{s-1}),\pF{y_{j-1}} - D_{i_j}\pF{\tilde{x}_{s-1}} \rangle} \nonumber\\
		&= \norm{\pF{y_{j-1}}}^2 - \langle \pF{y_{j-1}}, \pF{\tilde{x}_{s-1} }\rangle - \Eij{\langle \nabla f_{i_j}(y_{j-1}), D_{i_j}\pF{\tilde{x}_{s-1} }\rangle}\nonumber\\
		&\ \ \ \ + \Eij{\langle \nabla f_{i_j} (\tilde{x}_{s-1}), D_{i_j}\pF{\tilde{x}_{s-1}}\rangle}\nonumber\\
		&\meq{\star} \norm{\pF{y_{j-1}}}^2 - \langle \pF{y_{j-1}}, \pF{\tilde{x}_{s-1} }\rangle - \langle \pF{y_{j-1}}, D\pF{\tilde{x}_{s-1} }\rangle + \langle\pF{\tilde{x}_{s-1}}, D\pF{\tilde{x}_{s-1}}\rangle \label{L1_term2},
	\end{align}
	where \mar{$\star$} follows from $\langle \nabla f_{i_j}(y_{j-1}), D_{i_j}\pF{\tilde{x}_{s-1}}\rangle = \langle \nabla f_{i_j}(y_{j-1}), D\pF{\tilde{x}_{s-1}}\rangle$, since $\nabla f_{i_j}(y_{j-1})$ and $D_{i_j}$ are both supported on the sample $i_j$.
	
	By substituting~(\ref{L1_term1}) and~(\ref{L1_term2}) into~(\ref{S3_2}), we get
	\begin{align}
		&\ \ \ \ \Eij{\norm{\pF{y_{j-1}} - \tilde{\nabla}_{\!S}}^2} \nonumber\\
		&= \Eij{\norm{\nabla f_{i_j} (y_{j-1}) - \nabla f_{i_j} (\tilde{x}_{s-1})}^2}- \norm{\pF{y_{j-1}}}^2 + 2\langle \pF{y_{j-1}}, D\pF{\tilde{x}_{s-1} }\rangle - \norm{D\pF{\tilde{x}_{s-1}}}^2\nonumber\\
		& \ \ \ \  + \langle (D-I)\pF{\tilde{x}_{s-1}}, D\pF{\tilde{x}_{s-1}} \rangle \nonumber\\
		&= \Eij{\norm{\nabla f_{i_j} (y_{j-1}) - \nabla f_{i_j} (\tilde{x}_{s-1})}^2} -
		\norm{\pF{y_{j-1}} - D\pF{\tilde{x}_{s-1}}}^2 + \langle (D-I)\pF{\tilde{x}_{s-1}}, D\pF{\tilde{x}_{s-1}} \rangle \nonumber\\
		&\leq \Eij{\norm{\nabla f_{i_j} (y_{j-1}) - \nabla f_{i_j} (\tilde{x}_{s-1})}^2} + \langle(D-I) \pF{\tilde{x}_{s-1}}, D\pF{\tilde{x}_{s-1}} \rangle.\label{S3_3}
	\end{align}
	From Theorem 2.1.5 in~\cite{nesterov:co}, we have
	\begin{equation}\label{L1_nesterov}
		\norm{\nabla f_{i_j} (y_{j-1}) - \nabla f_{i_j} (\tilde{x}_{s-1})}^2
		\leq 2L\big( f_{i_j}(\tilde{x}_{s-1}) - f_{i_j}(y_{j-1}) - \langle \nabla f_{i_j}(y_{j-1}),
		\tilde{x}_{s-1} - y_{j-1}  \rangle \big).
	\end{equation}
	By substituting~(\ref{L1_nesterov}) into~(\ref{S3_3}), we obtain
	\[
	\begin{aligned}
		&\ \ \ \ \Eij{\norm{\pF{y_{j-1}} - \tilde{\nabla}_{\!S}}^2}\\
		&\leq 2L\big( F(\tilde{x}_{s-1}) - F(y_{j-1}) - \langle \pF{y_{j-1}}, \tilde{x}_{s-1} - y_{j-1}  \rangle \big) + \langle(D-I) \pF{\tilde{x}_{s-1}}, D\pF{\tilde{x}_{s-1}} \rangle.
	\end{aligned}
	\]
	Since $D_m = \max_{k=1\ldots d} {\frac{1}{p_k}}$, we can bound the last term as follows:
	\[
	\begin{aligned}
		&\ \ \ \ \langle (D-I)\pF{\tilde{x}_{s-1}}, D\pF{\tilde{x}_{s-1}} \rangle\\ &= \sum_{k=1}^d {(\frac{1}{p_k^2} - \frac{1}{p_k}) \brk{\pF{\tilde{x}_{s-1}}}_k^2} \\
		&\leq (D_m^2 - D_m)\norm{\pF{\tilde{x}_{s-1}} - \pF{x^*}}^2 \\
		&\mleq{\star} 2L(D_m^2 - D_m)\cdot\big(F(\tilde{x}_{s-1}) - F(x^*)\big),
	\end{aligned}
	\]
	where \mar{$\star$} follows from applying Theorem 2.1.5 in~\cite{nesterov:co}.
	\end{proof}
\end{s_variance}

\subsection{Proof of Theorem~\ref{serial_sparse_rate}}
For the purpose of analyzing in the asynchronous setting, we need to analyze the convergence of MiG starting with iterate difference.

We start with the iterate difference between $y_j$ and $x^*$. By expanding iterate difference and taking expectation with respect to the sample $i_j$, we get
\begin{align}
	\Eij{\norm{y_j - x^*}^2} &\meq{a} \Eij{\norm{\theta(x_{j-1} - \eta \cdot \tilde{\nabla}_{\!S}) + (1 - \theta)\tilde{x}_{s-1} - x^*}^2} \nonumber\\
	&= \Eij{\norm{y_{j-1} - \eta \theta \cdot \tilde{\nabla}_{\!S} - x^*}^2} \nonumber\\
	&\meq{b} \norm{y_{j-1} - x^*}^2 - 2\eta \theta\langle \pF{y_{j-1}}, y_{j-1} - x^* \rangle + \eta^2\theta^2\Eij{\norm{\tilde{\nabla}_{\!S}}^2}, \label{C_1_1}
\end{align}
where \mar{$a$} uses the definition of $y$, and \mar{$b$} follows from the unbiasedness of the sparse gradient estimator $\Eij{\tilde{\nabla}_{\!S}} = \pF{y_{j-1}}$.

Using $\sigma$-strongly convex, we get a bound for $-\langle \pF{y_{j-1}}, y_{j-1} - x^* \rangle$ as
\begin{equation} \label{C_1_2}
	\langle \pF{y_{j-1}}, y_{j-1} - x^* \rangle \geq F(y_{j-1}) - F(x^*) + \frac{\sigma}{2}\norm{y_{j-1} - x^*}^2.
\end{equation}
Using Lemma~\ref{its_variance}, we have the following variance bound:
\begin{equation}\label{C_1_3}
	\Eij{\norm{\tilde{\nabla}_{\!S}}^2} \leq 4L\big(F(y_{j-1}) - F(x^*)\big) + 4L\big(F(\tilde{x}_{s-1}) - F(x^*)\big).
\end{equation}
By combining~(\ref{C_1_1}),~(\ref{C_1_2}) and~(\ref{C_1_3}), we get
\[
\begin{aligned}
	\Eij{\norm{y_j - x^*}^2} &\leq \norm{y_{j-1} - x^*}^2 - \eta \theta\sigma\norm{y_{j-1} - x^*}^2 - 2\eta\theta \big(F(y_{j-1}) - F(x^*)\big) + \eta^2\theta^2\Eij{\norm{\tilde{\nabla}_{\!S}}^2} \\
	&\leq (1 - \eta\theta\sigma)\cdot\norm{y_{j-1} - x^*}^2 + (4L\eta^2\theta^2 - 2\eta\theta)\cdot \big(F(y_{j-1}) - F(x^*)\big)\\
	& \ \ \ \  +  4L\eta^2\theta^2\big(F(\tilde{x}_{s-1}) - F(x^*)\big), \\
\end{aligned}	
\]\[
\begin{aligned}
	&\ \ \ \ (2\eta\theta - 4L\eta^2\theta^2)\cdot \big(F(y_{j-1}) - F(x^*)\big) \\
	&\leq \norm{y_{j-1} - x^*}^2 - \Eij{\norm{y_j - x^*}^2} + 4L\eta^2\theta^2\big(F(\tilde{x}_{s-1}) - F(x^*)\big).
\end{aligned}	
\]
Summing the above inequality over $j = 1\ldots m$ and taking expectation with respect to all randomness in this epoch, we get
\[
\begin{aligned}
	&\ \ \ \ (2\eta\theta - 4L\eta^2\theta^2)\cdot \E{\sum_{j = 1}^m{\big(F(y_{j-1}) - F(x^*)\big)}} \\
	&\leq \norm{y_0 - x^*}^2 - \E{\norm{y_m - x^*}^2} + 4L\eta^2\theta^2m\big(F(\tilde{x}_{s-1}) - F(x^*)\big).
\end{aligned}
\]
Using Jensen's inequality and $\tilde{x}_s = \frac{1}{m} \sum_{j = 0}^{m-1}{y_j}$, we obtain
\[
\begin{aligned}
	(2\eta\theta - 4L\eta^2\theta^2)m\cdot \E{\big(F(\tilde{x}_s) - F(x^*)\big)} &\leq \norm{y_0 - x^*}^2 + 4L\eta^2\theta^2m\big(F(\tilde{x}_{s-1}) - F(x^*)\big), \\
	\E{\big(F(\tilde{x}_s) - F(x^*)\big)} &\mleq{\star} \frac{\frac{2}{\sigma} + 4L\eta^2\theta^2m}{(2\eta\theta - 4L\eta^2\theta^2)m}\cdot\big(F(\tilde{x}_{s-1}) - F(x^*)\big),
\end{aligned}
\]
where \mar{$\star$} follows from the fact $x_0 = \tilde{x}_{s-1} = y_0$ and $\sigma$-strongly convexity of $F(\cdot)$.

By choosing $m = 25\kappa$, $\eta = \frac{1}{L}$, $\theta = \frac{1}{10}$, the above inequality becomes
\[
\begin{aligned}
	\E{\big(F(\tilde{x}_s) - F(x^*)\big)} &\leq 0.75\cdot \big(F(\tilde{x}_{s-1}) - F(x^*)\big),
\end{aligned}
\]
which means that the total oracle complexity is $\bigO{(n+\kappa)\log{\frac{F(\tilde{x}_0) - F(x^*)}{\epsilon}}}$.

\subsection{Proof of Theorem~\ref{serial_sparse_frate}}
\label{sparse_mig_optionII}
Here we consider improving the convergence rate for the Serial Sparse MiG, referencing to Algorithm~\ref{sparse_mig} with Option II.
Our analysis is based on function difference similar to the proofs in Appendix~\ref{proofs_sec2}, which make the parameter $\theta$ effective.

Similarly, we first add the following constraint:
\begin{equation}\label{C234_cons}
	L\theta + \frac{L\theta}{1 - \theta} \leq \frac{1}{\eta}.
\end{equation}
We start with the convexity of $F(\cdot)$ at $y_{j-1}$ and $x^*$,
\begin{equation} \label{C_2_1}
\begin{aligned}
	&\ \ \ \ F(y_{j-1}) - F(x^*) \leq \langle \pF{y_{j-1}}, y_{j-1} - x^* \rangle\\
	&= \frac{1 - \theta}{\theta}\langle \pF{y_{j-1}}, \tilde{x}_{s-1} - y_{j-1} \rangle +  \langle \pF{y_{j-1}}, x_{j-1} - x^* \rangle.	
\end{aligned}
\end{equation}
Then we further expand $\langle \pF{y_{j-1}}, x_{j-1} - x^* \rangle$ as
\begin{equation}\label{C_2_2}
	\langle \pF{y_{j-1}}, x_{j-1} - x^* \rangle = \langle \pF{y_{j-1}} - \tilde{\nabla}_{\!S}, x_{j-1} - x^* \rangle + \langle \tilde{\nabla}_{\!S}, x_{j-1} - x_j \rangle +  \langle\tilde{\nabla}_{\!S}, x_j - x^* \rangle.
\end{equation}
Using $L$-smooth~(\ref{l-smooth}) of $F(\cdot)$ at $(y_{j}, y_{j-1})$, we can bound $\langle \tilde{\nabla}_{\!S}, x_{j-1} - x_{j} \rangle$ as
\[
\begin{aligned}
	F(y_j)-F(y_{j-1}) &\leq \langle \pF{y_{j-1}}, y_j - y_{j-1} \rangle + \frac{L}{2} \norm{y_j - y_{j-1}}^2 \\
	&\meq{\star} \theta \big[\langle \pF{y_{j-1}} - \tilde{\nabla}_{\!S}, x_j - x_{j-1} \rangle + \langle \tilde{\nabla}_{\!S}, x_j - x_{j-1} \rangle\big] + \frac{L\theta^2}{2} \norm{x_j - x_{j-1}}^2, \\
	\langle \tilde{\nabla}_{\!S}, x_{j-1} - x_{j} \rangle &\leq \frac{1}{\theta}\big(F(y_{j-1}) - F(y_{j})\big) + \langle \pF{y_{j-1}} - \tilde{\nabla}_{\!S}, x_j - x_{j-1} \rangle + \frac{L\theta}{2} \norm{x_j - x_{j-1}}^2,
\end{aligned}
\]
where \mar{$\star$} uses the definition of $y$.

Using the constraint~(\ref{C234_cons}), we have
\begin{align}
	\langle \tilde{\nabla}_{\!S}, x_{j-1} - x_{j} \rangle &\leq \frac{1}{\theta}\big(F(y_{j-1}) - F(y_{j})\big) + \langle \pF{y_{j-1}} - \tilde{\nabla}_{\!S}, x_j - x_{j-1} \rangle + \frac{1}{2\eta} \norm{x_j - x_{j-1}}^2 \label{C_2_3}\\
	& \ \ \ \ - \frac{L\theta}{2(1-\theta)}\norm{x_j - x_{j-1}}^2. \nonumber
\end{align}
Since we can write an equivalent update as $x_j = \argmin_{x} \lbrace \frac{1}{2} \norm{x - x_{j-1}}^2 + \eta \langle \tilde{\nabla}_{\!S}, x \rangle \rbrace$, by applying Lemma~\ref{3_point} with $z = x^*$, $z_0 = x_{j-1}$, $z^* = x_j$, $\tau = \frac{1}{\eta}$, $\psi(x) = \langle \tilde{\nabla}_{\!S}, x \rangle$, we have
\begin{equation} \label{C_2_4}
	\langle \tilde{\nabla}_{\!S}, x_{j} - x^* \rangle = -\frac{1}{2\eta}\norm{x_{j-1} - x_{j}}^2 + \frac{1}{2\eta}\norm{x_{j-1} - x^*}^2 - \frac{1}{2\eta}\norm{x_{j}-x^*}^2.
\end{equation}
Combining~(\ref{C_2_1}),~(\ref{C_2_2}),~(\ref{C_2_3}) and~(\ref{C_2_4}), we get
\[
\begin{aligned}
	F(y_{j-1}) - F(x^*)&\leq\frac{1 - \theta}{\theta}\langle \pF{y_{j-1}}, \tilde{x}_{s-1} - y_{j-1} \rangle + \frac{1}{\theta}\big(F(y_{j-1}) - F(y_{j})\big) + \langle \pF{y_{j-1}} - \tilde{\nabla}_{\!S}, x_j - x^* \rangle \\
	& \ \ \ \ - \frac{L\theta}{2(1-\theta)}\norm{x_j - x_{j-1}}^2 + \frac{1}{2\eta} \norm{x_{j-1} - x^*}^2 - \frac{1}{2\eta} \norm{x_j - x^*}^2.
\end{aligned}
\]
Taking expectation with respect to the sample $i_j$, we obtain
\[
\begin{aligned}
	F(y_{j-1}) - F(x^*) &\mleq{a} \frac{1 - \theta}{\theta}\langle \pF{y_{j-1}}, \tilde{x}_{s-1} - y_{j-1} \rangle + \frac{1}{\theta}\big(F(y_{j-1}) - \Eij{F(y_{j})}\big)- \frac{L\theta}{2(1-\theta)}\Eij{\norm{x_j - x_{j-1}}^2} \\
	& \ \ \ \ + \Eij{\langle \pF{y_{j-1}} - \tilde{\nabla}_{\!S}, x_j - x_{j-1} \rangle}  + \frac{1}{2\eta} \norm{x_{j-1} - x^*}^2 - \frac{1}{2\eta} \Eij{\norm{x_j - x^*}^2} \\
	&\mleq{b} \frac{1 - \theta}{\theta}\langle \pF{y_{j-1}}, \tilde{x}_{s-1} - y_{j-1} \rangle + \frac{1}{\theta}\big(F(y_{j-1}) - \Eij{F(y_{j})}\big)- \frac{L\theta}{2(1-\theta)}\Eij{\norm{x_j - x_{j-1}}^2} \\
	& \ \ \ \ + \frac{1}{2\beta} \Eij{\norm{\pF{y_{j-1}} - \tilde{\nabla}_{\!S}}^2} + \frac{\beta}{2}\Eij{\norm{x_j - x_{j-1}}^2}  + \frac{1}{2\eta} \norm{x_{j-1} - x^*}^2 - \frac{1}{2\eta} \Eij{\norm{x_j - x^*}^2},
\end{aligned}
\]
where \mar{$a$} uses the unbiasedness of the sparse gradient estimator $\Eij{\tilde{\nabla}_{\!S}} = \pF{y_{j-1}}$, and \mar{$b$} uses the Young's inequality with the parameter $\beta > 0$.

Using Lemma~\ref{s_variance} to bound the variance term, we get
\[
\begin{aligned}
	F(y_{j-1}) - F(x^*) &\leq \frac{1 - \theta}{\theta}\langle \pF{y_{j-1}}, \tilde{x}_{s-1} - y_{j-1} \rangle + \frac{1}{\theta}\big(F(y_{j-1})- \Eij{F(y_{j})}\big)  - \frac{L\theta}{2(1-\theta)}\Eij{\norm{x_j - x_{j-1}}^2} \\
	& \ \ \ \  + \frac{L}{\beta}\big( F(\tilde{x}_{s-1}) - F(y_{j-1}) - \langle \pF{y_{j-1}},\tilde{x}_{s-1} - y_{j-1}  \rangle \big) + \frac{L(D_m^2 - D_m)}{\beta}\big(F(\tilde{x}_{s-1}) - F(x^*)\big) \\
	& \ \ \ \  + \frac{\beta}{2}\Eij{\norm{x_j - x_{j-1}}^2}+ \frac{1}{2\eta} \norm{x_{j-1} - x^*}^2  - \frac{1}{2\eta} \Eij{\norm{x_j - x^*}^2}.
\end{aligned}
\]
By choosing $\beta = \frac{L\theta}{1-\theta} > 0$, the above inequality becomes
\[
\begin{aligned}
	F(y_{j-1}) - F(x^*) &\leq \frac{1}{\theta}\big(F(y_{j-1}) - \Eij{F(y_{j})}\big) + \frac{1-\theta}{\theta}\big( F(\tilde{x}_{s-1}) - F(y_{j-1})\big)\\
	& \ \ \ \ + \frac{1-\theta}{\theta}(D_m^2 - D_m)\cdot\big(F(\tilde{x}_{s-1}) - F(x^*)\big)  + \frac{1}{2\eta} \norm{x_{j-1} - x^*}^2 - \frac{1}{2\eta} \Eij{\norm{x_j - x^*}^2}, \\
	\frac{1}{\theta}\big(\Eij{F(y_{j})} - F(x^*)\big)&\leq \frac{1-\theta}{\theta}\big(F(\tilde{x}_{s-1}) - F(x^*)\big) + \frac{1-\theta}{\theta}(D_m^2 - D_m)\cdot\big(F(\tilde{x}_{s-1}) - F(x^*)\big)  \\
	& \ \ \ \ + \frac{1}{2\eta} \norm{x_{j-1} - x^*}^2 - \frac{1}{2\eta} \Eij{\norm{x_j - x^*}^2}.
\end{aligned}
\]
Summing the above inequality over $j=1\ldots m$ and taking expectation with respect to all randomness in this epoch, we get
\[
\begin{aligned}
	&\ \ \ \ \frac{1}{\theta}\Big(\frac{1}{m}\sum_{j = 1}^m{\E{F(y_{j})}} - F(x^*)\Big)\\
	&\leq \frac{1-\theta}{\theta}(1 + D_m^2 - D_m)\cdot\big(F(\tilde{x}_{s-1}) - F(x^*)\big) + \frac{1}{2\eta m} \Big(\norm{x_{0} - x^*}^2 - \E{\norm{x_m - x^*}^2}\Big).
\end{aligned}
\]

Using Jensen's inequality and $\tilde{x}_s = \frac{1}{m}\sum_{j = 1}^m y_j$ in Option II, we have
\[
\begin{aligned}
	&\ \ \ \ \frac{1}{\theta}\E{F(\tilde{x}_s) - F(x^*)}\\
	&\leq \frac{1-\theta}{\theta}\Big( 1 + D_m^2 - D_m\Big)\cdot\big(F(\tilde{x}_{s-1}) - F(x^*)\big) + \frac{1}{2\eta m} \Big(\norm{x_{0} - x^*}^2 - \E{\norm{x_m - x^*}^2}\Big).
\end{aligned}
\]
In order to give a clean proof, we set $\tilde{D}_s \triangleq F(\tilde{x}_s) - F(x^*)$, $P^s_0 \triangleq \norm{x^s_{0} - x^*}^2$ and the terms that related to $D_m$ as $\zeta = D_m^2 - D_m$. Then the inequality becomes
\[
\begin{aligned}
	\frac{1}{\theta}\E{\tilde{D}_s}&\leq \frac{1-\theta}{\theta}(1 + \zeta)\cdot\tilde{D}_{s-1} + \frac{1}{2\eta m}\E{P^s_0 - P^s_m}, \\
	\frac{\theta + \theta\zeta - \zeta}{\theta}\cdot\E{\tilde{D}_s}&\leq \frac{1-\theta}{\theta}(1 + \zeta)\cdot\E{\tilde{D}_{s-1} - \tilde{D}_s} + \frac{1}{2\eta m} \E{P^s_0 - P^s_m}.
\end{aligned}
\]
Suppose we run $\mathcal{S}$ epochs before a restart. By summing the above inequality over $s = 1\ldots\mathcal{S}$, we get
\[
	\frac{\theta + \theta\zeta - \zeta}{\theta}\cdot\sum_{s=1}^{\mathcal{S}}\E{\tilde{D}_s} \leq \frac{1-\theta}{\theta}( 1 + \zeta)\cdot\E{\tilde{D}_{0} - \tilde{D}_{\mathcal{S}}} + \frac{1}{2\eta m} \E{P^1_0 - P^{\mathcal{S}}_m}.
\]
Choosing the initial vector for next $\mathcal{S}$ epochs as $x^{new}_0 = \frac{1}{\mathcal{S}} \sum_{s=1}^{\mathcal{S}} {\tilde{x}_s}$, we have
\[
\begin{aligned}
	\mathcal{S}\cdot\frac{\theta + \theta\zeta - \zeta}{\theta}\cdot\E{\tilde{D}^{new}_0}&\leq \frac{1-\theta}{\theta}(1 + \zeta)\cdot\E{\tilde{D}_{0} - \tilde{D}_{\mathcal{S}}} + \frac{1}{2\eta m} \E{P^1_0 - P^{\mathcal{S}}_m} \\
	&\mleq{\star} \big(\frac{(1-\theta)( 1 + \zeta )}{\theta} + \frac{1}{\eta m\sigma}\big)\cdot\tilde{D}_{0}, \\
	\E{\tilde{D}^{new}_0}&\leq \frac{(1-\theta)\cdot( 1 + \zeta ) + \frac{\theta}{\eta m\sigma}}{\mathcal{S}\cdot(\theta + \zeta\theta - \zeta)}\tilde{D}_{0},
\end{aligned}
\]
where \mar{$\star$} uses the $\sigma$-strongly convexity of $F(\cdot)$ and $\tilde{x}_0 = x^1_0$, that is, $P^1_0 \leq \frac{2}{\sigma} \tilde{D}_0$.

Setting $\mathcal{S} = \Big\lceil {2\cdot\frac{(1-\theta)\cdot( 1 + \zeta ) + \frac{\theta}{\eta m\sigma}}{\theta + \zeta\theta - \zeta}}\Big\rceil$, we have that $\tilde{D}_0$ decreases by a factor of at least $\frac{1}{2}$ every $\mathcal{S}$ rounds of epochs. So in order to achieve an $\epsilon$-additive error, we need totally $O(\log{\frac{\tilde{D}_0}{\epsilon}})$ restarts of the above procedure.

\subsubsection{Parameter setting for two cases}\label{proof_T3}
\textbf{(I)} Consider the first case with $\frac{m}{\kappa} \leq \frac{3}{4}$, by choosing identical parameters settings $\eta = \sqrt{\frac{1}{3\sigma m L}}$,  $\theta = \sqrt{\frac{m}{3\kappa}} \leq \frac{1}{2}$ and $m = \Theta(n)$ as in Section~\ref{sequential} (so the constraint~(\ref{C234_cons}) is satisfied), we have
\[
	\mathcal{S} = \Bigg\lceil{2\cdot \frac{2 - \big(\theta + \zeta\cdot(\theta - 1)\big)}{\theta + \zeta\cdot(\theta - 1)}}\Bigg\rceil.
\]
Imposing an additional constraint on the sparse variance: $\zeta \leq \sqrt{\frac{m}{4\kappa}}$, we have $\mathcal{S} = O({\sqrt{\frac{\kappa}{n}}})$, which means that the total oracle complexity is
\[
	\mathcal{O}{\left(\mathcal{S}\cdot O\left({\log{\frac{\tilde{D}_0}{\epsilon}}}\right) \cdot (m+n)\right)} = \mathcal{O}{\left(\sqrt{\kappa n}\log{\frac{F(\tilde{x}_0) - F(x^*)}{\epsilon}}\right)}
\]

\textbf{(II)} Consider another case with $\frac{m}{\kappa} > \frac{3}{4}$ and $\zeta \leq C_{\zeta}$, let $\hat{\theta} \triangleq\frac{C_{\zeta} + 0.5}{C_{\zeta} + 1}$, by choosing $\theta = \frac{\zeta + 0.5}{\zeta + 1} \in [0.5, \hat{\theta}]$, $\eta = \frac{1-\hat{\theta}}{2m\sigma\hat{\theta}} \leq \frac{1-\theta}{L\theta(2-\theta)}$ (the constraint~(\ref{C234_cons}) is satisfied) and $m = \Theta(n)$, we have $\mathcal{S}=O(1)$ (correlated with $C_{\zeta}$), the total oracle complexity $\bigO{n\log{\frac{F(\tilde{x}_0) - F(x^*)}{\epsilon}}}$.

\subsubsection{Discussion about Sparse Variance Bound (Lemma~\ref{its_variance} and Lemma~\ref{s_variance})}\label{discuss_variance_bound}
In the dense update case (with $D_m = 1$), we see that Lemma~\ref{s_variance} degenerates to Lemma~\ref{variance}, this bound is much tighter than the bound in Lemma~\ref{its_variance} (Lemma~\ref{its_variance} is identical for both the sparse and dense cases).

However, in the sparse update case, Lemma~\ref{s_variance} highly correlates with dataset sparsity ($\propto D_m^2$), which could be much looser than Lemma~\ref{its_variance} in some extreme cases (imagine a dataset with some dimensions contain only one entry among the $n$ samples, so $D_m = n$). Unfortunately, our MiG algorithm (as well as Katyusha) relies on canceling the additional variance term to yield a tight correlation inside one iteration:
\[
	F(\tilde{x}_{s-1}) - F(y_{j-1}) - \boxed{\langle \pF{y_{j-1}},
	\tilde{x}_{s-1} - y_{j-1} \rangle} \rightarrow \text{canceled by coupling term } (1-\theta)\tilde{x}_s
\]
If $D_m$ is as large as $n$, the oracle complexity in Theorem~\ref{serial_sparse_frate} could be even worse than that in Theorem~\ref{serial_sparse_rate}.

It is still an open problem whether we can have a tighter variance bound in the sparse update setting that is uncorrelated with $D_m$.

\subsection{Proof of Theorem~\ref{asy_sparse_rate}}
\label{async_analysis}
Here we analyze Algorithm~\ref{async_mig} based on the ``perturbed iterate analysis'' framework~\cite{man:perturbed}.

To begin with, we need to specify the iterates labeling order, which is crucial in our asynchronous analysis.

\textit{Choice of labeling order. } There are ``Before Read''~\cite{man:perturbed} and ``After Read''~\cite{leb:asaga} labeling schemes proposed in recent years which are reasonable in asynchronous analysis. Among these two schemes, ``Before Read'' requires considering the updates from ``future'', which leads to a complex analysis. ``After Read'' enjoys a simpler analysis but requires changing the order of sampling step to ensure uniform distributed samples\footnote{So there are always two versions (analyzed, implemented) of algorithms in the works with ``After Read'' scheme~\cite{leb:asaga,ped:proxasaga}}. In order to give a clean proof, we adopt the ``After Read'' labeling scheme and make the following assumptions:
\begin{asy_assump_indp}\label{asy_assump_indp}
	The labeling order increases after the step (14) in Algorithm~\ref{async_mig} finished, so the future perturbation is not included in the effect of asynchrony in the current step.
\end{asy_assump_indp}
\begin{asy_assump_sample}\label{asy_assump_sample}
	We explicitly assume uniform distributed samples and the independence of the sample $i_j$ with $\hat{x}_{j-1}$.
\end{asy_assump_sample}
In other words, we are analyzing the following procedure:
\begin{enumerate}
	\item Inconsistent read the iterate $\hat{x}_{j-1}$.
	\item Increase iterates counter $j$ and sample a random index $i_j$.
	\item Compute an update $-\eta\cdot \tilde{\nabla}(\hat{y}_{j-1})$.
	\item Atomic write the update to shared memory coordinately.
\end{enumerate}

From~\cite{leb:asaga}, we can model the effect of asynchrony as follows:
\begin{equation} \label{asynchony}
	\hat{x}_{j} - x_{j} = \eta \sum_{k = (j-1-\tau)_+}^{j-2} \mathcal{S}^j_k\tilde{\nabla}(\hat{y}_{k}),
\end{equation}
where $\mathcal{S}^j_k$ is a diagonal matrix with entries in $\lbrace 0, +1\rbrace$. This definition models the coordinate perturbation from the past updates. Here $\tau$ represents the maximum number of overlaps between concurrent threads~\cite{man:perturbed}. We further denote $\Delta = \max_{k=1\ldots d}{p_k}$ following~\cite{leb:asaga}, which provides a measure of sparsity.

Then we start our analysis with the iterate difference between ``fake'' $y_j$ and $x^*$. By expanding the iterate difference and taking expectation with respect to the sample $i_j$, we get
\begin{align}
	\Eij{\norm{y_j - x^*}^2} &= \Eij{\norm{\theta(x_{j-1} - \eta
	\cdot\tilde{\nabla}(\hat{y}_{j-1})) + (1 - \theta)\tilde{x}_{s-1} - x^*}^2}\nonumber \\
	&=\Eij{\norm{y_{j-1} - \eta \theta\cdot\tilde{\nabla}(\hat{y}_{j-1}) - x^*}^2}\nonumber \\
	&\meq{\star} \norm{y_{j-1} - x^*}^2 - 2\eta \theta\langle \pF{\hat{y}_{j-1}}, \hat{y}_{j-1} - x^* \rangle + \eta^2\theta^2\Eij{\norm{\tilde{\nabla}(\hat{y}_{j-1})}^2}
	\label{T5_1}\\
	&\ \ \ \  + 2\eta \theta\Eij{\langle \tilde{\nabla}(\hat{y}_{j-1}), \hat{y}_{j-1} - y_{j-1} \rangle}, \nonumber
\end{align}
where \mar{$\star$} uses the unbiasedness $\Eij{\tilde{\nabla}(\hat{y}_{j-1})} = \pF{\hat{y}_{j-1}}$ and the independence Assumption~\ref{asy_assump_sample}.

Using Lemma~\ref{its_variance}, we get the variance bound
\begin{equation}\label{T5_2}
	\Eij{\norm{\tilde{\nabla}(\hat{y}_{j-1})}^2} \leq 4L\big(F(\hat{y}_{j-1}) - F(x^*)\big) + 4L\big(F(\tilde{x}_{s-1}) - F(x^*)\big).
\end{equation}
Using the $\sigma$-strongly convex of $F(\cdot)$, we can bound $-\langle \pF{\hat{y}_{j-1}}, \hat{y}_{j-1} - x^* \rangle$ as follows:
\begin{align}
	\langle \pF{\hat{y}_{j-1}}, \hat{y}_{j-1} - x^* \rangle &\geq F(\hat{y}_{j-1}) - F(x^*) + \frac{\sigma}{2}\norm{\hat{y}_{j-1} - x^*}^2 \nonumber \\
	&\mgeq{\star} F(\hat{y}_{j-1}) - F(x^*) + \frac{\sigma}{4}\norm{y_{j-1} - x^*}^2 - \frac{\sigma}{2}\norm{\hat{y}_{j-1} - y_{j-1}}^2, \label{T5_3}
\end{align}
where \mar{$\star$} uses the fact that $\norm{a + b}^2 \leq 2\norm{a}^2 + 2\norm{b}^2$.

Combining~(\ref{T5_1}),~(\ref{T5_2}) and~(\ref{T5_3}), we get
\begin{align}
	\Eij{\norm{y_j - x^*}^2}
	&\leq (1- \frac{\eta\theta\sigma}{2}) \norm{y_{j-1} - x^*}^2 + \eta\theta\sigma \norm{\hat{y}_{j-1} - y_{j-1}}^2+ 2\eta \theta\Eij{\langle \tilde{\nabla}(\hat{y}_{j-1}), \hat{y}_{j-1} - y_{j-1} \rangle}\label{T5_7}\\
	&\ \ \ \ + (4L\eta^2\theta^2 - 2\eta\theta)\big(F(\hat{y}_{j-1}) - F(x^*)\big) + 4L\eta^2\theta^2\big(F(\tilde{x}_{s-1}) - F(x^*)\big).\nonumber
\end{align}
From Lemma 1 in~\cite{leb:asaga}, we borrow the notations $C_1 = 1 + \sqrt{\Delta} \tau$, $C_2 = \sqrt{\Delta} + \eta\theta \sigma C_1$ and bound the asynchronous variance terms $ \norm{\hat{y}_{j-1} - y_{j-1}}^2$, $	\Eij{\langle\tilde{\nabla}(\hat{y}_{j-1}), \hat{y}_{j-1} - y_{j-1} \rangle}$ using~(\ref{asynchony}) as
\begin{gather}
	\Eij{\langle\tilde{\nabla}(\hat{y}_{j-1}), \hat{y}_{j-1} - y_{j-1} \rangle} \leq \frac{\eta \theta\sqrt{\Delta}}{2}\sum_{k = (j-1-\tau)_+}^{j-2}{ \norm{\tilde{\nabla}(\hat{y}_{k})}^2} + \frac{\eta \theta\sqrt{\Delta}\tau}{2} \Eij{\norm{\tilde{\nabla}(\hat{y}_{j-1})}^2},\label{T5_4}\\
	\norm{\hat{y}_{j-1} - y_{j-1}}^2 \leq \eta^2\theta^2 (1 + \sqrt{\Delta}\tau)\sum_{k = (j-1-\tau)_+}^{j-2} {\norm{\tilde{\nabla}(\hat{y}_{k})}^2}. \label{T5_5}
\end{gather}
Upper bounding the asynchronous terms in~(\ref{T5_7}) using~(\ref{T5_4}) and~(\ref{T5_5}), we get
\[
\begin{aligned}
	\Eij{\norm{y_j - x^*}^2}
	&\leq (1- \frac{\eta\theta\sigma}{2}) \norm{y_{j-1} - x^*}^2 + \eta^2 \theta^2(\sqrt{\Delta} + \eta\theta\sigma(1 + \sqrt{\Delta}\tau))\sum_{k = (j-1-\tau)_+}^{j-2}{ \norm{\tilde{\nabla}(\hat{y}_{k})}^2}\\
	& \ \ \ \  + (4L\eta^2\theta^2(1 + \sqrt{\Delta}\tau) - 2\eta\theta)\big(F(\hat{y}_{j-1}) - F(x^*)\big) + 4L\eta^2\theta^2(1 + \sqrt{\Delta}\tau)\big(F(\tilde{x}_{s-1}) - F(x^*)\big).
\end{aligned}
\]
Defining $a_j \triangleq \norm{y_j - x^*}^2$, $\hat{D}_{j-1} = F(\hat{y}_{j-1}) - F(x^*)$, $\tilde{D}_{s-1} = F(\tilde{x}_{s-1}) - F(x^*)$ for a clean proof and rearranging, we obtain
\begin{align}
	\Eij{a_j}
	&\leq (1- \frac{\eta\theta\sigma}{2}) a_{j-1} + \eta^2 \theta^2C_2\sum_{k = (j-1-\tau)_+}^{j-2}{ \norm{\tilde{\nabla}(\hat{y}_{k})}^2} +(4L\eta^2\theta^2C_1 - 2\eta\theta)\hat{D}_{j-1} + 4L\eta^2\theta^2C_1\tilde{D}_{s-1},\nonumber \\
	(2\eta\theta - &4L\eta^2\theta^2C_1)\hat{D}_{j-1}\mleq{\star} (a_{j-1} - \Eij{a_j})+ \eta^2 \theta^2C_2\sum_{k = (j-1-\tau)_+}^{j-2}{ \norm{\tilde{\nabla}(\hat{y}_{k})}^2} + 4L\eta^2\theta^2C_1\tilde{D}_{s-1},\label{T5_6}
\end{align}
where \mar{$\star$} uses the fact that $1-\frac{\eta\theta\sigma}{2} \leq 1$.

Summing~(\ref{T5_6}) over $j=1\ldots m$ and taking expectation with all randomness in this epoch, we get
\begin{equation}\label{T5_8}
	(2\eta\theta - 4L\eta^2\theta^2C_1) \sum_{j = 1}^m\E{\hat{D}_{j-1}} \leq (a_{0} - \E{a_m})+ \eta^2 \theta^2C_2\sum_{j = 1}^m{\sum_{k = (j-1-\tau)_+}^{j-2}{ \E{\norm{\tilde{\nabla}(\hat{y}_{k})}^2}}} + 4L\eta^2\theta^2C_1 m\tilde{D}_{s-1}.
\end{equation}
Then we focus on upper bounding the second term on the right side of~(\ref{T5_8}),
\[
	\sum_{j = 1}^m{\sum_{k = (j-1-\tau)_+}^{j-2}{\E{\norm{\tilde{\nabla}(\hat{y}_{k})}^2}}}\leq \tau \sum_{j=1}^{m-1} \E{\norm{\tilde{\nabla}(\hat{y}_{j-1})}^2} \leq \tau \sum_{j=1}^{m} \E{\norm{\tilde{\nabla}(\hat{y}_{j-1})}^2} \mleq{\star} 4L\tau\big(\sum_{j = 1}^m{\E{\hat{D}_{j-1}}} + m \tilde{D}_{s-1}\big),
\]
where \mar{$\star$} uses the variance bound~(\ref{T5_2}).

Substituting the above inequality into~(\ref{T5_8}), we get
\[
\begin{gathered}
	(2\eta\theta - 4L\eta^2\theta^2C_1 - 4L\eta^2 \theta^2C_2\tau) \sum_{j = 1}^m\E{\hat{D}_{j-1}} \leq a_{0} + (4L\eta^2\theta^2C_1 m + 4L\eta^2 \theta^2C_2\tau m)\tilde{D}_{s-1}, \\
	\tilde{D}_s \mleq{\star} \frac{\frac{2}{\sigma} + 4L\eta^2\theta^2C_1 m + 4L\eta^2 \theta^2C_2\tau m}{(2\eta\theta - 4L\eta^2\theta^2C_1 - 4L\eta^2 \theta^2C_2\tau)m}\cdot\tilde{D}_{s-1},
\end{gathered}
\]
where \mar{$\star$} uses the $\sigma$-strongly convex of $F(\cdot)$ and $\tilde{x}_0 = x_0 = y_0$, $\tilde{x}_s = \frac{1}{m}\sum_{j = 0}^{m-1} \hat{y}_j$.

By choosing $m = 60\kappa$, $\eta = \frac{1}{5L}$, $\theta = \frac{1}{6}$, we get
\[
	\tilde{D}_s \leq \frac{2 + \frac{4}{15}(C_1 + C_2\tau)}{4 - \frac{4}{15}(C_1 + C_2\tau)}\cdot\tilde{D}_{s-1}.
\]
In order to ensure linear speed up, $\tau$ needs to satisfy the following constraint:
\[
	\rho \triangleq \frac{2 + \frac{4}{15}(C_1 + C_2\tau)}{4 - \frac{4}{15}(C_1 + C_2\tau)} \leq 1.
\]
By simply setting $\tau \leq \min{\lbrace \frac{5}{4\sqrt{\Delta}}, 2\kappa, \sqrt{\frac{2\kappa}{\sqrt{\Delta}}} \rbrace}$, the above constraint is satisfied with $\rho \leq 0.979$, which implies that the total oracle complexity is $\bigO{(n + \kappa)\log{\frac{F(\tilde{x}_0) - F(x^*)}{\epsilon}}}$.


\section{Experimental Setup}
\label{experiment_setup}
All our algorithms were implemented in C++ and parameters were passed through MATLAB interface for fair comparison\footnote{The code of our method can be downloaded from the anonymous link:\\ \url{https://www.dropbox.com/s/1a5v1gvioqbjtv3/Async_Sparse_MiG.zip?dl=0}.}. Detailed settings are divided into the following cases.

\subsection{In Serial Dense Case}
\label{exp_parameter_settings}
In this case, we ran experiments on the HP Z440 machine with single Intel Xeon E5-1630v4 with 3.70GHz cores, 16GB RAM, Ubuntu 16.04 LTS with GCC 4.9.0, MATLAB R2017b. We are optimizing the following binary problems with $a_i \in \R^d$, $b_i \in \lbrace -1, +1\rbrace$, $i=1\ldots m$:
\begin{equation}
\begin{split}
\textup{Logistic Regression: }& \frac{1}{n} \sum_{i=1}^n {\log{(1 + \exp{(-b_ia_i^Tx))}}} + \frac{\lambda}{2} \norm{x}^2,\\
\!\!\!\!\!\!\textup{Ridge Regression: }& \frac{1}{n} \sum_{i=1}^n {(a_i^Tx + b_i)^2} + \frac{\lambda}{2} \norm{x}^2,\\
\!\!\!\!\!\!\textup{LASSO: }& \frac{1}{n} \sum_{i=1}^n {(a_i^Tx + b_i)^2} + \lambda\norm{x}_1,
\end{split}
\end{equation}
where $\lambda$ is the regularization parameter.

We used datasets from LibSVM website\footnote{https://www.csie.ntu.edu.tw/{\raise.17ex\hbox{$\scriptstyle\mathtt{\sim}$}}cjlin/libsvmtools/datasets/}, including \textsf{a9a} (32,561 samples, 123 features), \textsf{covtype.binary} (581,012 samples, 54 features), \textsf{w8a} (49,749 samples, 300 features), \textsf{ijcnn1} (49,990 samples, 22 features). We added one dimension as bias to all the datasets and then normalized all data vectors to $1$ for the ease of experimental setup.

We mainly compared MiG with the following state-of-the-art algorithms:
\begin{itemize}
	\item SVRG. For theoretical evaluation, we set the learning rate as $\frac{1}{4L}$, which is a reasonable learning rate for SVRG in theory. Otherwise in the strongly convex case, we tuned the learning rate. For non-smooth regularizers (e.g., LASSO), we ran Prox-SVRG~\cite{xiao:prox-svrg} instead.
	\item SAGA. For theoretical evaluation, we set the learning rate as $\frac{1}{2(\sigma n + L)}$ following~\cite{defazio:saga}. Otherwise (include the non-strongly convex case), we tuned the learning rate.
	\item Acc-Prox-SVRG. This algorithm is quite unstable if mini-batch size is set to $1$, we set same learning rate as in SVRG and tuned the momentum parameter $\beta$~\cite{nitanda:svrg}.
	\item Catalyst on SVRG. We set the same learning rate as in SVRG and carefully tuned the parameters $\alpha_0$ and $\kappa$~\cite{lin:vrsg}.
	\item Katyusha. As suggested by the author, we fixed $\tau_2 = \frac{1}{2}$ (sometimes we tuned $\tau_2$ for a better performance), set $\eta = \frac{1}{3\tau_1 L}$ and tuned only $\tau_1$~\cite{zhu:Katyusha}. For theoretical evaluation, we chose $\tau_1 = \sqrt{\frac{m}{3\kappa}}$. For Katyusha\textsuperscript{ns}, we used $\tau_1 = \frac{2}{s + 4}$, $\alpha = \frac{1}{a\tau_1 L}$ and tune $a$.
	\item MiG. Similarly, we set $\eta = \frac{1}{3 \theta L}$ and tuned only $\theta$. For theoretical evaluation, we chose $\theta = \sqrt{\frac{m}{3\kappa}}$. For MiG\textsuperscript{NSC}, we chose $\theta = \frac{2}{s + 4}$, $\eta = \frac{1}{aL\theta}$ and tuned $a$.
\end{itemize}

\begin{figure}[H]
	\begin{center}
		\centerline{
			\includegraphics[width=\columnwidth / 4]{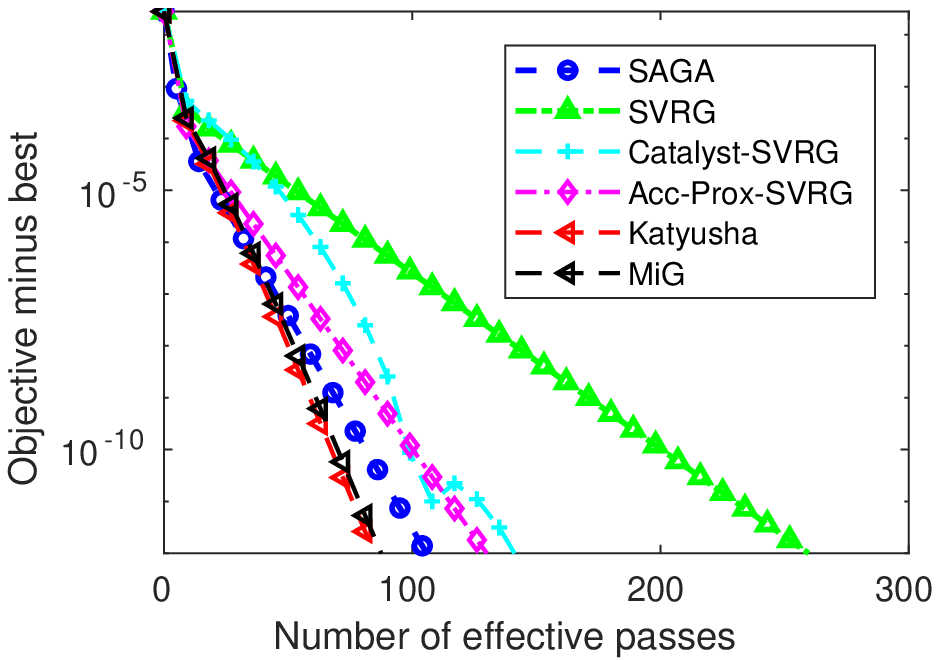}
			\includegraphics[width=\columnwidth / 4]{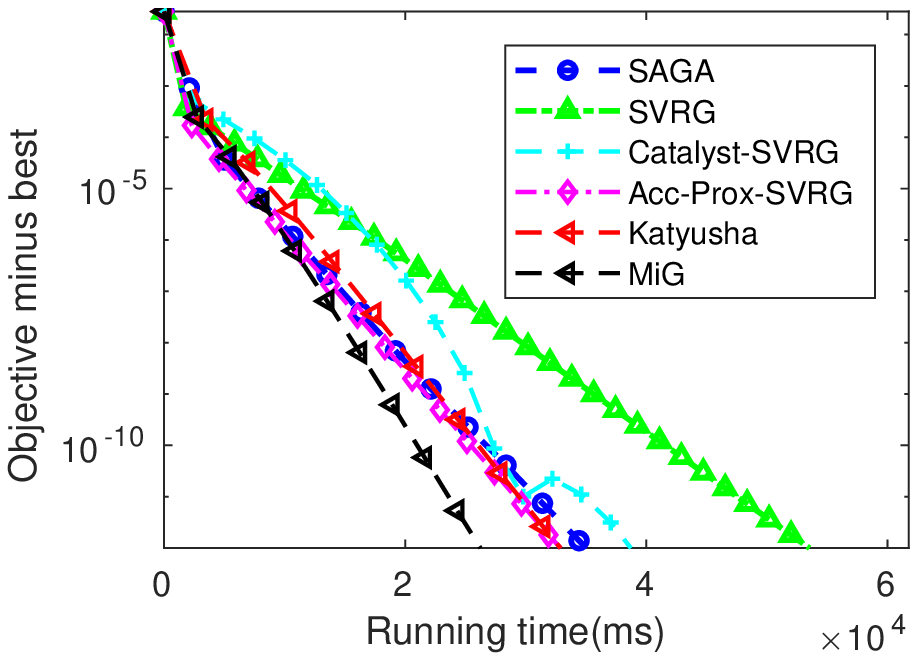}
			\includegraphics[width=\columnwidth / 4]{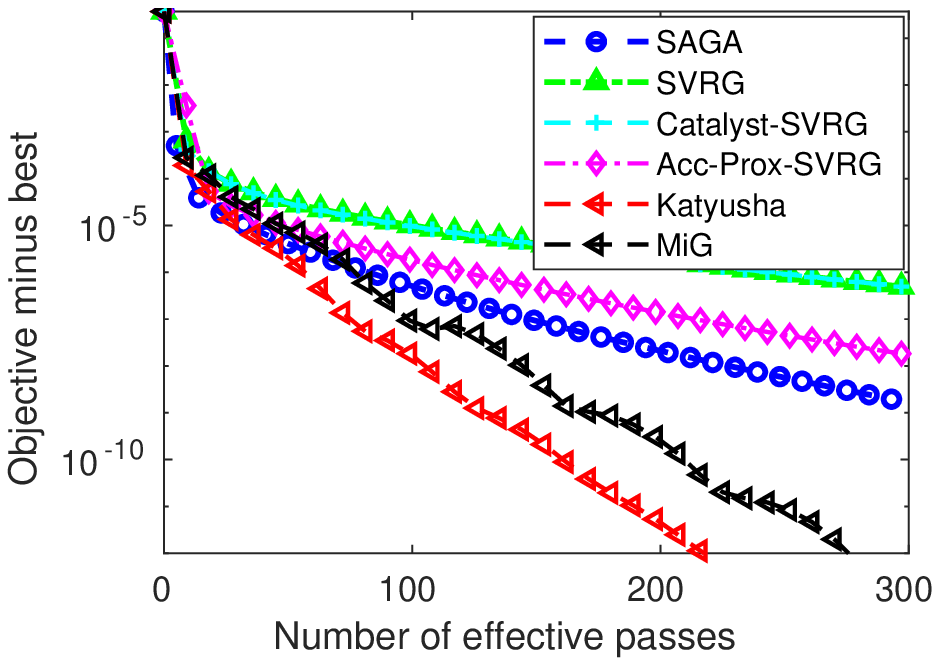}
			\includegraphics[width=\columnwidth / 4]{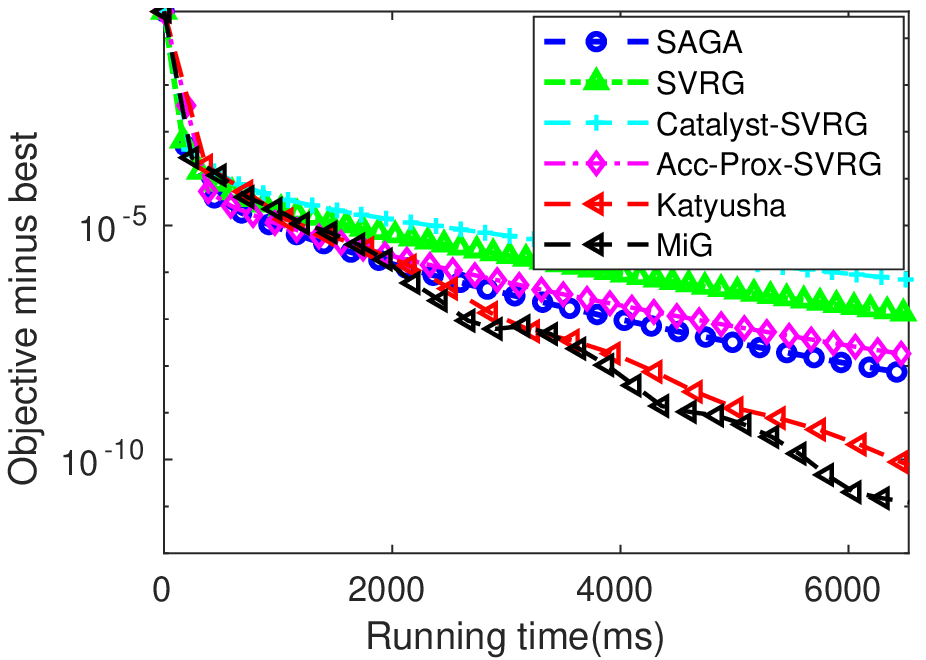}
		}
		\caption{Theoretical evaluations of MiG and other state-of-the-art algorithms for solving $\ell2$-logistic regression on \textsf{covtype} ($\lambda = 10^{-7}$, the first two figures) and \textsf{a9a} ($\lambda = 10^{-7}$, the last two figures).}
		\label{exp_theory_1}
	\end{center}
\end{figure}

\begin{figure}[H]
	\begin{center}
		\centerline{
			\includegraphics[width=\columnwidth / 4]{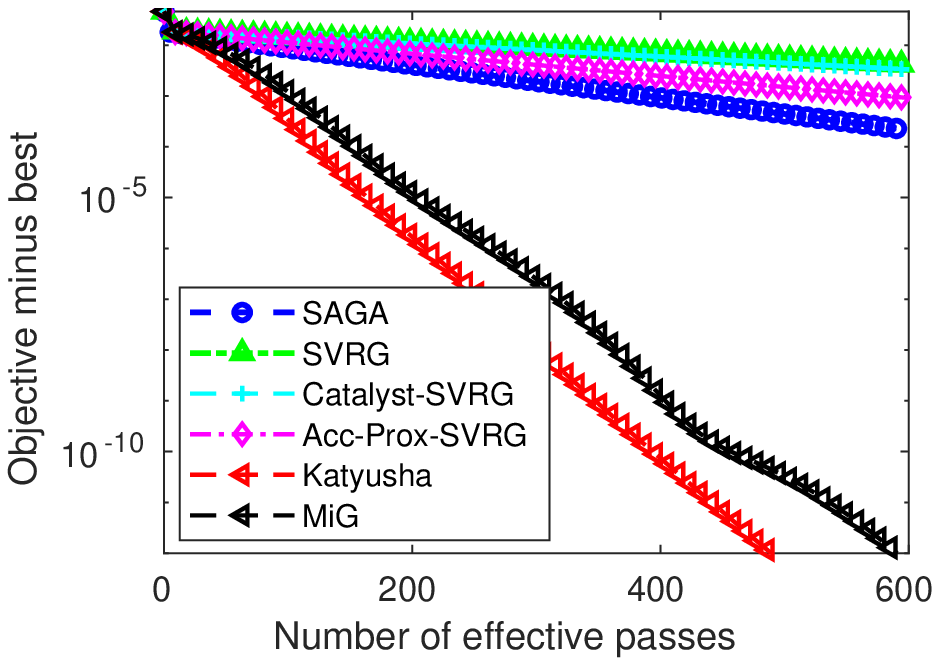}
			\includegraphics[width=\columnwidth / 4]{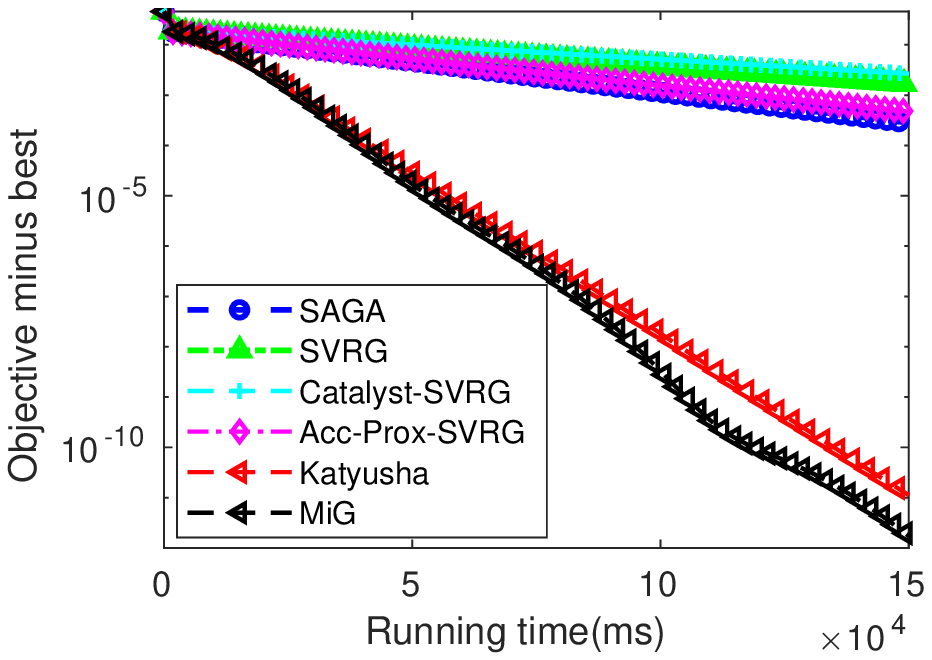}
			\includegraphics[width=\columnwidth / 4]{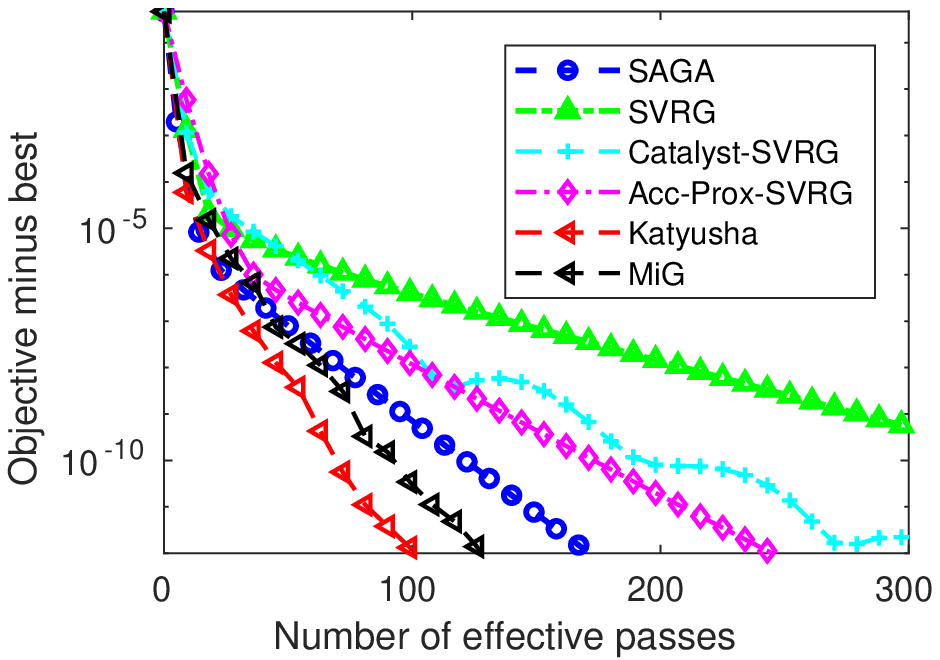}
			\includegraphics[width=\columnwidth / 4]{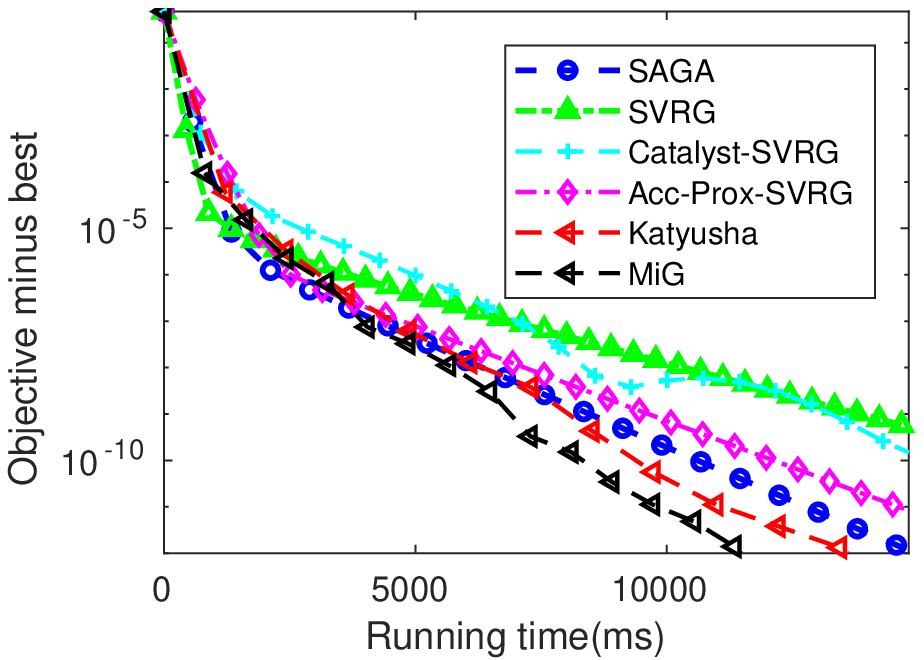}
		}
		\caption{Theoretical evaluations of MiG and other state-of-the-art algorithms for solving ridge regression on \textsf{covtype} ($\lambda = 10^{-8}$, the first two figures) and \textsf{w8a} ($\lambda = 10^{-6}$, the last two figures).}
		\label{exp_theory_2}
	\end{center}
\end{figure}

More theoretical evaluation results for $\ell2$-logistic regression and ridge regression problems are shown in Figures~\ref{exp_theory_1} and~\ref{exp_theory_2}, respectively, where the regularization parameter was set to some relatively small values, e.g., $10^{-6}$, $10^{-7}$, and $10^{-8}$.

\begin{figure}[H]
\centering
	\subfigure[covtype: $\lambda = 10^{-5}$]{\includegraphics[width=0.239\columnwidth]{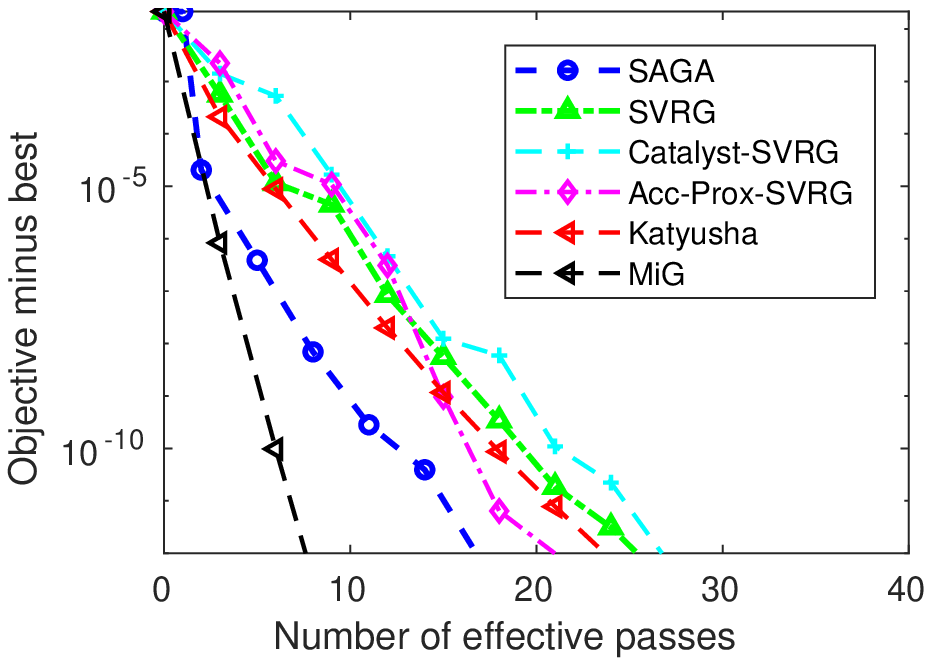}
	\includegraphics[width=0.239\columnwidth]{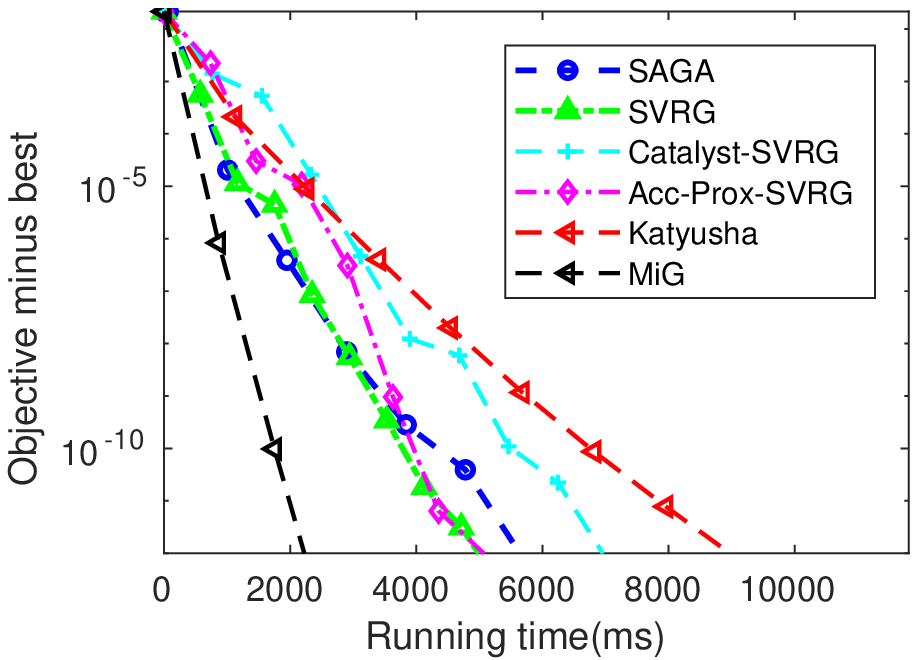}}\;
	\subfigure[ijcnn1: $\lambda = 10^{-4}$]{\includegraphics[width=0.239\columnwidth]{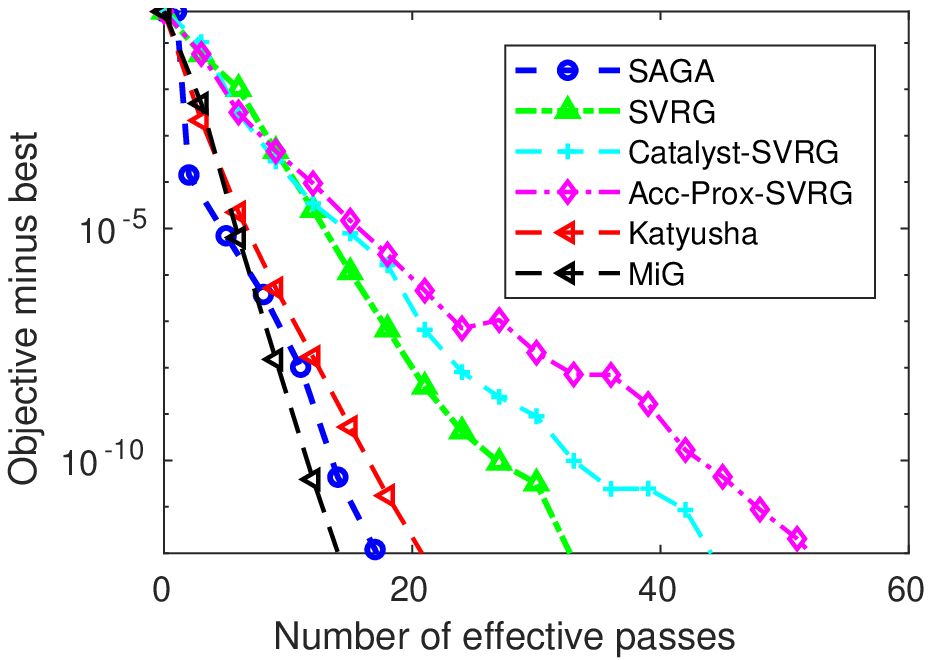}
	\includegraphics[width=0.239\columnwidth]{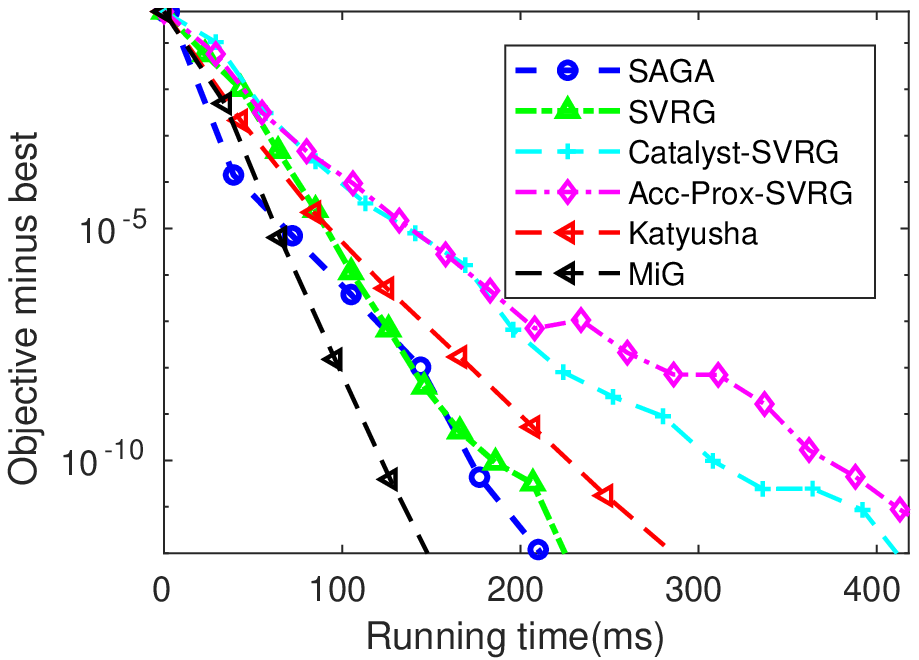}}
	\subfigure[w8a: $\lambda = 10^{-4}$]{\includegraphics[width=0.239\columnwidth]{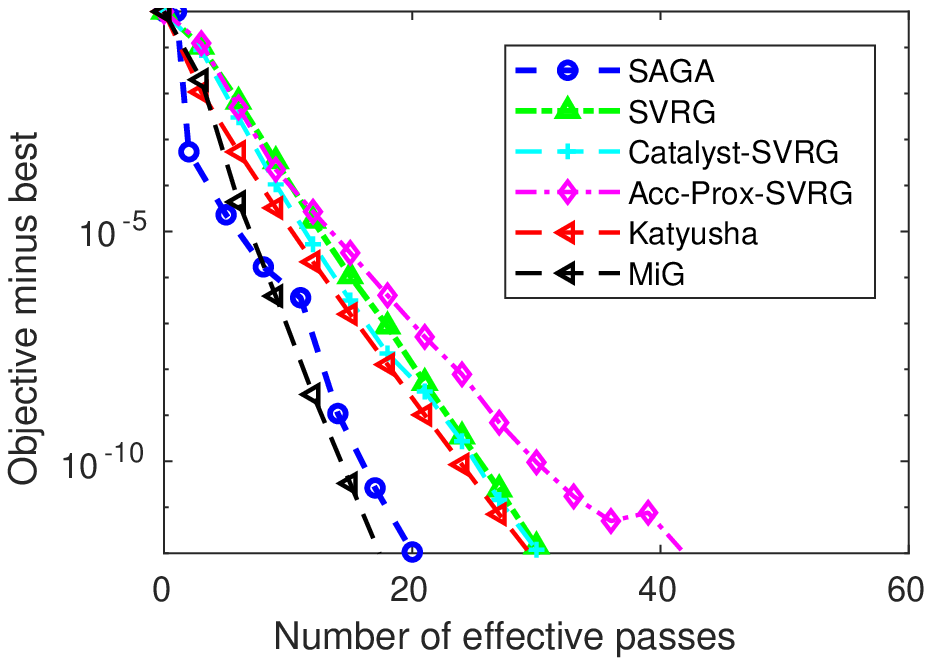}
	\includegraphics[width=0.239\columnwidth]{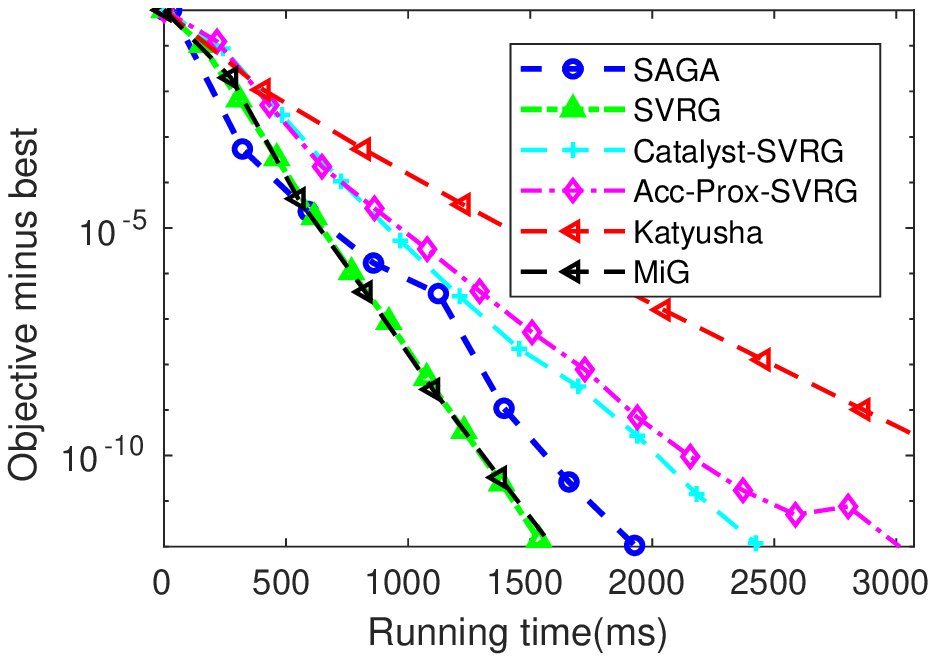}}\;
    \subfigure[a9a: $\lambda = 10^{-4}$]{\includegraphics[width=0.239\columnwidth]{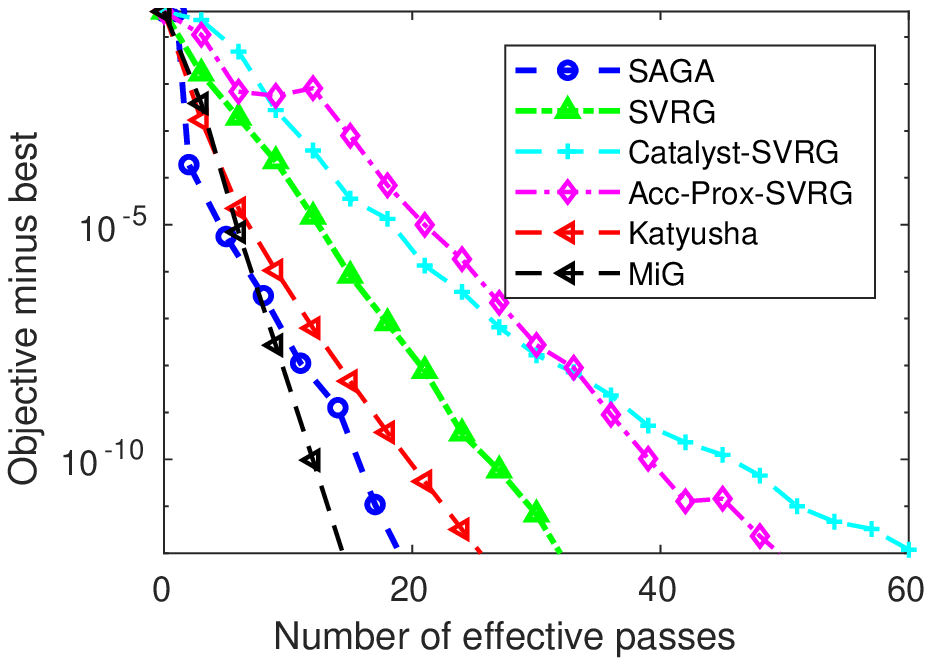}
	\includegraphics[width=0.239\columnwidth]{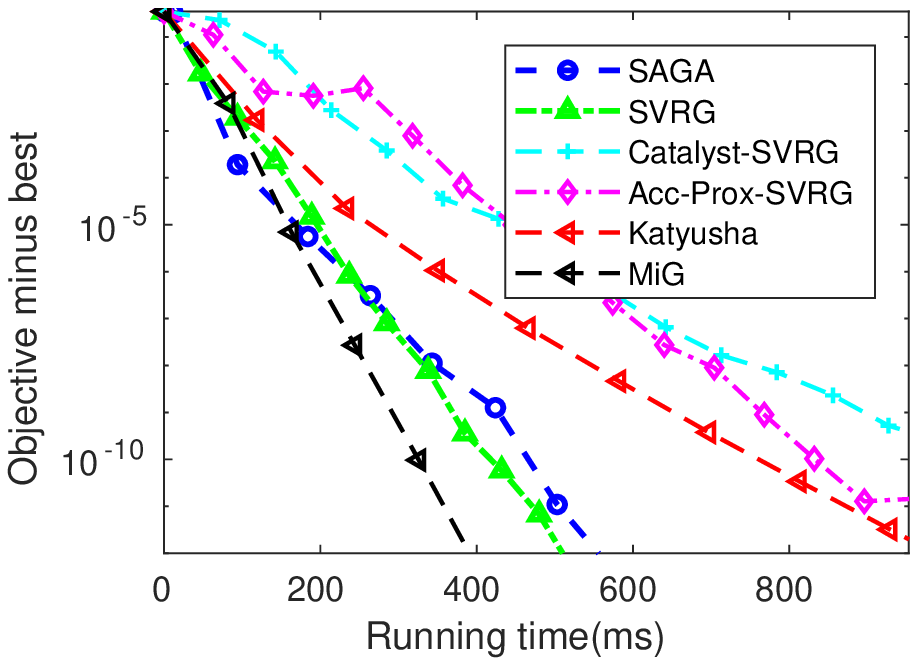}}
\caption{Practical evaluations of MiG and other state-of-the-art algorithms for solving $\ell2$-logistic regression on \textsf{covtype} (a), \textsf{ijcnn1} (b), \textsf{w8a} (c), and \textsf{a9a} (d).}
\label{exp_practice_1}
\end{figure}

\begin{figure}[H]
\centering
	\subfigure[covtype: $\lambda = 10^{-4}$]{\includegraphics[width=0.239\columnwidth]{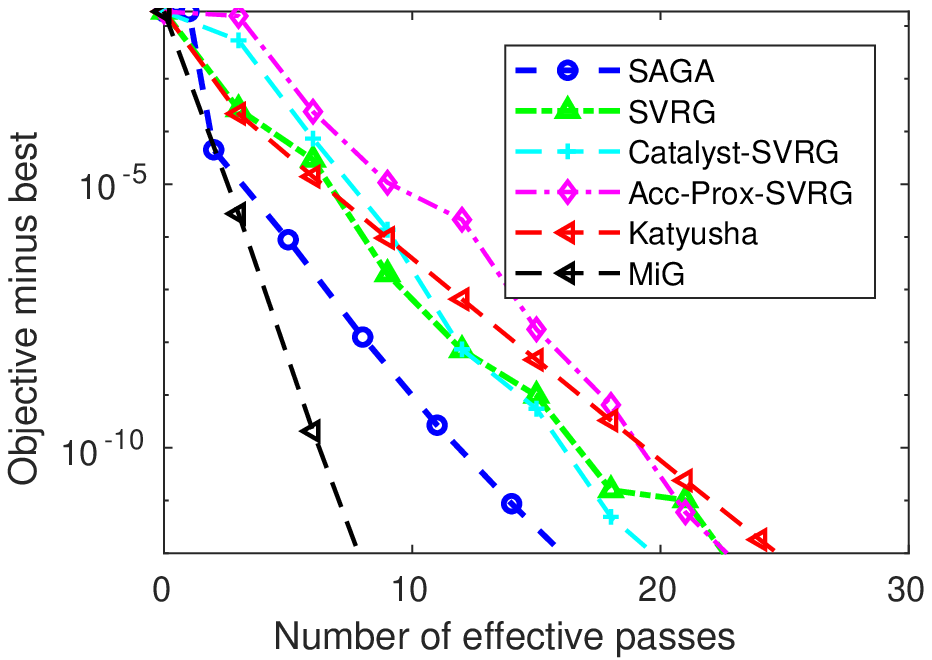}
	\includegraphics[width=0.239\columnwidth]{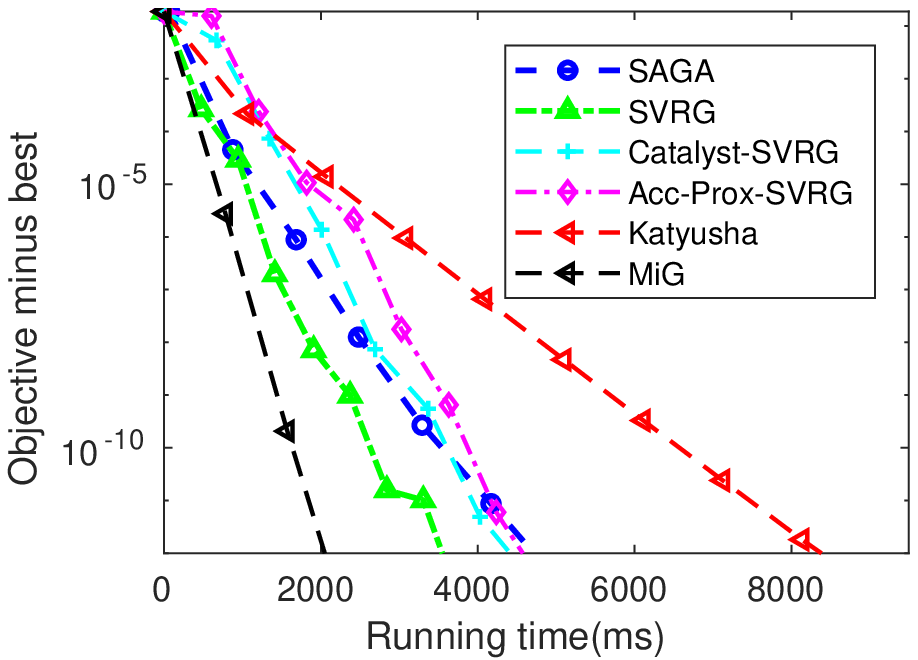}}\;
	\subfigure[ijcnn1: $\lambda = 10^{-4}$]{\includegraphics[width=0.239\columnwidth]{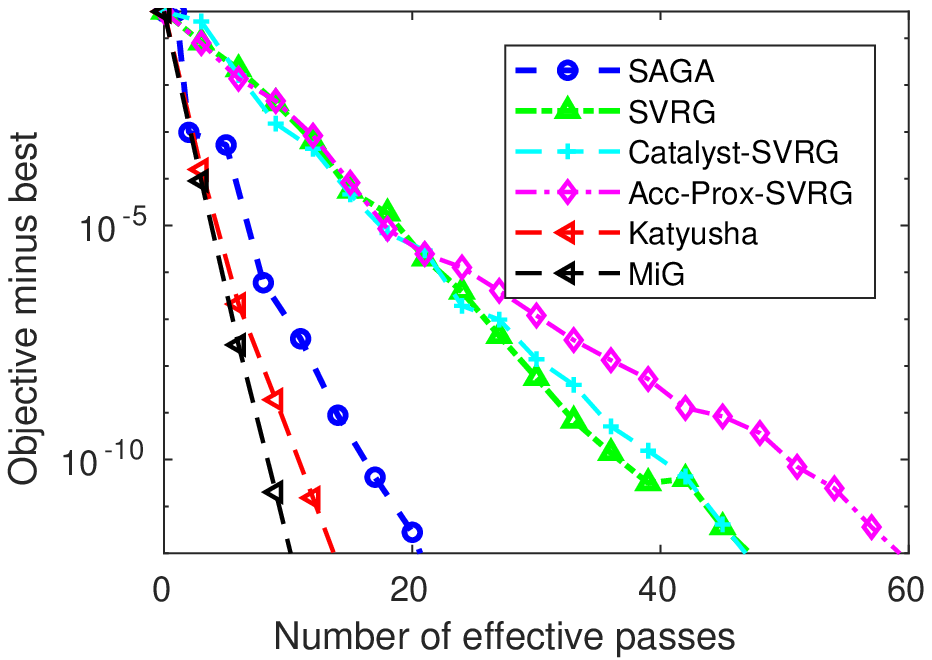}
	\includegraphics[width=0.239\columnwidth]{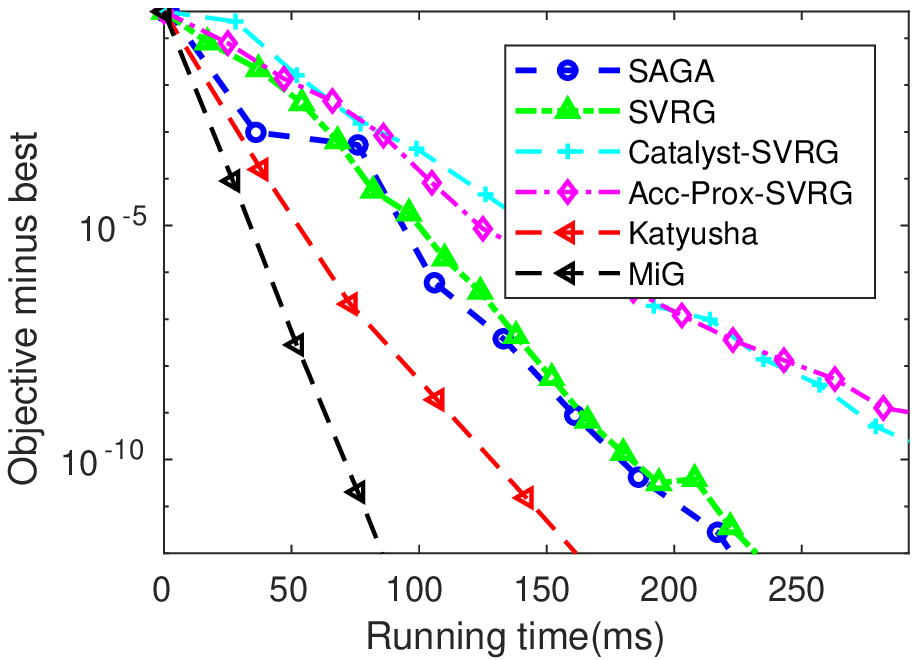}}
	\subfigure[covtype: $\lambda = 10^{-5}$]{\includegraphics[width=0.239\columnwidth]{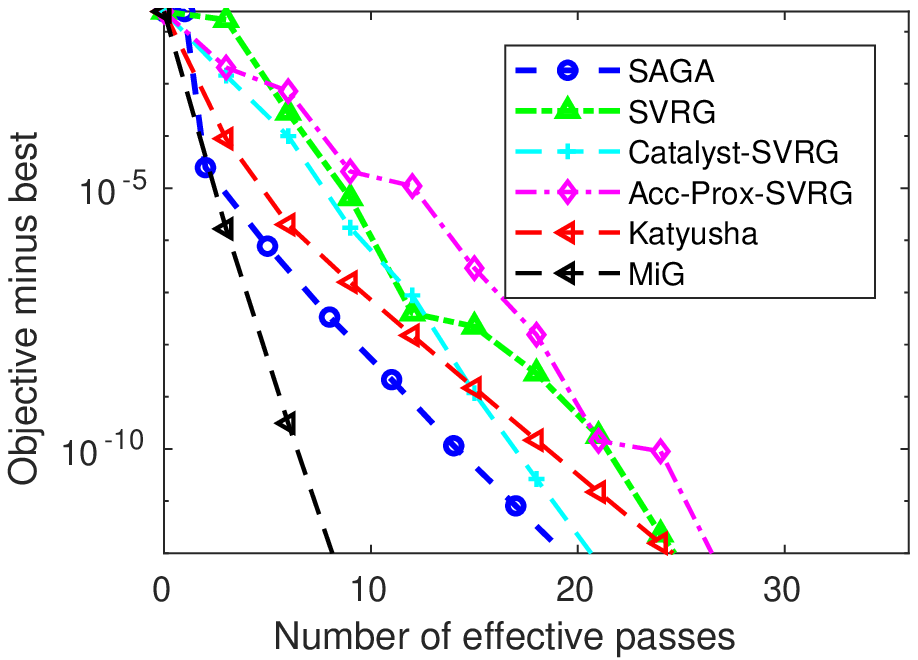}
	\includegraphics[width=0.239\columnwidth]{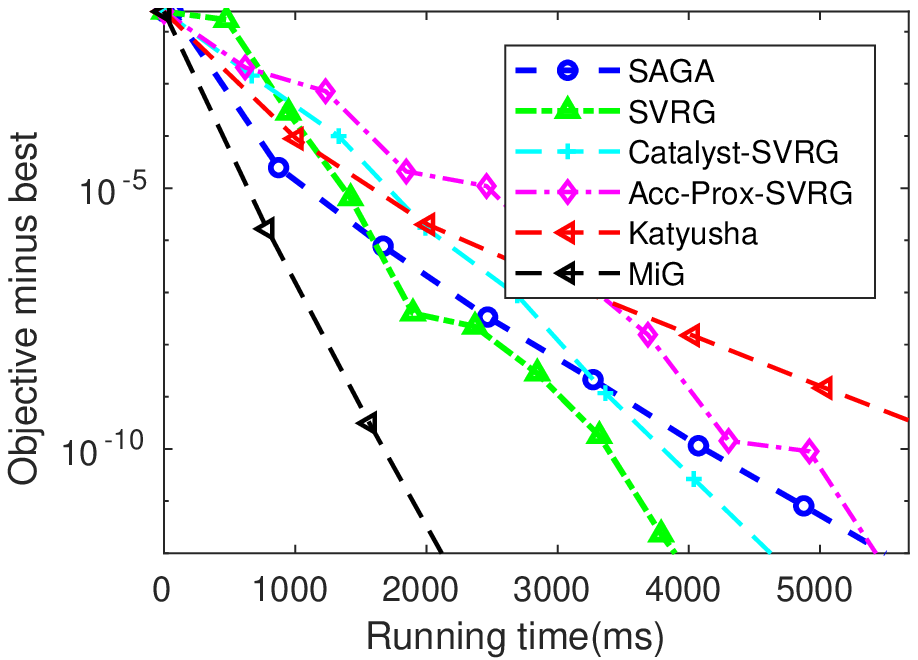}}\;
	\subfigure[w8a: $\lambda = 10^{-4}$]{\includegraphics[width=0.239\columnwidth]{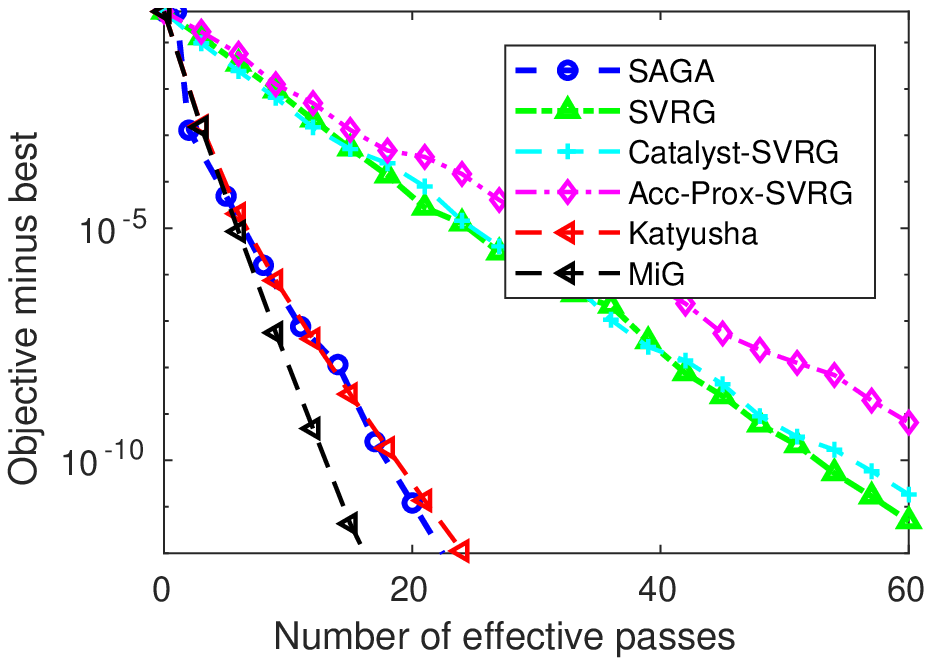}
	\includegraphics[width=0.239\columnwidth]{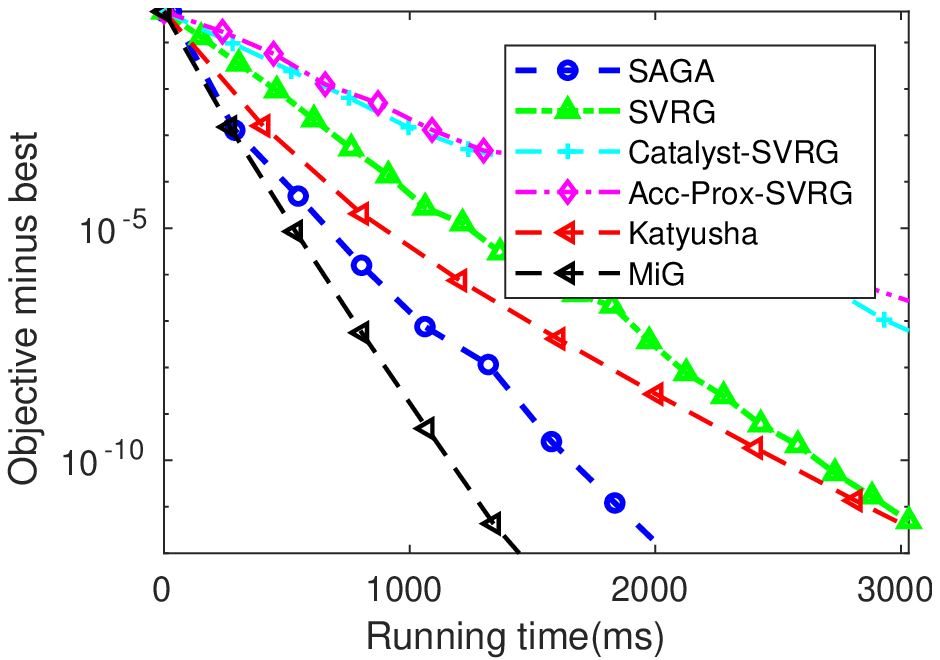}}
\caption{Practical evaluations of MiG and other state-of-the-art algorithms for solving ridge regression on \textsf{covtype} (a), \textsf{ijcnn1} (b), \textsf{covtype} (c), and \textsf{w8a} (d).}
		\label{exp_practice_4}
\end{figure}

More practical evaluation results (with parameter tuning for all the algorithms and for relatively large $\lambda$) are shown in Figures~\ref{exp_practice_1} and \ref{exp_practice_4}.

\begin{figure}[H]
	\vskip 0.2in
	\begin{center}
		\centerline{
			\includegraphics[width=\columnwidth / 4]{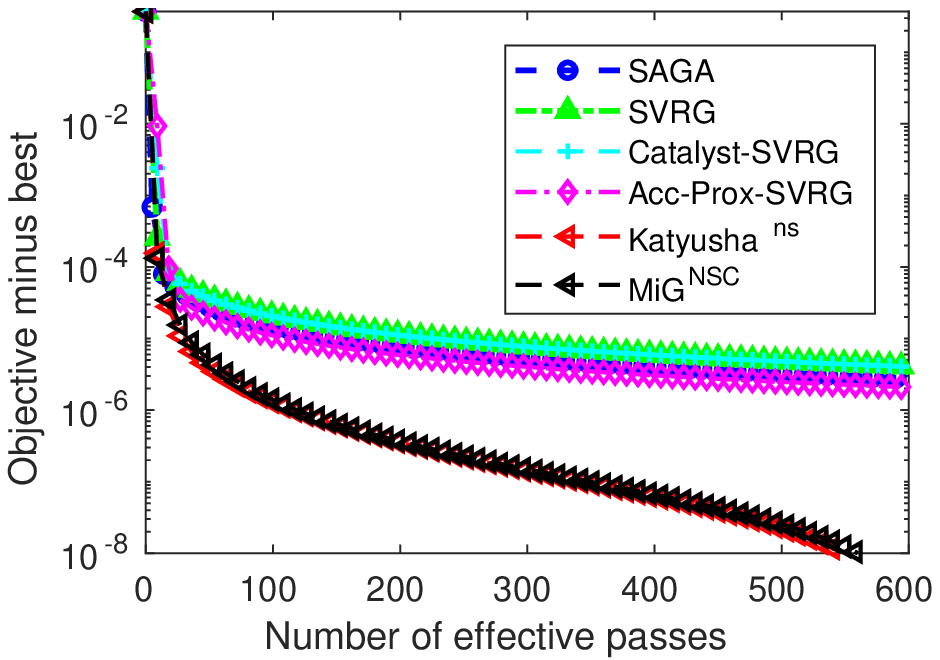}
			\includegraphics[width=\columnwidth / 4]{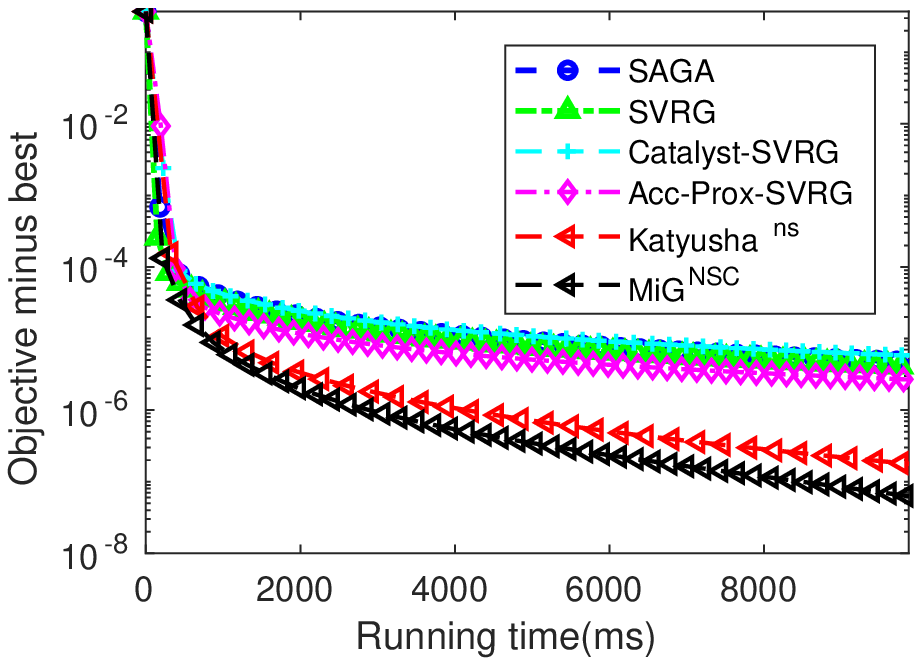}
			\includegraphics[width=\columnwidth / 4]{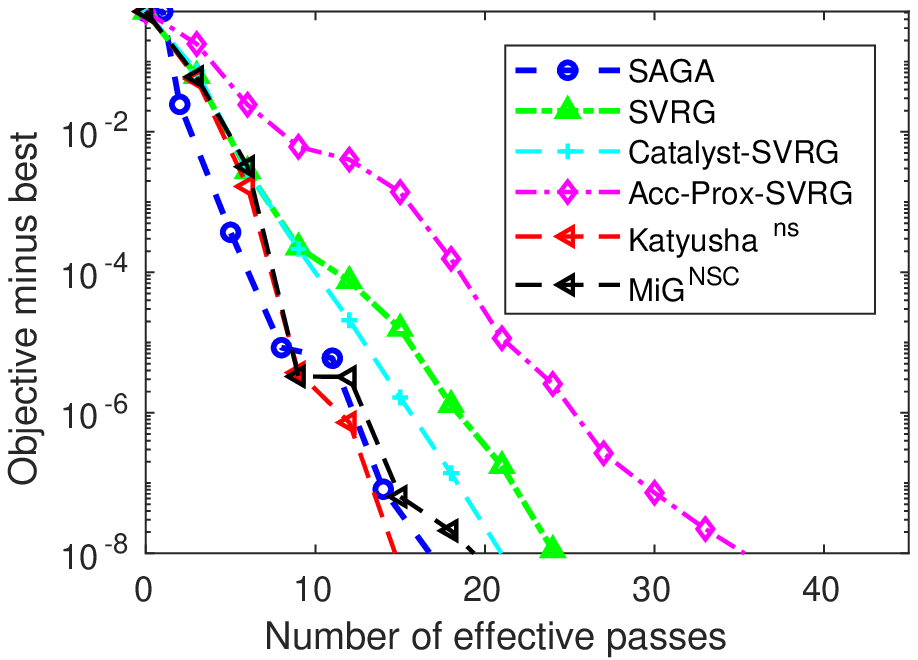}
			\includegraphics[width=\columnwidth / 4]{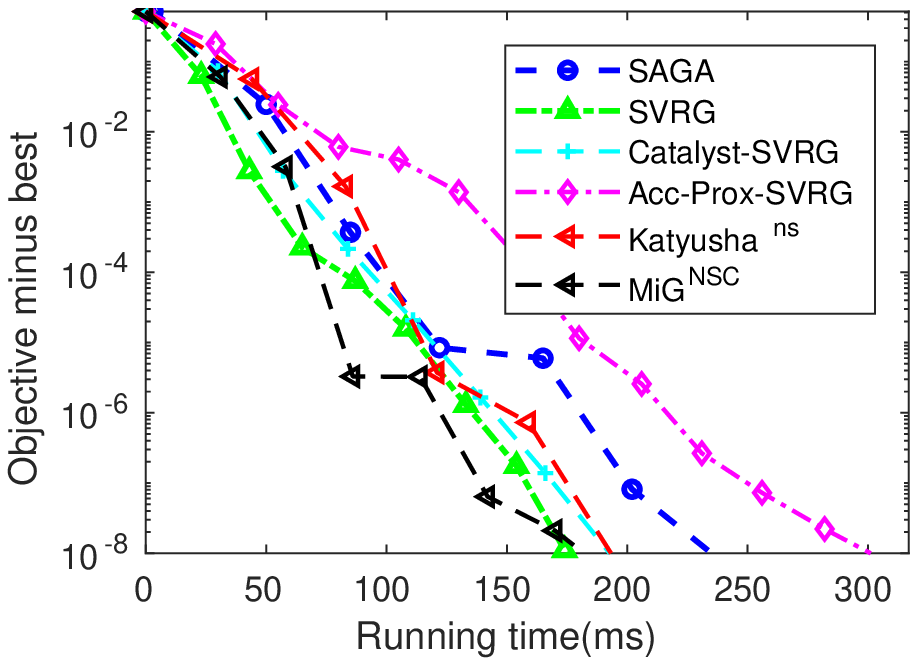}
		}
		\caption{Comparison of MiG\textsuperscript{NSC} and other state-of-the-art algorithms for solving non-strongly convex logistic regression without regularizer (i.e., $\lambda=0$) on \textsf{a9a} (the first two figures) and \textsf{ijcnn1} (the last two figures).}
		\label{exp_nsc_1}
	\end{center}
\end{figure}

\begin{figure}[H]
	\begin{center}
		\centerline{
			\includegraphics[width=\columnwidth / 4]{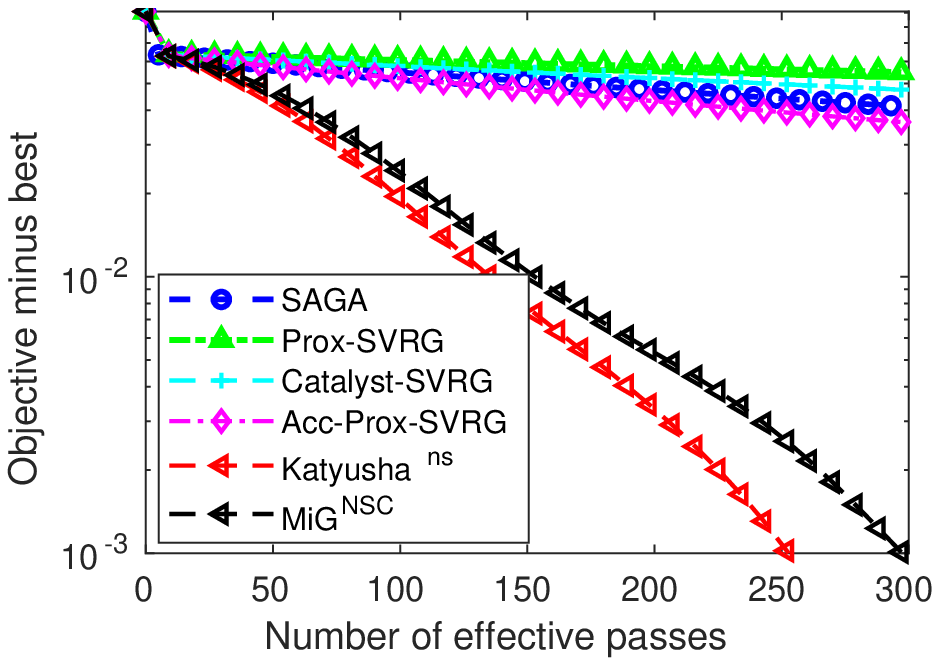}
			\includegraphics[width=\columnwidth / 4]{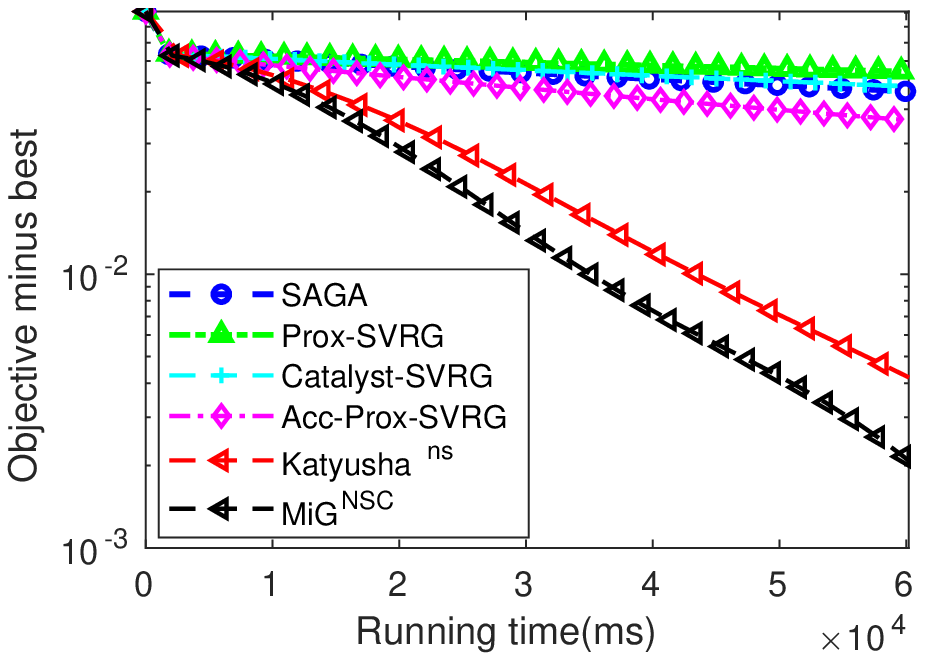}
			\includegraphics[width=\columnwidth / 4]{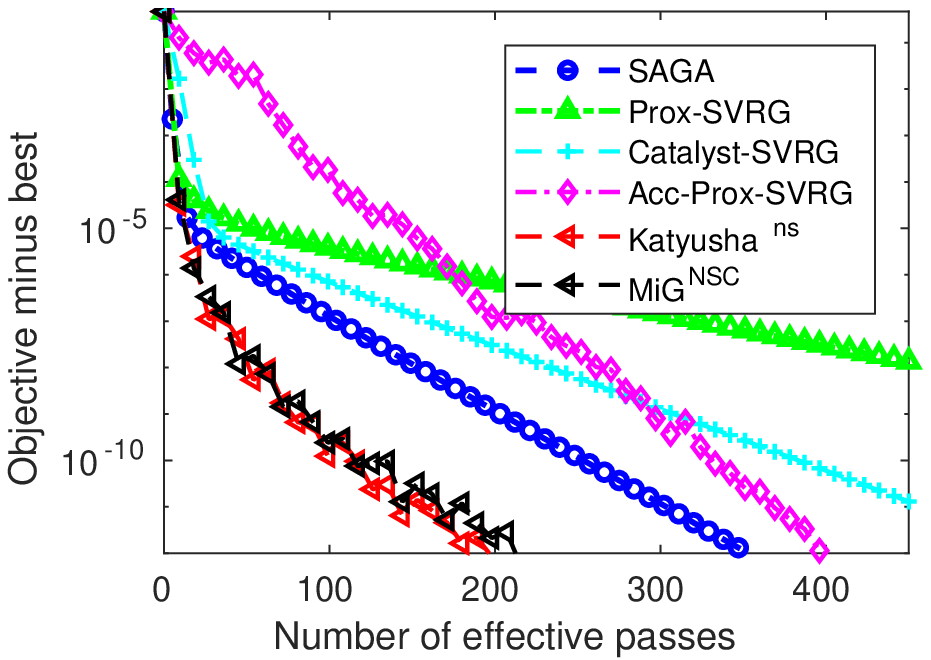}
			\includegraphics[width=\columnwidth / 4]{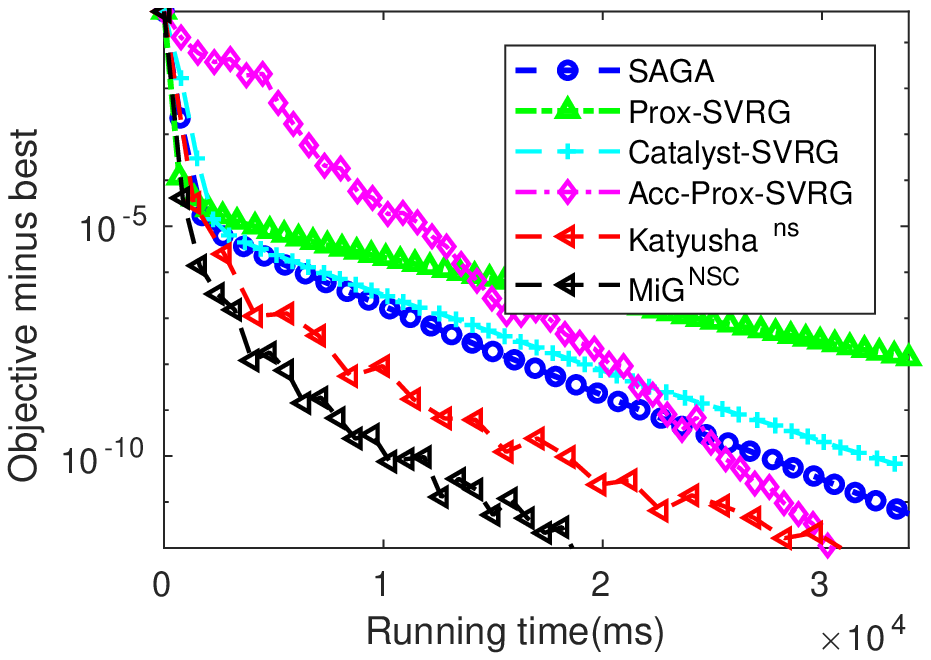}
		}
		\caption{Comparison of MiG\textsuperscript{NSC} and other state-of-the-art algorithms for solving the non-strongly convex problem (LASSO) on \textsf{covtype} ($\lambda = 10^{-6}$, the first two figures) and \textsf{w8a} ($\lambda = 10^{-7}$, the last two figures).}
		\label{exp_nsc_2}
	\end{center}
\end{figure}

We also compared the performance of MiG\textsuperscript{NSC} with that of all the algorithms mentioned above, as shown in Figures~\ref{exp_nsc_1} and~\ref{exp_nsc_2}, where the results are for the non-strongly convex logistic regression (i.e., $\lambda\!=\!0$) and LASSO, respectively.


\subsection{In Asynchronous Sparse Case}
\label{async_exp_parameter_settings}
Experiments in this setting were running on a server with 4 Intel Xeon E7-4820 v3 with 10 cores each 1.90GHz, 512 GB RAM, CentOS 7.4.1708 with GCC 4.8.5, MATLAB R2017a. Multi-threads experiments were running on certain CPU cores using \textsf{numactl}. We are optimizing the following smooth and strongly convex problem:
\begin{equation*}
	 \textup{Logistic Regression: }\frac{1}{n} \sum_{i=1}^n {\log{(1 + \exp{(-b_ia_i^Tx))}}} + \frac{\lambda}{2} \norm{x}^2,
\end{equation*}
where $\lambda$ is the regularization parameter. Notice that the $\ell2$ regularizer is dense with respect to the sample $i$, so we used the following sparse and unbiased stochastic gradient as in~\cite{leb:asaga}:
\[
	\nabla_i(x) = \nabla f_i(x) + \lambda D_i x.
\]
We tuned learning rates for ASAGA and KroMagnon, learning rate and $\theta$ for MiG.

\subsubsection{Parameter Tuning Criteria}
For the relatively small RCV1 dataset, we carefully tuned the learning rate for ASAGA and KroMagnon to achieve best performance. For MiG, following theoretical intuition in the dense case, we first chose a large learning rate and then carefully tuned $\theta$ to achieve best performance (thus the effect of $\theta$ can be regard as stabilize iterates and allow for a larger learning rate).

For the KDD2010 dataset, we first made 3 subsets with 19,000 samples, 190,000 samples and 1,900,000 samples (corresponds to $\lambda = 10^{-7}$, $10^{-8}$ and $10^{-9}$). Then we tuned parameters for them following the above criteria and finally used relatively ``safe'' parameter settings for the entire dataset.

\subsubsection{On the Effectiveness of Our Acceleration Trick}
\label{effective_theta}
In this part, we evaluated the effect of our acceleration trick in MiG. One can easily verify that when $\theta$ is set to $1$, the coupling term is neglected and the algorithm is quite similar to SVRG (different on the choice of the snapshot point). Based on the theoretical analysis in Section~\ref{sequential} and Section~\ref{sec_serial_sparse_mig}, we claim that the effect of $\theta$ is to stabilize the iterates to adopt a larger learning rate. Thus, we designed an experiment to justify this claim.

\begin{figure}[H]
	\vskip 0.2in
	\begin{center}
		\centerline{
			\includegraphics[width=\columnwidth / 3]{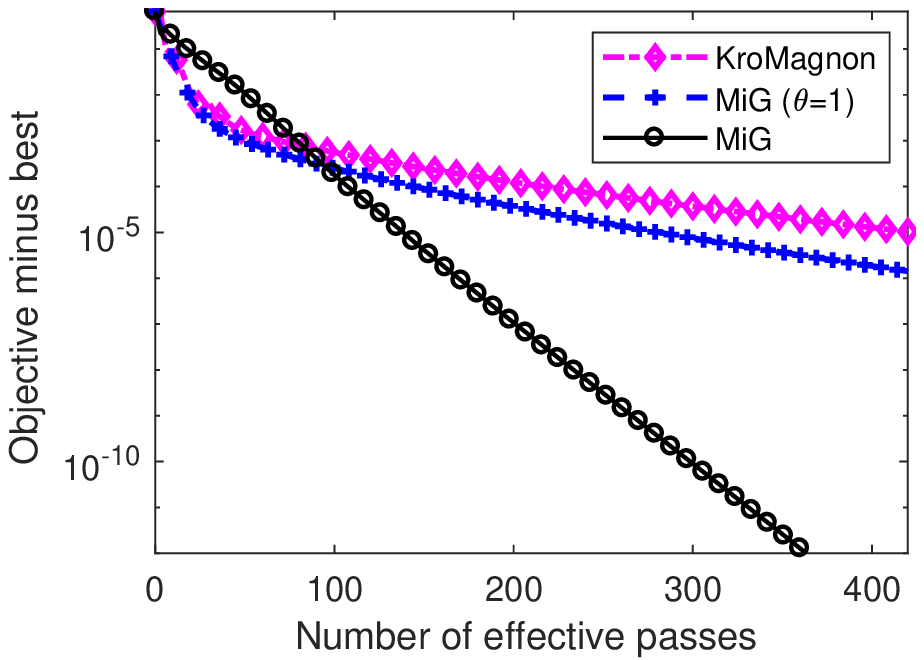}\qquad
			\includegraphics[width=\columnwidth / 3]{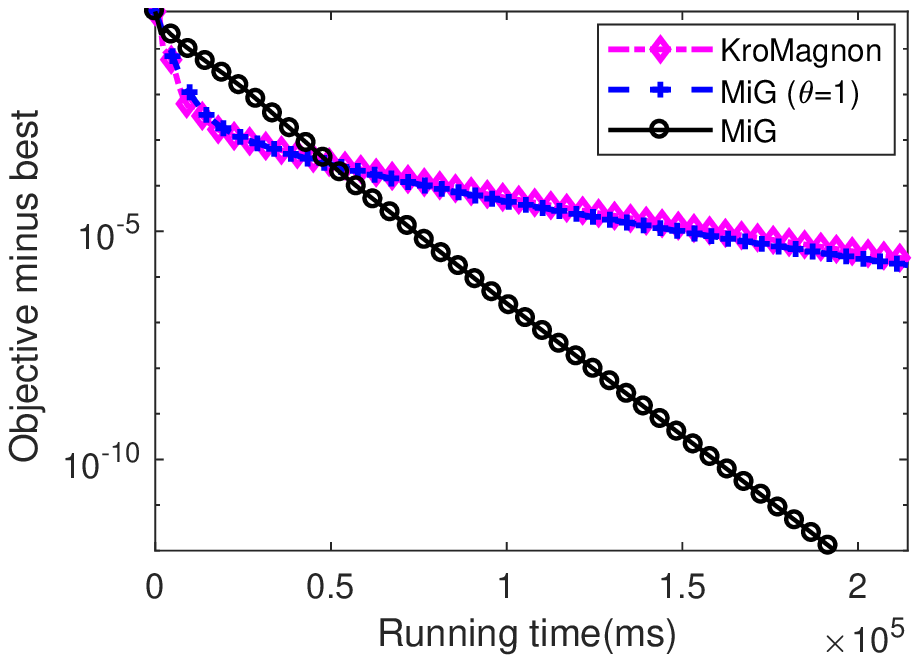}
		}
		\caption{Evaluation of the effectiveness of our acceleration trick of our algorithm for solving $\ell2$-logistic regression with $\lambda = 10^{-9}$.}
		\label{exp_async_comp}
	\end{center}
\end{figure}

We compared the best-tuned performance of MiG, MiG with $\theta\!=\!1$, and KroMagnon on \textsf{RCV1} on a server with the 16 threads, as shown in Figure~\ref{exp_async_comp}.

Based on the parameter choice in this experiment, MiG with $\theta\!=\!1$ cannot use the same large learning rate as MiG does. When the suitable parameter $\theta$ is chosen, MiG outperforms both algorithms as desired, which further indicates the effectiveness of our acceleration trick.

\end{document}